\documentclass[preprint,10pt]{elsarticle}
\usepackage{mathrsfs}
\usepackage{subcaption}
\usepackage{mathtools}
\usepackage{pifont}
\usepackage{algorithmic}
\usepackage{algorithm}
\usepackage{xcolor}
\usepackage{color}
\usepackage{lineno}
\usepackage{makecell} 
\usepackage{pdflscape}
\usepackage{adjustbox}
\usepackage[utf8]{inputenc}
\usepackage{tabularx}
\usepackage{blindtext}
\usepackage{multirow}
\usepackage{notoccite} 
\usepackage{lscape} 
\usepackage{caption} 
\usepackage{mwe}
\usepackage{tikz}
\usepackage{siunitx}
\usepackage{mathrsfs}
\usepackage{amsmath,amsfonts}
\usepackage{amsmath}
\usepackage{mathtools}
\usepackage{mathrsfs}
\usetikzlibrary{shapes,arrows}
\usepackage{xcolor}
\usepackage{booktabs}
\usepackage{hyperref}
\usepackage{amsthm}

\newcommand{\RNum}[1]{\lowercase\expandafter{\romannumeral #1\relax}}
\newcommand{\RNumU}[1]{\uppercase\expandafter{\romannumeral #1\relax}}
\usepackage{natbib}
\journal{Elsevier}

\begin{document}
\date{}
\begin{frontmatter}
\title{Multiview learning with twin parametric margin SVM}
\author[inst1]{A. Quadir}
\affiliation[inst1]{organization={Department of Mathematics, Indian Institute of Technology Indore},
            addressline={Simrol}, 
            city={Indore},
            postcode={453552}, 
            state={Madhya Pradesh},
            country={India}}
\author[inst1]{M. Tanveer\texorpdfstring{\corref}{corref}{Correspondingauthor}}
 \cortext[Correspondingauthor]{Corresponding author}

\begin{abstract}
Multiview learning (MVL) seeks to leverage the benefits of diverse perspectives to complement each other, effectively extracting and utilizing the latent information within the dataset. Several twin support vector machine-based MVL (MvTSVM) models have been introduced and demonstrated outstanding performance in various learning tasks. However, MvTSVM-based models face significant challenges in the form of computational complexity due to four matrix inversions, the need to reformulate optimization problems in order to employ kernel-generated surfaces for handling non-linear cases, and the constraint of uniform noise assumption in the training data. Particularly in cases where the data possesses a heteroscedastic error structure, these challenges become even more pronounced. In view of the aforementioned challenges, we propose multiview twin parametric margin support vector machine (MvTPMSVM). MvTPMSVM constructs parametric margin hyperplanes corresponding to both classes, aiming to regulate and manage the impact of the heteroscedastic noise structure existing within the data. The proposed MvTPMSVM model avoids the explicit computation of matrix inversions in the dual formulation, leading to enhanced computational efficiency. We perform an extensive assessment of the MvTPMSVM model using benchmark datasets such as UCI, KEEL, synthetic, and Animals with Attributes (AwA). Our experimental results, coupled with rigorous statistical analyses, confirm the superior generalization capabilities of the proposed MvTPMSVM model compared to the baseline models. The source code of the proposed MvTPMSVM model is available at \url{https://github.com/mtanveer1/MvTPMSVM}.
\end{abstract}

\end{frontmatter}
\section{Introduction}
Support vector machine (SVM) \cite{cortes1995support} is a widely acclaimed and proficient machine learning technique employed for classification and regression problems. SVM strives to maximize the margin between two classes, aiming to find the optimal hyperplane between two parallel supporting hyperplanes by solving a quadratic programming problem (QPP). A QPP is the process of solving certain mathematical optimization problems involving quadratic functions. QPP is minimizing or maximizing an objective function subject to bounds, linear equality, and inequality constraints. Over the decades, SVM has been effectively employed across a range of domains, including forest fire detection\cite{yang2023preferred}, pattern recognition \cite{sun2021multi}, significant memory concern (SMC) \cite{sajid2024decoding}, Alzheimer’s disease diagnosis \cite{richhariya2020diagnosis}, and so on. Although SVM has made significant strides in various domains, there is still considerable potential for enhancement. A notable obstacle associated with the standard SVM is the substantial computational intricacy associated with solving the QPP. In order to reduce the computational complexity of SVM, \citet{mangasarian2005multisurface} and \citet{khemchandani2007twin} introduced the generalized eigenvalue proximal SVM (GEPSVM) and the twin SVM (TSVM), respectively. GEPSVM solves the generalized eigenvalue problem instead of QPP, and TSVM tackles two smaller QPPs as opposed to a single large one, resulting in TSVM being four times faster than the standard SVM \cite{tanveer2022comprehensive, shao2011improvements}. This firmly establishes TSVM as the superior choice over SVM. Subsequently, a range of advancements to TSVM is introduced, including general TSVM with pinball loss function (Pin-GTSVM) \cite{tanveer2019general}, conditional probability function TSVM (CPTSVM) \cite{shao2023twin}, smooth linear programming TSVM (SLPTSVM) \cite{tanveer2015application},  sparse pinball TSVM (SPTWSVM) \cite{tanveer2019sparse}, large scale pinball TSVM (LPTWSVM) \cite{tanveer2022large}, improvement on reduced universum TSVM for imbalanced classes (IRUTSVM) \cite{moosaei2023inverse}, multi-task TSVM with Universum data (UMTSVM) \cite{moosaei2024multi} and granular ball TSVM with pinball loss (Pin-GBTSVM) \cite{abdul2024granular}. To further enhance the computational efficiency \citet{kumar2009least} proposed least squares TSVM (LSTSVM) by solving two linear systems of equations. This approach offers novel avenues for diminishing computational complexity, and it has spurred the development of various algorithms rooted in it such as intuitionistic fuzzy weighted LSTSVM (IFW-LSTSVM) \cite{tanveer2022intuitionistic}, large-scale fuzzy LSTSVM for class imbalance learning (LS-FLSTSVM-CIL) \cite{ganaie2022large}, laplacian $Lp$ norm LSTSVM (Lap-LpLSTSVM) \cite{xie2023laplacian}, least squares structural twin bounded SVM on class scatter (LS-STBSVM) \cite{gupta2023least}, energy-based LSTSVM (ELS-TSVM)\cite{tanveer2016robust} and least squares weighted multi-class TSVM (LS-KWMTSVM) \cite{tanveer2021least}. 

While the TSVM and its variants assume that the level of noise in the training data is consistent across the entire dataset \cite{scholkopf2000new}. Noise refers to random or irrelevant data that can interfere with the learning process by introducing errors or misleading patterns in the training data. The assumption of a uniform noise model may not hold true in real-world scenarios. In the case of a heteroscedastic noise structure, where noise strongly depends on the input value, the amount of noise varies based on the location. To generate a parametric-margin model, \citet{hao2010new} proposed the parametric margin $\nu$-support vector machine (par-$\nu$-SVM). This can be valuable in numerous scenarios, particularly when dealing with datasets exhibiting a heteroscedastic error structure, where the input values are greatly influenced by the noise. However, the par-$\nu$-SVMs learning rate is as slow as the standard SVM. To reduce the computational complexity, \citet{peng2011tpmsvm} proposed twin parametric-margin SVM (TPMSVM). TPMSVM constructs two distinct nonparallel hyperplanes, with each of them serving as the parametric-margin hyperplane for the $+1$ and $-1$ classes.

Multiview learning (MVL) concentrates on data sets that can be represented by multiple distinct feature sets. Different views often provide information complementary to each other. In contrast to traditional single-view learning, MVL constructs a learning function for each feature view separately and then jointly optimizes the learning function by exploiting redundant views of the same input data. MVL \cite{zhao2017multi} is a nascent avenue within the field of machine learning, dedicated to leveraging multiview data to enhance the overall performance of learning algorithms, particularly in terms of their generalization capabilities \cite{xie2020multi}. MVL has demonstrated its success in various domains, including financial analysis \cite{sun2021multi}, intelligent diagnostics \cite{yang2020inverse}, multiview clustering \cite{wang2019study}, and so on. However, a primary challenge currently confronting MVL is the effective utilization of the multiview data to its fullest potential. As per \cite{zhao2017multi}, the current approaches to MVL can be categorized into three groups: margin-consistency \cite{ chao2016consensus}, co-training \cite{li2006multitraining} and co-regularization \cite{sindhwani2005co, szedmak2007synthesis}. To keep the margins of classifiers operating on different views highly consistent, one can use margin-consistency regularization. An early framework for multiview semi-supervised learning is co-training. It operates through iterative processes aimed at maximizing consensus among classifiers across diverse views, ensuring coherence within the data. Co-regularization involves the amalgamation of regularization terms related to regression or discriminant functions from separate views into the overarching objective function.

The efficacy of MVL algorithms hinges on two fundamental principles: the consensus principle and the complementarity principle. These principles are essential in determining how multiple views are related and combined. The complementarity principle highlights the importance of leveraging the complementary information provided by different views to ensure a comprehensive description of the data. The consensus principle aims to maximize the correlation among multiple distinct perspectives to achieve accurate classifiers for each individual view. In the domain of MVL, these two principles play a pivotal role in guiding the construction of models. Multiview SVM (SVM-2K) \cite{farquhar2005two} is a consensus-based principle. SVM-2K model involves training two SVMs within distinct feature spaces, employing an additional $l_1$-norm constraint during the training process to uphold robustness. Considering the Rademacher complexity, incorporating the $l_1$-norm constraint results in a more rigorous generalization error bound for SVM-2K \cite{farquhar2005two}. The Multiview TSVM (MvTSVM) \cite{xie2015multi} is inspired by the co-regularization concept found in SVM-2K and utilizes the effective training attributes derived from TSVMs. By solving two smaller-sized QPPs, MvTSVM generates four non-parallel hyperplanes (associated with two views) in total, which makes it four times faster than SVM-2K. In recent years, several variants of MvTSVM have been proposed such as multiview twin hypersphere support vector machine (MvTHSVM) \cite{zhu2022fast}, multiview learning model based on nonparallel SVM (MvNPSVM) \cite{tang2018multi}, multiview robust double-sided TSVM (MvRDTSVM) \cite{ye2021multiview}, deep multiview TSVM (DMvTSVM) \cite{xie2023deep}, multiview one-class SVM method with privileged information learning (MOCPIL) \cite{xiao2024privileged}, multiview large margin distribution machine (MVLDM) \cite{hu2024multiview} and many more. However, MvTSVM encounters significant challenges, which are as follows: (1) MvTSVM necessitates four matrix inversion operations during the training process, resulting in avoidable computational overhead. (2) MvTSVM requires the reformulation of optimization problems when the training data is mapped into high-dimensional spaces using non-linear functions. (3) The uniformity of noise in the training data, or its pre-known functional dependency, presents significant obstacles to the effectiveness of the model. 

To surmount the aforementioned challenges, we propose a novel multiview twin parametric margin support vector machine (MvTPMSVM). MvTPMSVM generates a total of four non-parallel hyperplanes (associated with two views). Each hyperplane defines the positive or negative parametric margin, aligning closer to its respective class.
The MvTPMSVM autonomously adjusts the parametric insensitive hyperplanes to assimilate the structural information inherent in the data, effectively capturing the intrinsic characteristics of heteroscedastic noise within the dataset. The optimization problem of MvTPMSVM is to remove the need for computing extensive matrix inverses and employ the kernel method directly. The integration of multiple views in the proposed MvTPMSVM model helps mitigate the impact of missing or noisy data in any single view, thereby enhancing the model’s robustness and reliability.

The main contributions of this work are outlined as follows:
\begin{enumerate}
    \item We propose a novel multiview twin parametric margin support vector machine (MvTPMSVM). The proposed MvTPMSVM generates parametric margin hyperplanes corresponding to both classes, thereby controlling the impact of heteroscedastic noise present in the data.
    \item The MvTPMSVM circumvents the need for computing matrix inverses, unlike other baseline models, which necessitate the calculation of matrix inverse that becomes infeasible for real-world scenarios.
    \item The MvTPMSVM model adeptly captures intricate, nonlinear relationships present within the data. This ability frequently results in enhanced performance, achieved with notably lower computational complexity.
    \item We provide rigorous mathematical frameworks for MvTPMSVM, for both linear and nonlinear kernel spaces. MvTPMSVM can directly apply the kernel method with an elegant mathematical formulation.
    \item We conduct experiments encompassing $55$ UCI and KEEL, synthetic, and Animals with Attributes (AwA) datasets. Through exhaustive numerical experiments and comprehensive statistical analyses, our findings establish the superior performance of the proposed MvTPMSVM model in comparison to the baseline models.
\end{enumerate}

The subsequent sections of the paper are structured as follows: Section \ref{Related Work} introduces notation and an overview of the related work.  Section \ref{Proposed multi-view learning with twin parametric support vector machine} explains the formulation of the proposed MvTPMSVM model. The experimental results and discussions are presented in Section \ref{Experiments}. The conclusion and avenues for future work are presented in Section \ref{Conclusions}.
\section{Related Work}
\label{Related Work}
In this section, we briefly outline the mathematical formulations of SVM-2K and MvTSVM. Consider the sample space denoted as $\mathscr{T}$, which is a product of two distinct feature views, 
$A$ and $B$, expressed as $\mathscr{T} = \mathscr{T}^A \times \mathscr{T}^B$, where, $\mathscr{T}^A \subseteq \mathbb{R}^{m_1}$ $\mathscr{T}^B \subseteq \mathbb{R}^{m_2}$ and the label space is presented by $\mathscr{Y} = \{-1, +1\}$. Suppose $\mathscr{H} = \{(x_i^A, x_i^B, y_i) | x_i^A \in \mathscr{T}^A, x_i^B \in \mathscr{T}^B, y_i \in \mathscr{Y}\}_{i=1}^m$ represent a two-view data set. Let us consider the input matrices $\textbf{A}^A$ and $\textbf{B}^A$ represent the $+1$ and $-1$ class sample of view $A$, and $\textbf{A}^B$ and $\textbf{B}^B$ represent the $+1$ and $-1$ class sample of view $B$, respectively. The non-parallel hyperplanes are given by $h_1^A = <w_1^A, x^A> + b_1^A$ and $h_2^A = <w_2^A, x^A> + b_2^A$ for view $A$, and two non-parallel hyperplanes for view $B$ is given by $h_1^B = <w_1^B, x^B> + b_1^B$ and $h_2^B = <w_2^B, x^B> + b_2^B$. Let $A_1 = [\textbf{A}^A, \hspace{0.1cm} e_1]$, $B_1 = [\textbf{A}^B, \hspace{0.1cm} e_1]$, $A_2 = [\textbf{B}^A, \hspace{0.1cm} e_2]$, $B_2 = [\textbf{B}^B, \hspace{0.1cm} e_2]$, $v_1 = [w_1^A; \hspace{0.1cm} b_1^A]$, $v_2 = [w_1^B; \hspace{0.1cm} b_1^B]$, $u_1 = [w_2^A; \hspace{0.1cm} b_2^A]$, and $u_2 = [w_2^B; \hspace{0.1cm} b_2^B]$, where $e_1 \in \mathbb{R}^{m_1}$, $e_2 \in \mathbb{R}^{m_2}$ are the vector of ones, respectively. Table \ref{List of notations} provides a summary of the primary notations used in this paper.

\begin{table*}[]
    \centering
    \caption{List of notations.}
    \label{List of notations}
    \resizebox{1\textwidth}{!}{
\begin{tabular}{ll}
\hline
Notation & Description                                             \\ \hline
$m_1$, $m_2$   & number of $+1$ and $-1$ training samples, respectively.        \\
$(x_i^A, x_i^B, y_i)$        & $i^{th}$ training sample with its label.                      \\
$<x_i, x_j>$     & dot product of $x_i$ and $x_j$, or termed as $x_i^Tx_j$.         \\
$\phi_A(\cdot), \phi_B(\cdot)$      & transformations from input spaces to high-dimensional feature spaces. \\
$\lvert \cdot \rvert, \|\cdot\|$       & $1$-norm, $2$-norm, respectively.                                          \\
$k_A(x_i^A, x_j^A)$       & kernel function $<\phi_A(x_i^A), \phi_A(x_j^A)>$.                       \\
$k_B(x_i^B, x_j^B)$       & kernel function $<\phi_B(x_i^B), \phi_B(x_j^B)>$.              \\ 
$w_1^A, w_2^A, w_1^B, w_2^B$       & weight vectors corresponding to $+1$ and $-1$ classes of views $A$ and $B$, respectively.              \\ \hline
\end{tabular}}
\end{table*}
\subsection{Two view learning: SVM-2K, theory and practice}
SVM-2K \cite{farquhar2005two} finds two distinct optimal hyperplanes: one associated with view $A$ and another with view $B$, and given as:
\begin{align}
    w^{A^T}\phi_A(x^A) + b^{A} =0,  \hspace{0.4cm}   w^{B^T}\phi_B(x^B) + b^{B} =0.
\end{align}
The optimization problem of SVM-2K can be written as follows:

\begin{align}
\label{eq:1}
\underset{w^{B},~ w^{A},~ b^{B},~ b^{A}}{\min}  \hspace{0.1cm}~&\frac{1}{2}\|w^{A}\|^2+\frac{1}{2}\|w^{B}\|^2+C_{1}\sum_{i=1}^{m} \xi_{i}^A + C_{2}\sum_{i=1}^{m} \xi_{i}^B + D\sum_{i=1}^{m} \eta_{i} \nonumber \\
 \text { s.t. }\hspace{0.1cm}  & \lvert <w^B, \phi_B(x_i^B)> + b^B - <w^A, \phi_A(x_i^A)> - b^A \rvert \leq \epsilon + \eta_i, \nonumber \\
 & y_i(<w^B, \phi_B(x_i^B)> + b^B) \geq 1 - \xi_i^B, \nonumber \\
 & y_i(<w^A, \phi_A(x_i^A)> + b^A) \geq 1 - \xi_i^A, \nonumber \\
 & \xi_{i}^A \geq 0, \xi_{i}^B \geq 0, \eta_{i} \geq 0, ~~ i=1, 2, \ldots m, 
\end{align}
where $C_{1}$, $C_2$, $D$ $(> 0)$ are penalty parameters, $\epsilon > 0$ is an insensitive parameter and $\xi_i^A$, $\xi_i^B$, and $\eta_i$ are slack variables, respectively. The amalgamation of the two views through the  $\epsilon$-insensitive $l_1$-norm similarity constraint is incorporated as the initial constraint in the problem \eqref{eq:1}. The dual problem of \eqref{eq:1} is given by:
\begin{align}
\label{eq:2}
\underset{ \beta_i^{+},~ \beta_i^{-},~ \alpha_i^{A},~ \alpha_i^{B}}{\max}  \hspace{0.05cm}~&-\frac{1}{2} \sum_{i, j=1}^m(g_i^A g_j^Ak_A(x_i, x_j) + g_i^Bg_j^Bk_B(x_i, x_j)) + \sum_{i=1}^m(\alpha_i^A+ \alpha_i^B)\nonumber \\
 \text { s.t. }\hspace{0.1cm}  & g_i^A = \alpha_i^Ay_i - \beta_i^+ + \beta_i^-, \nonumber \\
 & g_i^B = \alpha_i^By_i + \beta_i^+ - \beta_i^-, \nonumber \\
 & \sum_{i=1}^n g_i^A = \sum_{i=1}^n g_i^B = 0, \nonumber \\
 & 0 \leq \beta_i^+, \beta_i^-, \beta_i^+ + \beta_i^- \leq D, \nonumber \\
 & 0 \leq \alpha _i^{A/B} \leq C_{1/2}, 
\end{align}
where $\alpha_i^A, \alpha_i^B, \beta_i^+,$ and $\beta_i^-$ are the vectors of Lagrangian multipliers. The predictive function for each view is expressed as:
\begin{align}
    f_{A/B} = \sum_{i=1}^m g_i^{A/B}k_{A/B}(x_i, x) +b^{A/B}.
\end{align}

\subsection{Multiview twin support vector machine (MvTSVM)}
MvTSVM \cite{xie2015multi} needs to construct four non-parallel hyperplanes. The optimization problem of MvTSVM is given as follows:
\begin{align}
\label{eq:4}
\underset{ v_{1},~ v_{2},~ \xi_1,~ \xi_2,~ \eta_1}{\min}  \hspace{0.1cm}~&\frac{1}{2}\|A_1v_1\|^2+\frac{1}{2}\|A_2v_2\|^2+C_{1}e_2^T\xi_{1} + C_{2}e_2^T\xi_{2} + D_1e_1^T\eta_{1} \nonumber \\
 \text { s.t. }\hspace{0.1cm}  & \lvert A_1v_1 - A_2v_2 \rvert \leq \epsilon + \eta_1, \nonumber \\
 & -B_1v_1 \geq e_2 - \xi_1, \nonumber \\
 & -B_2v_2 \geq e_2 - \xi_2, \nonumber \\
 & \xi_{1} \geq 0, \xi_{2} \geq 0, \eta_{1}\geq 0, 
\end{align}
and
\begin{align}
\label{eq:5}
\underset{ u_{1},~ u_{2},~ \xi_3,~ \xi_4,~ \eta_2}{\min}  \hspace{0.1cm}~&\frac{1}{2}\|B_1u_1\|^2+\frac{1}{2}\|b_2u_2\|^2+C_{3}e_1^T\xi_{3} + C_{4}e_1^T\xi_{4} + D_2e_2^T\eta_{2} \nonumber \\
 \text { s.t. }\hspace{0.1cm}  & \lvert B_1u_1 - B_2u_2 \rvert \leq \epsilon + \eta_2, \nonumber \\
 & A_1u_1 \geq e_1 - \xi_3, \nonumber \\
 & A_2u_2 \geq e_1 - \xi_4, \nonumber \\
 & \xi_{3} \geq 0, \xi_{4} \geq 0, \eta_{2}\geq 0, 
\end{align}
where $C_1$, $C_2$, $C_3$, $C_4$, $D_1$, $D_2$ are tunable parameters, and $\epsilon > 0$ is an insensitive parameter; $\xi_1$, $\xi_2$, $\xi_3$, $\xi_4$, $\eta_1$, and $\eta_2$ are slack variables, respectively.

The dual optimization problem of \eqref{eq:4} and \eqref{eq:5} are given by:
\begin{align}
\label{eq:6}
\underset{\beta_1,~ \beta_2,~ \alpha_1,~ \alpha_2}{\max}  \hspace{0.1cm}~&\frac{1}{2} \zeta_1^T(A_1^TA_1)^{-1} \zeta_1 + \zeta_2^T(A_2^TA_2)^{-1} \zeta_2 - (\alpha_1 + \alpha_2)^Te_2  \nonumber \\
 \text { s.t. }\hspace{0.1cm}  & \zeta_1 = A_1^T(\beta_2 - \beta_1) - B_1^T\alpha_1, \nonumber \\
 & \zeta_2 = A_2^T(\beta_1 - \beta_2) - B_2^T\alpha_2, \nonumber \\
 & 0 \leq \beta_1, \beta_2, \beta_1 + \beta_2 \leq D_1 e_1, \nonumber \\
 & 0 \leq \alpha_{1/2} \leq C_{1/2}e_2,
\end{align}
and
\begin{align}
\label{eq:7}
\underset{\delta_1,~ \delta_2, ~\gamma_1, ~\gamma_2}{\max}  \hspace{0.1cm}~&\frac{1}{2} \rho_1^T(B_1^TB_1)^{-1} \rho_1 + \rho_2^T(B_2^TB_2)^{-1} \rho_2 - (\gamma_1 + \gamma_2)^Te_1  \nonumber \\
 \text { s.t. }\hspace{0.1cm}  & \rho_1 = B_1^T(\delta_2 - \delta_1) - A_1^T\gamma_1, \nonumber \\
 & \rho_2 = B_2^T(\delta_1 - \delta_2) - A_2^T\gamma_2, \nonumber \\
 & 0 \leq \delta_1, \delta_2, \delta_1 + \delta_2 \leq D_2 e_2, \nonumber \\
 & 0 \leq \gamma_{1/2} \leq C_{3/4}e_1,
\end{align}
where $\alpha_1$, $\alpha_2$, $\beta_1$, $\beta_2$, $\gamma_1$, $\gamma_2$, $\delta_1$ and $\delta_2$ are the vectors of Lagrange multipliers. \\
The categorization of a new input data point $x=(x^A, x^B)$ into either the $+1$ or $-1$ class can be determined as follows:
\begin{align}
    f(x) = \lvert h_1^A + h_1^B \rvert - \lvert h_2^A + h_2^B \rvert.
\end{align}
If the function $f(x)$ yields a value less than 0, it will be assigned to the $+1$ class; otherwise, it will be assigned to the $-1$ class.
\section{Proposed Multiview Twin Parametric Margin Support Vector Machine (MvTPMSVM)}
\label{Proposed multi-view learning with twin parametric support vector machine}
\begin{figure}
    \centering
\includegraphics[width=0.9\textwidth,height=4.5cm]{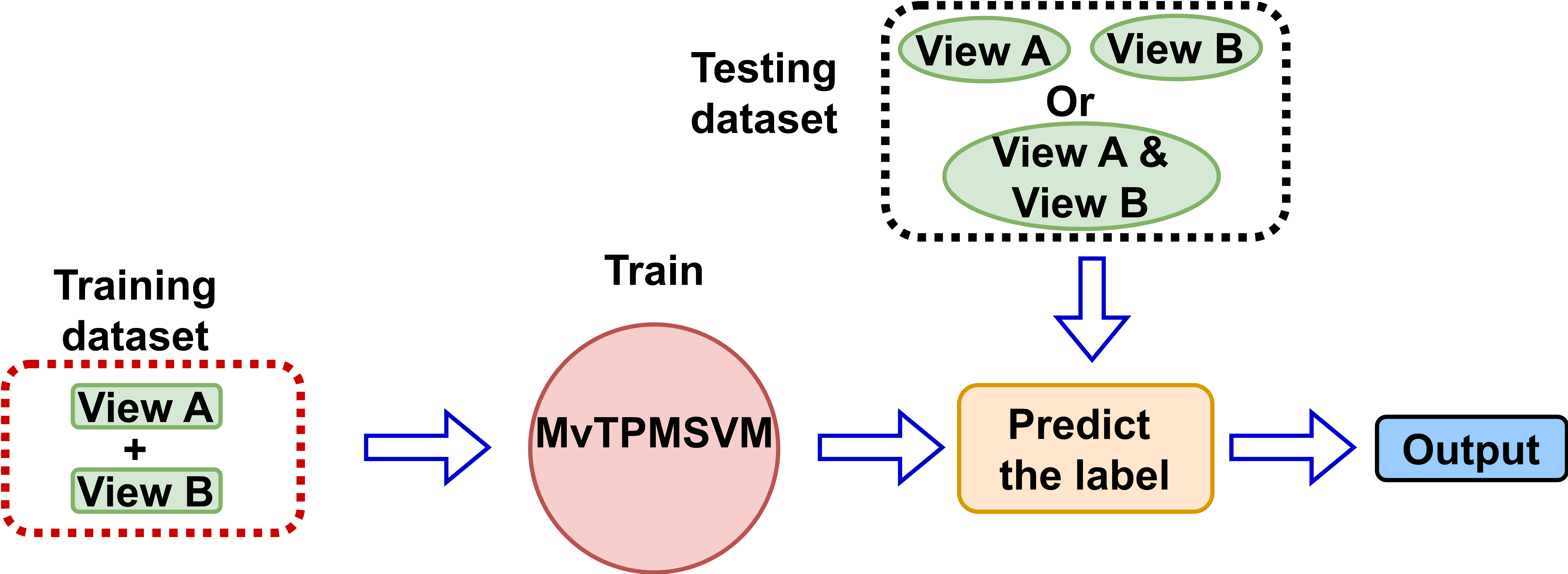}
    \caption{Flow diagram of the proposed MvTPMSVM model.}
    \label{Flow diagrams of the proposed MvTPMSVM model}
\end{figure}

In this section, we provide a detailed mathematical formulation of the proposed MvTPMSVM model tailored for linear and non-linear cases. MvTPMSVM generates a pair of nonparallel parametric-margin hyperplanes solved by two smaller-sized QPPs while eliminating the requirement for calculating large matrix inverses. Also, MvTPMSVM effectively captures more intricate heteroscedastic noise structures, through the utilization of these parametric-margin hyperplanes. Flow diagram of the proposed MvTPMSVM model is shown in Fig. \ref{Flow diagrams of the proposed MvTPMSVM model}.

\subsection{Linear MvTPMSVM}
The optimization problem for linear MvTPMSVM is formulated as:
\begin{align}
\label{eq:9}
\underset{ v_{1},~ v_{2},~ \xi_1,~ \xi_2,~ \eta_1}{\min}  \hspace{0.1cm}~&\frac{1}{2}(\|v_1\|^2+\|v_2\|^2)+C_{1}e_2^T(B_1v_1 + B_2v_2) + C_{2}e_1^T(\xi_{1} + \xi_2) + D_1e_1^T\eta_{1} \nonumber
\end{align}
\begin{align}
\text { s.t. }\hspace{0.1cm}  & \lvert A_1v_1 - A_2v_2 \rvert \leq \eta_1 + \epsilon_1, \nonumber \\
 & A_1v_1 \geq 0 - \xi_1, \nonumber \\
 &  A_2v_2 \geq 0 - \xi_2, \nonumber \\
 &  \xi_{1} \geq 0, \xi_{2} \geq 0, \eta_{1}\geq 0, 
\end{align}
and
\begin{align}
\label{eq:10}
\underset{ u_{1},~ u_{2},~ \xi_3,~\xi_4,~ \eta_2}{\min}  \hspace{0.1cm}~&\frac{1}{2}(\|u_1\|^2+\|u_2\|^2)-C_{3}e_1^T(A_1u_1 + A_2u_2) + C_{4}e_2^T(\xi_{3} + \xi_4) + D_2e_2^T\eta_{2} \nonumber \\
 \text { s.t. }\hspace{0.1cm}  & \lvert B_1v_1 - B_2v_2 \rvert \leq \eta_2 + \epsilon_2, \nonumber \\
 & -B_1u_1 \geq 0 - \xi_3, \nonumber \\
 & -B_2u_2 \geq 0 - \xi_4, \nonumber \\
 & \xi_{3} \geq 0, \xi_{4} \geq 0, \eta_{2}\geq 0, 
\end{align}
where, $C_i \hspace{0.1cm} (i=1,2,3,4)$, $D_j \hspace{0.1cm} (j=1,2)$ are positive tunable parameters, $\xi_i \hspace{0.1cm} ( i=1,2,3,4)$, $\eta_i \hspace{0.1cm} ( j=1,2)$, represents the slack variables.

The demonstration of the MvTPMSVM mechanism is illustrated as follows:
\begin{enumerate}
\item The terms $\|v_1\|$ and $\|v_2\|$ are regularization components for view $A$, while $\|u_1\|$ and $\|u_2\|$ are regularization components for view $B$. These terms are intended to prevent overfitting by constraining the capabilities of the classifier sets for both views.
\item The second term in the objective function minimizes the sum of projection values for the $-1$ class training samples of both views. By leveraging label correlations, it incorporates the $\epsilon$-insensitive loss function, which gives rise to the inclusion of a regularization term between different views. Nonnegative slack variables $\eta_1$ and $\eta_2$ are introduced, along with tunable parameters $D_1$ and $D_2$, to evaluate the degree of deviation from $\epsilon$-similarity between two $+1$ ($-1$) classifiers across two views.
\item The first constraint in the objective function of the proposed MvTPMSVM model is the $\epsilon$-insensitive $l_1$-norm, which ensures the alignment of the classifiers from both views concerning $+1$ and $-1$ classes. Samples deviating from these constraints are accommodated by introducing the parameter $\epsilon$. By leveraging label correlations, this approach diminishes the gap between different perspectives, thereby ensuring their consistency. 
\item The second component of the objective function is made to minimize the total projection values of the $+1$ ($-1$) training points in each view. Optimizing this component encourages the effective positioning of $-1$ ($+1$) training points to attain maximal distance from the $+1$ ($-1$) parametric-margin hyperplane.
\item The second and third constraints in the optimization problem of the proposed MvTPMSVM model stipulate that the parametric-margin hyperplane should not be less than zero. The fault tolerance
parameters $\xi_1$, $\xi_2$, $\xi_3$, and $\xi_4$ are employed to quantify the extent of disagreement on each view. In order to avoid overfitting to the $+1$ ($-1$) training points, the third term of the objective function minimizes the sum of error variables.
\end{enumerate}
The corresponding Lagrangian function of problem \eqref{eq:9} can be formulated as:
\begin{align}
\label{eq:12}
    L=&\frac{1}{2}(\|v_1\|^2+\|v_2\|^2) + C_{1}e_2^T(B_1v_1 + B_2v_2) + C_{2}e_1^T(\xi_{1} + \xi_2) + D_1e_1^T\eta_{1} \nonumber \\
    &- \beta_1^{+^T}(\eta_1 + \epsilon_1 - A_1v_1 + A_2v_2) -\beta_2^{+^T}(A_1v_1-A_2v_2 + \eta_1 + \epsilon_1) - \alpha_1^{+^T}(A_1v_1 + \xi_1)  \nonumber \\
    & -\alpha_2^{+^T}(A_2v_2 + \xi_2) - \lambda_1^{+^T}\xi_1 - \lambda_2^{+^T}\xi_2 - \sigma^{+^T}\eta_1.
\end{align}
Using the Karush-Kuhn-Tucker (K.K.T.) conditions, we have
\begin{align}
   &\frac{\partial L}{\partial v_1} =  v_1 + C_1B_1^Te_2 + A_1^T\beta_1^{+} - A_1^T\beta_2^{+} - A_1^T\alpha_1^{+} = 0, \label{eq:13}\\
    &\frac{\partial L}{\partial v_2} = v_2 + C_1B_2^Te_2 - A_2^T\beta_1^{+} + A_2^T\beta_2^{+} - A_2^T\alpha_2^{+} = 0, \label{eq:14} \\
    & \frac{\partial L}{\partial \xi_1} = C_2e_1 - \alpha_1^{+} - \lambda_1^{+} = 0, \label{eq:15} \\
    & \frac{\partial L}{\partial \xi_2} = C_2e_1 - \alpha_2^{+} - \lambda_2^{+} = 0,\label{eq:16} \\
    & \frac{\partial L}{\partial \eta_1} = D_1e_1 - \beta_1^{+} -\beta_2^{+} - \sigma^{+} =0, \label{eq:17}, \\
      & \frac{\partial L}{\partial \beta_1^+} = \eta_1+\epsilon_1 - A_1v_1 + A_2v_2 = 0,  \\
      &\frac{\partial L}{\partial \beta_2^+} = \eta_1+\epsilon_1 - A_1v_1 + A_2v_2 = 0, \\  
    & \frac{\partial L}{\partial \alpha_1^{+}} = A_1v_1 + \xi_1 = 0, \\
    & \frac{\partial L}{\partial \alpha_2^{+}} = A_2v_2 + \xi_2 = 0. 
\end{align}
From Eqs. \eqref{eq:13} and \eqref{eq:14}, we obtain 
\begin{align}
    v_1 = A_1^T (\alpha_1^{+} - \beta_1^{+} + \beta_2^{+})-C_1B_1^Te_2, \label{eq:22} \\
    v_2 = A_2^T (\alpha_2^{+} + \beta_1^{+} - \beta_2^{+}) - C_1B_2^Te_2. \label{eq:23}
\end{align}
Putting Eqs. \eqref{eq:22} and \eqref{eq:23} in \eqref{eq:12}, we get
\begin{align}
\label{eq:24}
    L&= \frac{1}{2}(\alpha_1^+ - \beta_1^+ +\beta_2^+)^TA_1A_1^T(\alpha_1^+ - \beta_1^+ +\beta_2^+) -\frac{1}{2} C_1 (\alpha_1^+ - \beta_1^+ +\beta_2^+)^TA_1B_1^Te_2 \nonumber \\
    & -\frac{1}{2}C_1e_2^TB_1A_1^T(\alpha_1^+ - \beta_1^ + + \beta_2^+) + \frac{1}{2}C_1^2e_2^TB_1B_1^Te_2 + \frac{1}{2}(\alpha_2^+ + \beta_1^+ -\beta_2^+)^TA_2A_2^T(\alpha_2^+ + \beta_1^+ -\beta_2^+) \nonumber \\
    & -\frac{1}{2} C_1 (\alpha_2^+ + \beta_1^+ -\beta_2^+)^TA_2B_2^Te_2 -\frac{1}{2}C_1e_2^TB_2A_2^T(\alpha_2^+ + \beta_1^+ -\beta_2^+) + \frac{1}{2}C_1^2e_2^TB_2B_2^Te_2 \nonumber \\
    &+C_1e_2^TB_1A_1^T(\alpha_1^+ - \beta_1^+ +\beta_2^+) - C_1^2e_2^TB_1B_1^Te_2 + C_1e_2^TB_2A_2^T(\alpha_2^+ + \beta_1^+ -\beta_2^+) \nonumber \\
    &- C_1^2e_2^TB_2B_2^Te_2 + \beta_1^{+^T}A_1A_1^T(\alpha_1^+ - \beta_1^+ +\beta_2^+) - C_1\beta_1^{+^T}A_1B_1^Te_2 - \beta_1^{+^T}A_2A_2^T(\alpha_2^+ + \beta_1^+ -\beta_2^+) \nonumber \\
    &+ \beta_1^{+^T}C_1A_2B_2^Te_2 - \beta_2^{+^T}A_1A_1^T(\alpha_1^+ - \beta_1^+ +\beta_2^+) + C_1\beta_2^{+^T}A_1B_1^Te_2 + \beta_2^{+^T} A_2A_2^T(\alpha_2^+ + \beta_1^+ -\beta_2^+) \nonumber \\
    & - C_1\beta_2^{+^T}A_2B_2^Te_2 - \alpha_1^{+^T}A_1A_1^T(\alpha_1^+ - \beta_1^+ +\beta_2^+) + C_1 \alpha_1^{+^T}A_1B_1^Te_2 - \alpha_2^{+^T}A_2A_2^T(\alpha_2^+ + \beta_1^+ -\beta_2^+) \nonumber \\
    & + C_1 \alpha_2^TA_2B_2^Te_2  + C_2e_1^T(\xi_1 + \xi_2) + D_1e_1^T\eta_1 -\beta_1^{+T}\eta_1 -\beta_2^{+T}\eta_1 \nonumber \\
    &- \alpha_1^{+^T}\xi_1 - \alpha_2^T \xi_2\xi_2 - \lambda_1^{+^T}\xi_1 - \lambda_2^{+^T}\xi_2 - \sigma^{+^T}\eta_1
 \end{align}
Collecting the terms of Eq. \eqref{eq:24} and using the aforementioned K.K.T. conditions, we obtained the dual form of \eqref{eq:9} as follows:
\begin{align}
\label{eq:25}
    \underset{\alpha_1^+,~ \alpha_2^+,~ \beta_1^+,~ \beta_2^+}{\max} &-\frac{1}{2}(\alpha_1^+ - \beta_1^+ +\beta_2^+)^TA_1A_1^T(\alpha_1^+ - \beta_1^+ +\beta_2^+) \nonumber \\ 
    &- \frac{1}{2}(\alpha_2^+ + \beta_1^+ -\beta_2^+)^TA_2A_2^T(\alpha_2^+ + \beta_1^+ -\beta_2^+) \nonumber \\
    & + C_1e_2^T(B_1A_1^T, B_1A_1^T , B_1A_1^T) (\alpha_1^+ - \beta_1^+ +\beta_2^+) -\beta_1^{+^T}\epsilon_1 \nonumber \\
    & + C_1e_2^T(B_2A_2^T, B_2A_2^T, B_2A_2^T) (\alpha_2^+ + \beta_1^+ -\beta_2^+) -\beta_2^{+^T}\epsilon_1 \nonumber \\
    \text { s.t. }\hspace{0.1cm}  & 0 \leq \alpha_1^+, \leq C_1, \hspace{0.1cm} 0 \leq \alpha_2^+ \leq C_2, \nonumber \\
    & \beta_1^+, \beta_2^+ \geq 0, \hspace{0.1cm} \beta_1^++ \beta_2^+ \leq D_1.
\end{align}
Likewise, the Wolfe dual for \eqref{eq:10} can be obtain as
\begin{align}
\label{eq:26}
    \underset{\alpha_1^-,~ \alpha_2^-,~ \beta_1^-,~ \beta_2^-}{\max} &-\frac{1}{2}(\alpha_1^- - \beta_1^- +\beta_2^-)^TB_1B_1^T(\alpha_1^- - \beta_1^- +\beta_2^-) \nonumber \\
    & - \frac{1}{2}(\alpha_2^- + \beta_1^- -\beta_2^-)^TB_2B_2^T(\alpha_2^- + \beta_1^- -\beta_2^-) \nonumber \\
    & + C_1e_2^T(A_1B_1^T, -A_1B_1^T, -A_1B_1^T) (\alpha_1^- - \beta_1^- +\beta_2^-) - \beta_1^{+^T}\epsilon_2 \nonumber \\
    & + C_1e_2^T(A_2B_2^T, A_2B_2^T, A_2B_2^T) (\alpha_2^- + \beta_1^- -\beta_2^-) - \beta_2^{+^T}\epsilon_2 \nonumber \\
    \text { s.t. }\hspace{0.1cm}  & 0 \leq \alpha_1^-, \leq C_3, \hspace{0.1cm} 0 \leq \alpha_2^- \leq C_4, \nonumber \\
    & \beta_1^-, \beta_2^- \geq 0, \hspace{0.1cm} \beta_1^- + \beta_2^- \leq D_2.
\end{align}

The dual problems of \eqref{eq:25} and \eqref{eq:26} can be concisely formulated as:
\begin{align}
\label{eq:27}
\underset{ \tilde{\tau}}{\max}  \hspace{0.1cm}~&\frac{1}{2} \tilde{\tau}^T \tilde{\Lambda} \tilde{\tau} + \kappa_1 \tilde{\tau} \nonumber \\
 \text { s.t. }\hspace{0.1cm}  & 0 \leq \tilde{\tau} \leq p_1, \nonumber \\
 & H_1\tilde{\tau} \leq D_1e_2, 
\end{align}
where
\begin{align}
    &\tilde{\tau}= (\beta_1^{+^T}, \beta_2^{+^T}, \alpha_1^{+^T}, \alpha_2^{+^T})^T,\label{eq:28}\\
    &p_1 = (0, 0, e_2^T, e_2^T)^T, \label{eq:29}\\
    &H_1 = (E, E, O, O), \label{eq:30}\\
    &\kappa_1 = C_1e_2^T(B_2A_2^T-B_1A_1^T - \epsilon_1I, -B_2A_2^T + B_1A_1^T - \epsilon_1I, B_1A_1^T, B_2A_2^T), 
\end{align}
\begin{align}
    &\tilde{\Lambda} = \begin{bmatrix} 
        F_1 + F_2 & -F_1 - F_2 &  -F_1 &  F_2\\
        -F_1 - F_2 & F_1 + F_2 & F_1 &  -F_2 \\
        -F_1 &  F_1 & F_1 & O\\
         F_2  & -F_2 & O & F_2 \\
    \end{bmatrix}, \\
    & F_1=A_1A_1^T,   \hspace{0.4cm} \text{and} \hspace{0.4cm}  F_2=A_2A_2^T,
\end{align} 
where $\tilde{\tau}$ represents a vector of Lagrangian multipliers. $E$ and $O$ represent an identity matrix and zeros matrix of appropriate dimensions, respectively.
\begin{align}
\label{eq:33}
\underset{ \hat{\tau}}{\max}  \hspace{0.1cm}~&\frac{1}{2} \hat{\tau}^T \hat{\Lambda} \hat{\tau} + \kappa_2 \hat{\tau} \nonumber \\
 \text { s.t. }\hspace{0.1cm}  & 0 \leq \hat{\tau} \leq p_2, \nonumber \\
 & H_2\hat{\tau} \leq D_2e_1, 
\end{align}
where
\begin{align}
    &\hat{\tau}= (\beta_1^{-^T}, \beta_2^{-^T}, \alpha_1^{-^T}, \alpha_2^{-^T})^T, \label{eq:34}\\
    &p_2 = (0, 0, e_1^T, e_1^T)^T, \label{eq:35} \\
    &H_2 = (E, E, O, O), \label{eq:36} \\
      &\kappa_2 = C_3e_1^T(A_1B_1^T+A_2B_2^T - \epsilon_2I, -A_1B_1^T -A_2B_2^T - \epsilon_2I, A_1B_1^T, A_2B_2^T), 
\end{align}
    \begin{align} 
    &\hat{\Lambda} = \begin{bmatrix} 
        F_3 + F_4 & -F_3 - F_4 &  F_3 &  F_4\\
        -F_3 - F_4 & F_3 + F_4 & -F_3 &  -F_4 \\
        F_3 &  -F_3 & F_3 & O\\
         F_4  & -F_4 & O & F_4 \\
    \end{bmatrix}, \\
   & F_3 = B_1B_1^T, \hspace{0.4cm} \text{and} \hspace{0.4cm}  F_4 = B_2B_2^T.
\end{align}
Analogously, $u_1$ and $u_2$ can be calculated by the subsequent equations:
\begin{align}
    u_1 = C_3A_1^Te_1 - B_1^T(\beta_1^{-} - \beta_2^{-} + \alpha_1^{-}), \\
    u_1 = C_3A_2^Te_1 - B_2^T(\beta_1^{-} - \beta_2^{-} + \alpha_2^{-}).
\end{align}
Once the optimal values of $u_1$, $v_1$, $u_2$ and $v_2$ are calculated. Then the categorization of a new input data point $x=(x^A, x^B)$ into either the $+1$ or $-1$ classes can be determined as follows:
\begin{align}
\label{eq:145}
    f(x) = \lvert h_1^A + h_1^B \rvert - \lvert h_2^A + h_2^B \rvert.
\end{align}
If the function $f(x)$ yields a value less than 0, $x$ will be assigned to the $+1$ class; otherwise, $x$ will be assigned to the $-1$ class.
\subsection{Nonlinear MvTPMSVM}
To extend our proposed MvTPMSVM model to the non-linear case, we consider the hyperplanes,
$h_1^{A^{\phi}} = \phi_A(x^A)w_1^A + b_1^A$ and $h_2^{A^{\phi}} = \phi_A(x^A)w_2^A + b_2^A$ for view $A$, and $h_1^{B^{\phi}} = \phi_B(x^B)w_1^B + b_1^B$ and $h_2^{B^{\phi}} = \phi_B(x^B)w_2^B + b_2^B$ for view $B$.
The optimization problem for linear MvTPMSVM is formulated as:
\begin{align}
\label{eq:42}
\underset{ v_{1},~ v_{2},~ \xi_1,~ \xi_2,~ \eta_1}{\min}  \hspace{0.1cm}~&\frac{1}{2}(\|v_1\|^2+\|v_2\|^2)+C_{1}e_2^T(\phi_B(B_1)v_1  + \phi_B(B_2)v_2) + C_{2}e_1^T(\xi_{1} + \xi_2) + D_1e_1^T\eta_{1} \nonumber \\
 \text { s.t. }\hspace{0.1cm}  & \lvert \phi_A(A_1)v_1 - \phi_A(A_2)v_2 \rvert \leq \eta_1+\epsilon_1, \nonumber \\
 & \phi_A(A_1)v_1 \geq 0 - \xi_1, \nonumber \\
 & \phi_A(A_2)v_2 \geq 0 - \xi_2, \nonumber \\
 & \xi_{1} \geq 0, \xi_{2} \geq 0, \eta_{1}\geq 0, 
\end{align}
and
\begin{align}
\label{eq:43}
\underset{ u_{1},~ u_{2},~ \xi_3,~ \xi_4,~ \eta_2}{\min}  \hspace{0.1cm}~&\frac{1}{2}(\|u_1\|^2+\|u_2\|^2)-C_{3}e_1^T(\phi_A(A_1)u_1 + \phi_A(A_2)u_2) + C_{4}e_2^T(\xi_{3} + \xi_4) + D_2e_2^T\eta_{2} \nonumber \\
 \text { s.t. }\hspace{0.1cm}  & \lvert \phi_B(B_1)v_1 - \phi_B(B_2)v_2 \rvert \leq \eta_2+\epsilon_2, \nonumber \\
 & -\phi_B(B_1)u_1 \geq 0 - \xi_3, \nonumber \\
 & -\phi_B(B_2)u_2 \geq 0 - \xi_4, \nonumber \\
 & \xi_{3} \geq 0, \xi_{4} \geq 0, \eta_{2}\geq 0, 
\end{align}
where, $C_i \hspace{0.1cm} (i=1,2,3,4)$, \text{and} $D_j \hspace{0.1cm} (j=1,2)$ are positive tunable parameters. $\xi_i \hspace{0.1cm} ( i=1,2,3,4)$, \text{and} $\eta_i \hspace{0.1cm} ( j=1,2)$, represents the slack variables.

The dual formulations of \eqref{eq:42} and \eqref{eq:43} can be calculated in a similar way as in the linear case, and are given by:
\begin{align}
\label{eq:44}
\underset{ \tilde{\tau}}{\max}  \hspace{0.1cm}~&\frac{1}{2} \tilde{\tau}^T \tilde{\Lambda} \tilde{\tau} + \kappa_1 \tilde{\tau} \nonumber \\
 \text { s.t. }\hspace{0.1cm}  & 0 \leq \tilde{\tau} \leq p_1, \nonumber \\
 & H_1\tilde{\tau} \leq D_1e_2, 
\end{align}
and
\begin{align}
\label{eq:45}
\underset{ \hat{\tau}}{\max}  \hspace{0.1cm}~&\frac{1}{2} \hat{\tau}^T \hat{\Lambda} \hat{\tau} + \kappa_2 \hat{\tau} \nonumber \\
 \text { s.t. }\hspace{0.1cm}  & 0 \leq \hat{\tau} \leq p_2, \nonumber \\
 & H_2\hat{\tau} \leq D_2e_1, 
\end{align}
where
\begin{align}
    &\kappa_1 = C_1e_2^T(k_B(B_2, A_2)-k_A(B_1, A_1) - \epsilon_1 I, \hspace{0.05cm} -k_B(B_2, A_2) + k_A(B_1, A_1) - \epsilon_1 I, \nonumber \\
    & \hspace{1.5cm} k_A(B_1, A_1), \hspace{0.05cm} k_B(B_2, A_2), 
\end{align}
\begin{align}
    &\tilde{\Lambda} = \begin{bmatrix} 
        F_1^\phi + F_2^\phi & -F_1^\phi -F_2^\phi &  -F_1^\phi &  F_2^\phi\\
        -F_1^\phi - F_2^\phi & F_1^\phi + F_2^\phi & F_1^\phi &  -F_2^\phi \\
        -F_1^\phi &  F_1^\phi & F_1^\phi & O\\
         F_2^\phi  & -F_2^\phi & O & F_2^\phi \\
    \end{bmatrix}, \\
    & F_1^\phi = k_A(A_1, A_1), \hspace{0.4cm}  F_2^\phi = k_B(A_2, A_2), \\
    &\kappa_2 = C_3e_1^T(k_A(A_1, B_1)+k_B(A_2, B_2) - \epsilon_2 I, -k_A(A_1, B_1)  -k_B(A_2, B_2) - \epsilon_2 I, \nonumber \\
    & \hspace{1.4cm} k_A(A_1, B_1), k_B(A_2, B_2), \\
    &\hat{\Lambda} = \begin{bmatrix} 
        F_3^\phi + F_4^\phi & -F_3^\phi - F_4^\phi &  F_3^\phi &  F_4^\phi\\
        -F_3^\phi - F_4^\phi & F_3^\phi + F_4^\phi & -F_3^\phi &  -F_4^\phi \\
        F_3^\phi &  -F_3^\phi & F_3^\phi & O\\
         F_4^\phi  & -F_4^\phi & O & F_4^\phi \\
    \end{bmatrix} \\
    &  F_3^\phi = k_A(B_1, B_1), \hspace{0.4cm} \text{and} \hspace{0.4cm}   F_4^\phi = k_B(B_2, B_2).
\end{align}

Once the optimal values of $v_1$, $v_2$, $u_1$ and $u_2$ are calculated. The class of a new input data point $x=(x^A, x^B)$ can be determined as follows:

\begin{align}
    f(x) = \lvert h_1^{A^{\phi}} + h_1^{B^{\phi}} \rvert - \lvert h_2^{A^{\phi}} + h_2^{B^{\phi}} \rvert.
\end{align}
If the function $f(x)$ yields a value less than 0, $x$ will be assigned to the $+1$ class; otherwise, $x$ will be assigned to the $-1$ class.

\subsection{Computational complexity and algorithm}
In this section, we briefly discuss the computational complexity of the proposed MvTPMSVM model. Suppose the number of samples of each class are equal, namely $m_1 = m_2 = m/2$, where $m$ represents the number of training samples. The computational complexity of SVM-2K \cite{farquhar2005two} and MvTSVM \cite{xie2015multi} are $\mathcal{O}((4m)^3)$ and $2\times\mathcal{O}((2m)^3)$, respectively. Since our proposed MvTPMSVM model solves two smaller-sized QPP, which are roughly of size $m/2$, the computational complexity of the dual problems of \eqref{eq:27} and \eqref{eq:33} are both $ \ll \mathcal{O}((2m)^3)$. Therefore the computational complexity of the proposed model MvTPMSVM is $ \ll 2\times\mathcal{O}((2m)^3)$. Thus, the computational complexity of the proposed MvTPMSVM model is much lower than that of the SVM-2K and MvTSVM models. The algorithm of the proposed MvTPMSVM model is briefly described in Algorithm \ref{MvTPMSVM classifier}.

\begin{algorithm}
\caption{MvTPMSVM classifier}
\label{MvTPMSVM classifier}
\textbf{Input:} $A_1$ ($B_1$) and $A_2$ ($B_2$) are the matrices of $+1$ ($-1$) class corresponding to view A and view B.\\
\textbf{Output:} Decision function as in \eqref{eq:145}. \\
\vspace{-0.5cm}
\begin{algorithmic}[1]
\STATE Select two suitable kernels $k_A(x_{i}^A, x_{j}^A)$ and $k_B(x_{i}^B, x_{j}^B)$, and initialize the kernel parameters accordingly. 
\STATE Establish and solve QPPs for \eqref{eq:27} and \eqref{eq:33} through the implementation of $5$-fold cross-validation, and subsequently determine the optimal parameters.
\STATE Testing sample is classified into class $+1$ or $-1$ using \eqref{eq:145}.
\end{algorithmic}
\end{algorithm}

\section{Experimental Results}
\label{Experiments}
To test the efficiency of the proposed MvTPMSVM model, we conduct experiment on publicly available benchmark datasets including, $3$ synthetic multiview datasets \cite{wang2023safe}, $55$ real-world UCI \cite{dua2017uci} and KEEL \cite{derrac2015keel} datasets and $45$ binary classification datasets obtained from Animal with Attributes (AwA) \cite{tang2017multiview}.

\begin{figure}
\begin{minipage}{.30\linewidth}
\centering
\subfloat[Double Vortex]{\includegraphics[scale=0.14]{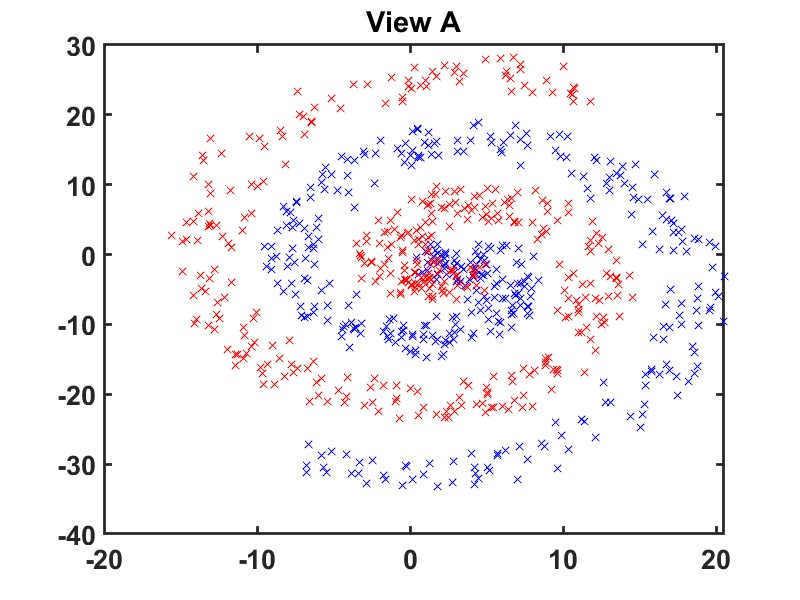}}
\end{minipage}
\begin{minipage}{.30\linewidth}
\centering
\subfloat[Check Board]{\includegraphics[scale=0.14]{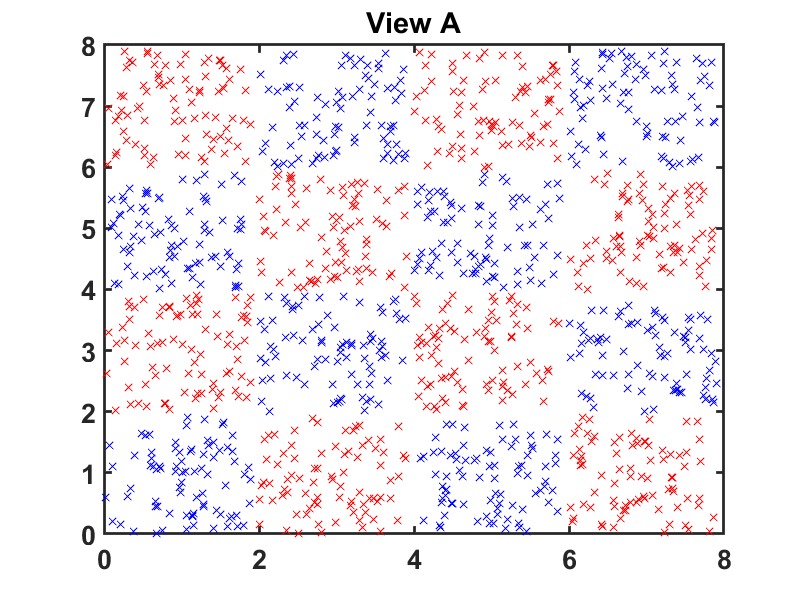}}
\end{minipage}
\begin{minipage}{.30\linewidth}
\centering
\subfloat[Double Moon]{\includegraphics[scale=0.14]{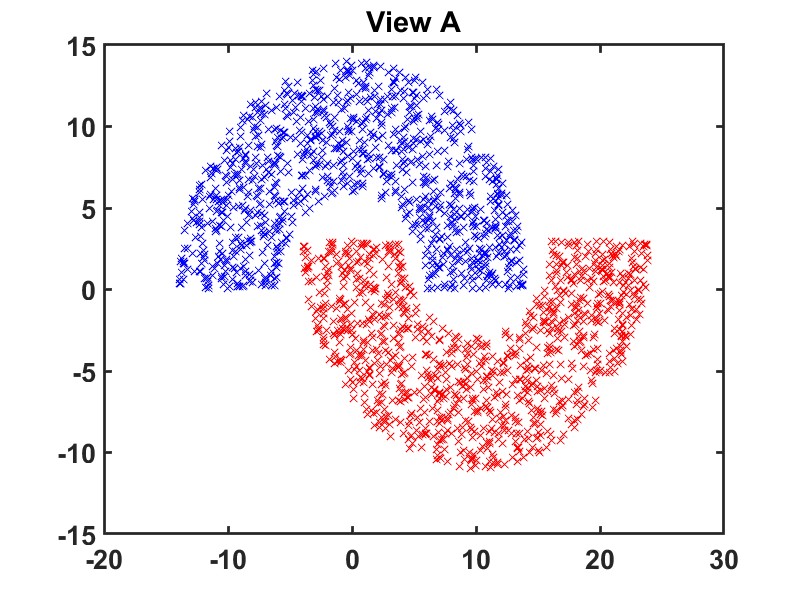}}
\end{minipage}
\par\medskip
\begin{minipage}{.30\linewidth}
\centering
\subfloat[Concentric Circle]{\includegraphics[scale=0.14]{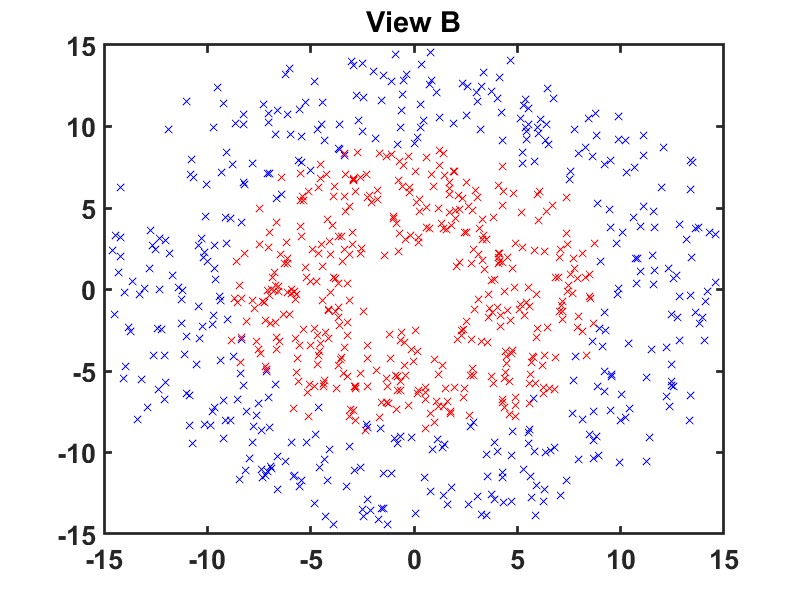}}
\end{minipage}
\begin{minipage}{.30\linewidth}
\centering
\subfloat[Gaussian Cloud]{\includegraphics[scale=0.14]{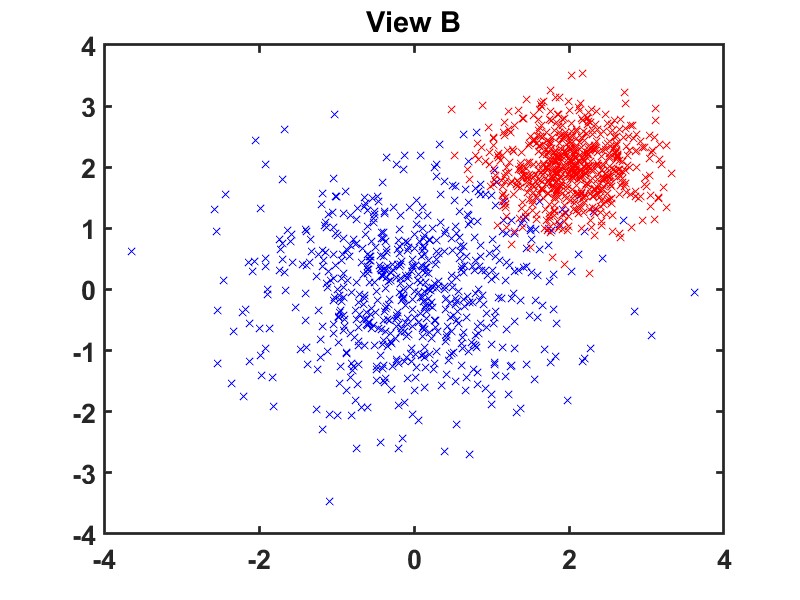}}
\end{minipage}
\begin{minipage}{.30\linewidth}
\centering
\subfloat[Double Line]{\includegraphics[scale=0.14]{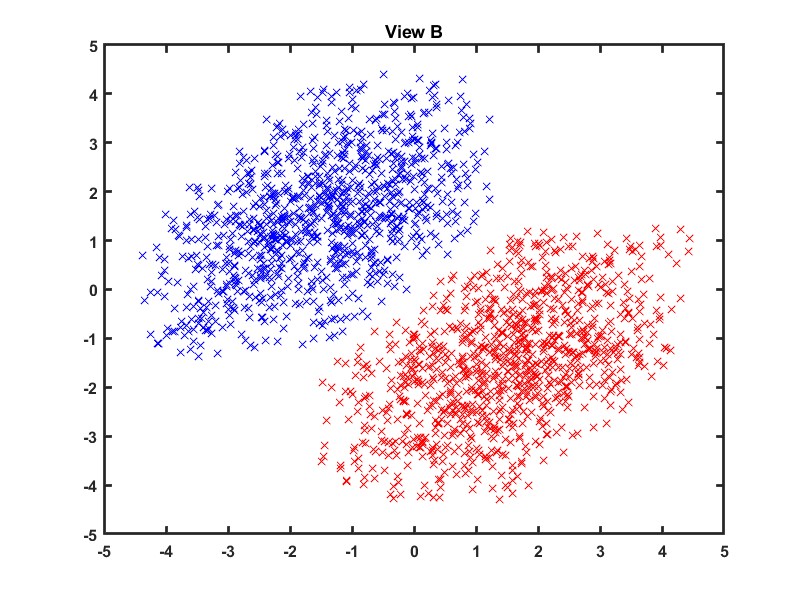}}
\end{minipage}
\caption{
Illustration of three synthetic multiview datasets.}
\label{gl}
\end{figure}

\subsection{Experimental setup}
The experimental hardware setup comprises a PC with an Intel(R) Xeon(R) Gold 6226R CPU running at $2.90$GHz and equipped with $128$ GB of RAM, running Windows $11$ operating system possessing Matlab R2023a. 
The dataset is randomly partitioned into a ratio of $70:30$, allocating $70\%$ of the data for training and $30\%$ for testing. We employ a five-fold cross-validation technique and a grid search approach to optimize the hyperparameters of the models. For all experiments, we opt the Gaussian kernel function represented by $k(x_i, x_j) = e^{-\frac{\|x_i - x_j\|}{2\sigma^2}}$ for each model. The kernel parameter $\sigma$ is selected from the following range: $\{2^{-5}, 2^{-4}, \ldots 2^{4}, 2^{5}\}$. We adopt equal penalty parameters, i.e., $C_1=C_3$ and $C_2=C_4=D_1=D_2$, to mitigate computational costs and are selected from the range $\{2^{-5}, 2^{-4}, \ldots 2^{4}, 2^{5}\}$. For the baseline MvTSVM and MVNPSVM model, we set $C_1 = C_2 = C_3= C_4$ and selected from the range $\{2^{-5}, 2^{-4}, \ldots 2^{4}, 2^{5}\}$. In SVM-2K and PSVM-2V, we set $C_1 = C_1 = D$ and selected from the range $\{2^{-5}, 2^{-4}, \ldots 2^{4}, 2^{5}\}$. For MVLDM, the parameter $C_1, v_1, v_2, \theta, \sigma$ are chosen from $\{ 2^{-5}, 2^{-4}, \ldots, 2^4, 2^5\}$. The parameter $\epsilon$ in the proposed MvTPMSVM model along with the baseline models is set to $0.1$. The generalization performance of the proposed MvTPMSVM model has been evaluated by comparing it with baseline models across various metrics including $accuracy$, $sensitivity$, $precision$, $specificity$, and $error$ $rate$. Mathematically,
\begin{align}
    Accuracy~(\text{ACC}) = \frac{\mathcal{TN} + \mathcal{TP}}{\mathcal{FP}+\mathcal{FN} + \mathcal{TP}+\mathcal{TN}},
\end{align}
 \begin{align}
     Sensitivity ~(\text{Seny})= \frac{\mathcal{TP}}{\mathcal{FN} + \mathcal{TP}},
 \end{align}
 \begin{align}
     Precision ~(\text{Pren})= \frac{\mathcal{TP}}{\mathcal{FP} + \mathcal{TP}},
 \end{align}
\begin{align}
    Specificity ~(\text{Spey}) = \frac{\mathcal{TN}}{\mathcal{TN} + \mathcal{FP}},
\end{align}
\begin{align}
    Error ~ rate = \frac{\mathcal{FP} + \mathcal{FN}}{\mathcal{TP} + \mathcal{TN} + \mathcal{FP} + \mathcal{FN}}
\end{align}
where true positive ($\mathcal{TP}$) represents the count of patterns belonging to $+1$ class that are accurately classified, while false negative ($\mathcal{FN}$) signifies the count of patterns belonging to $-1$ class that are inaccurately classified, false positive ($\mathcal{FP}$) denotes the count of patterns belonging to $-1$ class that are inaccurately classified, and true negative ($\mathcal{TN}$) describes the number of data points of $-1$ class that are correctly classified.

\subsection{Experiments on synthetic datasets} 
The artificial datasets \cite{wang2023safe} exhibit three distinct point distributions with two views. These distributions are categorized as follows: concentric circle and double vortex (synthetic 1), Gaussian cloud and checkboard (synthetic 2), and double square and double moon (synthetic 3). The distributions of these datasets are illustrated in Figure \ref{gl}. The three synthetic datasets consist of 800, 1200, and 2000 samples, respectively.
\begin{table*}[ht!]
\centering
    \caption{Performance comparison of the proposed MvTPMSVM along with the baseline models using non-linear kernel for synthetic datasets.}
    \label{Classification performance of synthetic datasets in nonLinear Case.}
    \resizebox{1.0\textwidth}{!}{
\begin{tabular}{lcccccc}
\hline
\text{Model} $\rightarrow$ & SVM-2K \cite{farquhar2005two} & MvTSVM \cite{xie2015multi} & MVNPSVM \cite{tang2018multi}& PSVM-2V \cite{tang2017multiview} & MVLDM \cite{hu2024multiview} & MvTPMSVM$^{\dagger}$ \\  \hline
Dataset $\downarrow$ &  ACC $(\%)$ &  ACC $(\%)$ & ACC $(\%)$ & ACC $(\%)$ & ACC $(\%)$ & ACC $(\%)$  \\ \hline
synthetic 1 & 93.58 & 89.82 & 85.83 & 93.75 & 93.45& 95.42 \\ 
synthetic 2 & 97.7 & 94.42 & 91.41 & 97.18 & 95.89 & 97.92 \\
synthetic 3 & 100 & 100 & 92.67 & 95.67 & 100 & 100 \\ \hline
Average ACC & \underline{97.09} & 94.75  &  89.97   &  95.53  &  96.44  &  \textbf{97.78} \\ \hline
\multicolumn{7}{l}{$^{\dagger}$ represents the proposed model.}\\
 \multicolumn{7}{l}{The boldface and underline indicate the best and second-best models, respectively, in terms of ACC.}
\end{tabular}}
\end{table*}

Table \ref{Classification performance of synthetic datasets in nonLinear Case.} displays the comparative experimental outcomes of the proposed MvTPMSVM model, along with the baseline models encompassing non-linear case. From Table \ref{Classification performance of synthetic datasets in nonLinear Case.}, the average ACC of the proposed MvTPMSVM model is $97.78\%$, surpassing the performance of the baseline models. It is evident that the proposed MvTPMSVM model consistently outperforms the baseline models. Thus, the experimental comparison of datasets demonstrates the superior performance of our proposed MvTPMSVM model.

\subsection{Experiments on real-world UCI and KEEL datasets} 
In this subsection, we provide a detailed analysis that includes a comparison of the proposed MvTPMSVM model with SVM-2K \cite{farquhar2005two}, MvTSVM \cite{xie2015multi}, MVNPSVM \cite{tang2018multi}, MVLDM \cite{hu2024multiview} and PSVM-2V \cite{tang2017multiview} models across $55$ UCI \cite{dua2017uci} and KEEL \cite{derrac2015keel} benchmark datasets. Our investigation spans a spectrum of scenarios, specifically focusing on non-linear cases, and is subjected to meticulous statistical analysis. Given that the UCI and KEEL datasets lack inherent multiview characteristics, we designate the $95\%$ principal component that we extracted from the original data as view $B$, and we refer to the original data as view $A$. \cite{wang2023safe}.

\begin{table}[htp]
\centering
    \caption{Performance evaluation of the proposed MvTPMSVM model along with the baseline models on UCI and KEEL datasets with the non-linear kernel.}
    \label{Classification performance of UCI in nonLinear Case.}
    \resizebox{1\textwidth}{!}{
\begin{tabular}{lcccccc}
\hline
 \text{Model} $\rightarrow$ & SVM-2K \cite{farquhar2005two} & MvTSVM \cite{xie2015multi} & PSVM-2V \cite{tang2017multiview}  & MVNPSVM \cite{tang2018multi}  & MVLDM \cite{hu2024multiview}  & MvTPSVM$^{\dagger}$ \\ \hline
Dataset $\downarrow$  & (ACC, Seny) & (ACC, Seny) & (ACC, Seny) & (ACC, Seny) & (ACC, Seny) & (ACC, Seny)  \\
 & (Spey, Pren) & (Spey, Pren) & (Spey, Pren) & (Spey, Pren) & (Spey, Pren) & (Spey, Pren) \\ \hline
abalone9-18 & $(93.52, 75.38)$ & $(92.31, 71.03)$ & $(93.61, 78.18)$ & $(92.89, 78.7)$ & $(93.15, 71.76)$ & $(94.06, 78.54)$ \\
 & $(86.67, 75)$ & $(81.82, 75)$ & $(87.25, 70.77)$ & $(88.48, 77.44)$ & $(89.78, 71.05)$ & $(90.45, 78.98)$ \\
acute\_inflammation & $(100, 100)$ & $(90.48, 100)$ & $(100, 100)$ & $(100, 97.76)$ & $(100, 100)$ & $(100, 100)$ \\
 & $(100, 100)$ & $(82.98, 90.7)$ & $(100, 100)$ & $(98.57, 99.24)$ & $(100, 100)$ & $(100, 100)$ \\
acute\_nephritis & $(100, 100)$ & $(100, 100)$ & $(100, 100)$ & $(100, 96.38)$ & $(100, 100)$ & $(100, 100)$ \\
 & $(100, 100)$ & $(100, 100)$ & $(100, 100)$ & $(98.26, 98.36)$ & $(100, 100)$ & $(100, 100)$ \\
aus & $(81.16, 80.9)$ & $(82.61, 77.06)$ & $(83.06, 80.24)$ & $(80.34, 73.68)$ & $(87.92, 84.78)$ & $(83.57, 81.59)$ \\
 & $(76.6, 78.69)$ & $(84.85, 80.77)$ & $(83.15, 80.55)$ & $(86.27, 80)$ & $(87.64, 83.19)$ & $(87.5, 85.75)$ \\
Balance & $(98.84, 88.78)$ & $(97.28, 88.53)$ & $(97.09, 87.74)$ & $(97.09, 87.2)$ & $(75, 80.19)$ & $(95.35, 89.44)$ \\
 & $(98.78, 98.78)$ & $(95.71, 97.1)$ & $(92.5, 97.06)$ & $(90.24, 97.16)$ & $(72.73, 92.67)$ & $(96.59, 95.51)$ \\
bank & $(88.86, 58.16)$ & $(87.74, 52.22)$ & $(84.45, 59.05)$ & $(85.67, 60)$ & $(70.94, 65.69)$ & $(89.31, 66.07)$ \\
 & $(65.91, 88.33)$ & $(67.95, 85.37)$ & $(66.67, 82.22)$ & $(62.76, 84.29)$ & $(66.67, 85.4)$ & $(67.69, 87.14)$ \\
Bankruptcy & $(98.67, 76.97)$ & $(99.43, 78.59)$ & $(98.67, 78.51)$ & $(74.67, 83.33)$ & $(98.67, 89.19)$ & $(100, 100)$ \\
 & $(100, 98.46)$ & $(100, 99.29)$ & $(97.06, 97.78)$ & $(100, 100)$ & $(86.09, 87.61)$ & $(100, 100)$ \\
blood & $(78.57, 81.37)$ & $(75.95, 83.94)$ & $(78.13, 83.33)$ & $(79.56, 86.67)$ & $(80.8, 80.19)$ & $(81.7, 84.65)$ \\
 & $(95.17, 65.78)$ & $(95.45, 67.25)$ & $(96.55, 64.92)$ & $(100, 67.84)$ & $(92.73, 62.67)$ & $(100, 68.89)$ \\
breast\_cancer & $(74.12, 75)$ & $(71.64, 76.56)$ & $(72.35, 76.36)$ & $(76.67, 62.5)$ & $(81.18, 72.86)$ & $(78.82, 74.73)$ \\
 & $(80.75, 75.29)$ & $(81.47, 72.31)$ & $(86.67, 77.06)$ & $(88.55, 73.33)$ & $(89.23, 72.94)$ & $(89.33, 75.71)$ \\
breast\_cancer\_wisc & $(97.61, 97.53)$ & $(95.31, 92.5)$ & $(95.69, 97.58)$ & $(94.35, 66.67)$ & $(97.61, 97.06)$ & $(95.69, 97.88)$ \\
 & $(76.34, 96.93)$ & $(71.93, 92.21)$ & $(74.53, 96.03)$ & $(75.47, 100)$ & $(75.65, 96.35)$ & $(78.31, 100)$ \\
breast\_cancer\_wisc\_diag & $(96.24, 93.15)$ & $(92.73, 82.5)$ & $(92.06, 90)$ & $(92.35, 85)$ & $(97.06, 93.94)$ & $(96.47, 94.33)$ \\
 & $(98.04, 97.09)$ & $(100, 90.41)$ & $(94.05, 91.95)$ & $(100, 95.78)$ & $(98.41, 96.12)$ & $(96.55, 97.92)$ \\
breast\_cancer\_wisc\_prog & $(71.36, 60)$ & $(71.38, 63.51)$ & $(67.97, 70.91)$ & $(71.02, 70)$ & $(81.36, 68.33)$ & $(72.88, 71.18)$ \\
 & $(84.44, 82.11)$ & $(82.5, 82.22)$ & $(81.08, 85.71)$ & $(82.74, 87.24)$ & $(85.47, 85.38)$ & $(86.92, 87.89)$ \\
brwisconsin & $(97.04, 78.45)$ & $(95.82, 76.19)$ & $(96.08, 75)$ & $(83.82, 77.44)$ & $(95.59, 77.06)$ & $(96.57, 78.41)$ \\
 & $(98.45, 98.45)$ & $(97.43, 96.81)$ & $(98.24, 91.08)$ & $(92.86, 90.48)$ & $(96.35, 96.7)$ & $(96.72, 97.25)$ \\
bupa or liver-disorders & $(71.67, 72.22)$ & $(69.42, 75)$ & $(71.84, 79.41)$ & $(66.99, 75.68)$ & $(67.96, 72.08)$ & $(72.34, 72.86)$ \\
 & $(70.91, 85.71)$ & $(72.75, 81.15)$ & $(76.2, 82.68)$ & $(77.67, 85.68)$ & $(72.08, 82.08)$ & $(78.29, 86.43)$ \\
checkerboard\_Data & $(81.16, 77.17)$ & $(79.3, 75.35)$ & $(85.44, 79.41)$ & $(85.44, 78.37)$ & $(85.51, 88.35)$ & $(86.47, 80.22)$ \\
 & $(79.78, 78.45)$ & $(81, 78.07)$ & $(82.2, 72.68)$ & $(66, 74.29)$ & $(83.49, 75.85)$ & $(87.95, 79.91)$ \\
chess\_krvkp & $(99.16, 98.99)$ & $(97.5, 99.06)$ & $(90, 75.93)$ & $(86.74, 97.1)$ & $(97.6, 97.6)$ & $(94.99, 94.85)$ \\
 & $(99.39, 99.19)$ & $(96.76, 97.9)$ & $(78.85, 97.36)$ & $(83.33, 91.78)$ & $(98, 97.8)$ & $(95.61, 99.23)$ \\
cleve.mat & $(87.64, 69.55)$ & $(81.73, 86.02)$ & $(80.9, 75.93)$ & $(74.16, 83.75)$ & $(82.02, 76.92)$ & $(77.53, 79.09)$ \\
 & $(90.59, 86.42)$ & $(84.77, 80)$ & $(88.85, 87.36)$ & $(90.24, 83.75)$ & $(85.71, 81.08)$ & $(92.86, 82.22)$ \\
cmc.mat & $(68.93, 86.07)$ & $(69.38, 75)$ & $(68.93, 81.3)$ & $(77.97, 84.41)$ & $(68.48, 87.55)$ & $(64.41, 87.62)$ \\
 & $(67.09, 75.4)$ & $(71.54, 73.23)$ & $(78.82, 74.92)$ & $(79.17, 77.44)$ & $(73.53, 79.93)$ & $(75.74, 76.67)$ \\
congressional\_voting & $(61.54, 82.74)$ & $(55.41, 76.95)$ & $(59.23, 81.3)$ & $(56.15, 78.51)$ & $(62.31, 82.04)$ & $(62.31, 82.84)$ \\
 & $(75.28, 73.79)$ & $(74.89, 73.74)$ & $(78.82, 74.92)$ & $(74, 74.62)$ & $(78.42, 73.92)$ & $(78.68, 75.78)$ \\
conn\_bench\_sonar\_mines\_rocks & $(88.71, 78.57)$ & $(71.23, 77.83)$ & $(90.32, 73.65)$ & $(64.52, 70.73)$ & $(82.26, 78.56)$ & $(82.26, 79.17)$ \\
 & $(85.65, 86.27)$ & $(82.5, 80.55)$ & $(89.39, 81.47)$ & $(81.76, 88.24)$ & $(82.76, 81.36)$ & $(86, 87.55)$ \\
credit\_approval & $(85.99, 81.97)$ & $(76.4, 87.36)$ & $(85.51, 83.65)$ & $(83.09, 80.89)$ & $(86.47, 82.19)$ & $(85.51, 84.48)$ \\
 & $(83.46, 82.34)$ & $(87.29, 82.01)$ & $(89.39, 81.47)$ & $(86.67, 85.89)$ & $(86.09, 87.61)$ & $(89.09, 87.73)$ \\
cylinder\_bands & $(73.2, 76.14)$ & $(72.7, 73.48)$ & $(73.2, 75.28)$ & $(62.75, 76.83)$ & $(77.78, 73.7)$ & $(75.16, 77.32)$ \\
 & $(77.01, 76.57)$ & $(76.02, 79.57)$ & $(78.98, 80.43)$ & $(78.82, 81.71)$ & $(73.33, 78.17)$ & $(77.89, 79.57)$ \\
echocardiogram & $(79.49, 61.54)$ & $(78.26, 66.67)$ & $(74.36, 64.85)$ & $(76.92, 68.28)$ & $(84.62, 69.23)$ & $(82.05, 69.36)$ \\
 & $(72.73, 66.67)$ & $(70, 61)$ & $(70.84, 63.77)$ & $(73.56, 69.57)$ & $(75.49, 71.82)$ & $(75.87, 73.33)$ \\
fertility & $(86.67, 84.78)$ & $(85.57, 86.45)$ & $(73.33, 87.38)$ & $(86.67, 82.07)$ & $(90, 85.23)$ & $(86.67, 88.79)$ \\
 & $(88.97, 82.36)$ & $(84.56, 84.25)$ & $(85.71, 86.54)$ & $(81.22, 87.45)$ & $(85.27, 80.47)$ & $(89.74, 88.57)$ \\
German & $(74, 89.27)$ & $(69.57, 87.22)$ & $(72, 87.38)$ & $(73.67, 90.91)$ & $(76.67, 87.39)$ & $(74.67, 90.79)$ \\
 & $(76.57, 82.43)$ & $(74.91, 80.6)$ & $(75.71, 86.54)$ & $(79.17, 85)$ & $(76.67, 87.59)$ & $(75.93, 88.28)$ \\
Haberman & $(68.13, 89.28)$ & $(70.23, 89.81)$ & $(69.23, 83.66)$ & $(67.03, 88.16)$ & $(76.92, 80)$ & $(69.64, 90.12)$ \\
 & $(68.13, 81.05)$ & $(74.6, 81.5)$ & $(73.66, 83.66)$ & $(78.31, 85.71)$ & $(75, 87.42)$ & $(82, 87.69)$ \\
heart\_hungarian & $(82.95, 89.13)$ & $(79.13, 86.9)$ & $(82.95, 83.66)$ & $(84.09, 84.74)$ & $(84.09, 87.14)$ & $(85.27, 88.05)$ \\
 & $(81.75, 73.9)$ & $(83.33, 71.43)$ & $(83.66, 73.66)$ & $(86.74, 72.31)$ & $(81.82, 79.41)$ & $(86.96, 76.67)$ \\
hepatitis & $(84.78, 97.14)$ & $(80.73, 100)$ & $(84.78, 95.56)$ & $(78.26, 100)$ & $(82.61, 100)$ & $(91.3, 100)$ \\
 & $(85, 90.67)$ & $(80.73, 89.34)$ & $(83.48, 94.51)$ & $(82.47, 92.78)$ & $(81.4, 89.74)$ & $(85.89, 94.44)$ \\
horse\_colic & $(80.82, 65.12)$ & $(74.42, 68.82)$ & $(80.73, 65.56)$ & $(73.64, 68.75)$ & $(81.82, 60.53)$ & $(80.91, 68.5)$ \\
 & $(84.85, 73.68)$ & $(89.57, 69.19)$ & $(83.48, 74.51)$ & $(87.98, 70.49)$ & $(82.14, 69.7)$ & $(85.65, 70.42)$ \\
 \hline
 \multicolumn{7}{l}{$^{\dagger}$ represents the proposed model.}\\
 \multicolumn{7}{l}{Seny, Spey, and Pren denote the Sensitivity, Specificity, and Precision, respectively.}
\end{tabular}}
\end{table}

 \begin{table}[htbp]
\ContinuedFloat
\centering
    \caption{(Continued)}
  \resizebox{1\textwidth}{!}{                                                     
    \begin{tabular}{lcccccc}
\hline
 \text{Model} $\rightarrow$ & SVM-2K \cite{farquhar2005two} & MvTSVM \cite{xie2015multi} & PSVM-2V \cite{tang2017multiview}  & MVNPSVM \cite{tang2018multi}  & MVLDM \cite{hu2024multiview}  & MvTPSVM$^{\dagger}$ \\ \hline
Dataset $\downarrow$ & (ACC, Seny) & (ACC, Seny) & (ACC, Seny) & (ACC, Seny) & (ACC, Seny) & (ACC, Seny)  \\
 & (Spey, Pren) & (Spey, Pren) & (Spey, Pren) & (Spey, Pren) & (Spey, Pren) & (Spey, Pren) \\ \hline
 Ilpd & $(82.93, 92.74)$ & $(80.07, 90.17)$ & $(82.53, 92.5)$ & $(80.48, 90)$ & $(72.99, 94.44)$ & $(83.33, 95.83)$ \\
 & $(80.55, 71.63)$ & $(85, 75.71)$ & $(82.61, 70.32)$ & $(89.77, 70.89)$ & $(83.33, 72.84)$ & $(85.83, 78.26)$ \\
 Iris & $(100, 100)$ & $(100, 100)$ & $(100, 97.5)$ & $(100, 98.7)$ & $(100, 97.14)$ & $(100, 100)$ \\
 & $(100, 100)$ & $(100, 100)$ & $(82.61, 90.32)$ & $(98.06, 88.16)$ & $(87.18, 91.89)$ & $(100, 100)$ \\
 mammographic & $(80.29, 71.75)$ & $(80.87, 74.92)$ & $(80.9, 74.15)$ & $(75.69, 77.76)$ & $(80.9, 71.48)$ & $(81.25, 79.22)$ \\
 & $(78.63, 80.16)$ & $(74.45, 79.4)$ & $(70.77, 92.43)$ & $(78.5, 91.78)$ & $(78.57, 80)$ & $(80.9, 77.87)$ \\
molec\_biol\_promoter & $(83.87, 71.43)$ & $(72, 74.1)$ & $(67.74, 75.82)$ & $(74.52, 76.38)$ & $(83.87, 78.26)$ & $(80.65, 80.44)$ \\
 & $(80.91, 80)$ & $(80.65, 81.43)$ & $(82.47, 82.45)$ & $(87.36, 84.74)$ & $(83.68, 84.85)$ & $(87.27, 85)$ \\
monks\_1 & $(80.18, 75.6)$ & $(80.38, 79.14)$ & $(80.77, 75.09)$ & $(75.66, 73.68)$ & $(84.94, 78.57)$ & $(81.33, 79.27)$ \\
 & $(85.6, 85.6)$ & $(88.62, 83.62)$ & $(82.11, 82.22)$ & $(76.71, 87.35)$ & $(80.41, 84.08)$ & $(82.28, 87.75)$ \\
monks\_2 & $(77.33, 74.29)$ & $(73.16, 77.21)$ & $(75, 75)$ & $(71.11, 87.2)$ & $(75.56, 75.31)$ & $(78.89, 86.67)$ \\
 & $(81.25, 77.61)$ & $(78.52, 78.95)$ & $(78.24, 71.08)$ & $(80.08, 78.75)$ & $(79.05, 74.72)$ & $(81.42, 78.64)$ \\
monks\_3 & $(90.96, 89.36)$ & $(88.66, 82.27)$ & $(93.37, 83.88)$ & $(71.69, 86.35)$ & $(93.37, 83.02)$ & $(93.98, 90.12)$ \\
 & $(84.38, 81.8)$ & $(86.06, 85.05)$ & $(84.62, 84.62)$ & $(85, 80)$ & $(87.36, 86.39)$ & $(87.33, 86.59)$ \\
musk\_1 & $(85.85, 88.52)$ & $(83.53, 89.86)$ & $(81.55, 82.67)$ & $(86.34, 83.33)$ & $(88.73, 83.33)$ & $(87.32, 90.31)$ \\
 & $(87.46, 89.26)$ & $(87.93, 87.86)$ & $(83.1, 83.1)$ & $(80.91, 83.51)$ & $(82.35, 87.5)$ & $(82.19, 86.96)$ \\
new-thyroid1 & $(98.44, 66.67)$ & $(81.46, 62.5)$ & $(100, 62.67)$ & $(85.94, 66.67)$ & $(100, 68.89)$ & $(100, 69.48)$ \\
 & $(100, 80)$ & $(100, 82.22)$ & $(93.1, 83.1)$ & $(80, 87.36)$ & $(82.73, 84.21)$ & $(100, 87.78)$ \\
oocytes\_trisopterus\_nucleus\_2f & $(71.68, 87.42)$ & $(70.13, 81.91)$ & $(72.98, 82.68)$ & $(70.41, 82.5)$ & $(81.32, 85.68)$ & $(73.63, 87.18)$ \\
 & $(80.98, 84.08)$ & $(84.85, 83.36)$ & $(88.26, 88.26)$ & $(81.43, 84.16)$ & $(84.22, 83.33)$ & $(87.23, 88.11)$ \\
Pima & $(69.13, 68.75)$ & $(69.42, 67.76)$ & $(69.87, 72.68)$ & $(68.7, 66.67)$ & $(69.13, 71.25)$ & $(70, 71.46)$ \\
 & $(78.57, 73.33)$ & $(72.61, 74.96)$ & $(78.26, 78.26)$ & $(80, 74.29)$ & $(86.16, 73.59)$ & $(87.78, 74.8)$ \\
pittsburg\_bridges\_T\_OR\_D & $(76.67, 80.89)$ & $(84.72, 79.68)$ & $(86.67, 80.43)$ & $(86.67, 75)$ & $(83.33, 73.33)$ & $(86.67, 81.82)$ \\
 & $(78.57, 86.36)$ & $(75.47, 85.68)$ & $(83.33, 83.33)$ & $(85.71, 89.13)$ & $(80, 82.47)$ & $(86.83, 85.75)$ \\
planning & $(70.37, 75.88)$ & $(73.44, 72.86)$ & $(70.37, 76)$ & $(69.26, 70)$ & $(72.22, 76.89)$ & $(83.33, 76.95)$ \\
 & $(100, 75.11)$ & $(100, 72.56)$ & $(97.65, 77.65)$ & $(92.35, 73.64)$ & $(85.37, 74.28)$ & $(100, 77.89)$ \\
 ripley & $(86.33, 90.05)$ & $(85.17, 93.87)$ & $(86.47, 90.53)$ & $(85, 97.44)$ & $(92, 87.89)$ & $(87.2, 99.07)$ \\
 & $(90.05, 90.05)$ & $(86.15, 89.84)$ & $(84.75, 84.75)$ & $(93.83, 84.14)$ & $(90.27, 89.07)$ & $(95.6, 91.36)$ \\
segment0 & $(95.12, 86.32)$ & $(93, 86.76)$ & $(95.47, 87.7)$ & $(87.43, 85.68)$ & $(73.81, 87.07)$ & $(96.82, 89.74)$ \\
 & $(100, 92.66)$ & $(93.75, 89.73)$ & $(95.1, 95.1)$ & $(95.68, 92.21)$ & $(93.52, 90.18)$ & $(91.58, 91.78)$ \\
shuttle-6\_vs\_2-3 & $(97.1, 77.58)$ & $(96, 78.25)$ & $(96.67, 71.07)$ & $(96.89, 78.37)$ & $(93.98, 74.85)$ & $(98.55, 78.89)$ \\
 & $(89.35, 72.36)$ & $(87.5, 75.98)$ & $(84.08, 74.08)$ & $(88.57, 72.6)$ & $(85.78, 77.36)$ & $(85.78, 76.67)$ \\
sonar & $(80.48, 75.86)$ & $(79.45, 75.28)$ & $(79.03, 86.75)$ & $(80.65, 87.75)$ & $(81.9, 86.67)$ & $(82.26, 87.73)$ \\
 & $(91.67, 83.02)$ & $(91.06, 87.39)$ & $(85.51, 85.51)$ & $(94.37, 85.53)$ & $(86.96, 85.47)$ & $(85, 86.67)$ \\
spectf & $(80, 90)$ & $(82.77, 90.26)$ & $(83.75, 86.13)$ & $(81.25, 93.75)$ & $(85.85, 86.15)$ & $(83.75, 91.9)$ \\
 & $(84.38, 87.1)$ & $(85.21, 82.67)$ & $(84.27, 84.27)$ & $(83.75, 85.53)$ & $(80.32, 88.19)$ & $(84.83, 89.43)$ \\
statlog\_heart & $(75.72, 84.62)$ & $(73.54, 79.14)$ & $(77.01, 87.95)$ & $(77.65, 84.41)$ & $(84.56, 85.29)$ & $(77.78, 85.04)$ \\
 & $(80.49, 82.5)$ & $(82.14, 79.14)$ & $(82.77, 82.17)$ & $(87.55, 87.78)$ & $(86.32, 85.56)$ & $(87.85, 85.68)$ \\
tic\_tac\_toe & $(100, 100)$ & $(98.51, 99.31)$ & $(97.21, 87.95)$ & $(98.26, 98.51)$ & $(98.94, 100)$ & $(100, 100)$ \\
 & $(92.78, 100)$ & $(97.74, 98.52)$ & $(92.77, 92.17)$ & $(91.03, 98.71)$ & $(100, 100)$ & $(100, 100)$ \\
Vehicle & $(77.05, 78.64)$ & $(73.93, 79.11)$ & $(72.49, 77.43)$ & $(76.68, 75.73)$ & $(76.46, 79.31)$ & $(77.08, 80.22)$ \\
 & $(87.36, 83.17)$ & $(87.66, 87.41)$ & $(86.67, 86.67)$ & $(89.46, 87.4)$ & $(83.77, 82)$ & $(88.89, 88.98)$ \\
vertebral\_column\_2clases & $(84.95, 78.57)$ & $(86.18, 76.39)$ & $(76.89, 77.43)$ & $(80.65, 87.56)$ & $(75.4, 74.29)$ & $(84.42, 81.61)$ \\
 & $(83.33, 75.86)$ & $(82.09, 79.14)$ & $(86.67, 76.67)$ & $(82.78, 77.69)$ & $(81.25, 77.61)$ & $(81.43, 80.57)$ \\
Vote & $(93.08, 89.8)$ & $(93.77, 93.44)$ & $(92.15, 96.14)$ & $(94.62, 96.83)$ & $(90.58, 92.31)$ & $(94.62, 94.84)$ \\
 & $(81.67, 90.72)$ & $(89.76, 91.57)$ & $(84.58, 94.58)$ & $(83.13, 95)$ & $(86, 94.12)$ & $(87.55, 92.04)$ \\
vowel & $(100, 100)$ & $(98.99, 94.83)$ & $(98.89, 96.14)$ & $(95.27, 100)$ & $(97.82, 100)$ & $(100, 100)$ \\
 & $(100, 92.36)$ & $(98.21, 96.49)$ & $(94.58, 94.58)$ & $(82.05, 84.17)$ & $(100, 100)$ & $(100, 100)$ \\
wine & $(85.74, 93.75)$ & $(85, 95.87)$ & $(84.74, 91.02)$ & $(80.77, 92.07)$ & $(92.31, 94.29)$ & $(87.18, 97.8)$ \\
 & $(83.33, 88.24)$ & $(97.78, 87.88)$ & $(92.25, 82.25)$ & $(89.67, 87.64)$ & $(91.24, 85)$ & $(88.24, 88.71)$ \\
wpbc & $(72.41, 81.45)$ & $(75.74, 87.62)$ & $(72.69, 81.02)$ & $(70.69, 80.91)$ & $(61.01, 85.65)$ & $(79.31, 86.13)$ \\
 & $(75.23, 64.33)$ & $(77.37, 66.73)$ & $(72.25, 62.25)$ & $(72.19, 66.72)$ & $(78.33, 67.99)$ & $(79.89, 68.43)$ \\ \hline
(Average ACC, Average Seny)  & $(\underline{84.48}, 82.46)$ & $(82.25, 82.13)$ & $(83.22, 81.8)$ & $(80.89, 82.32)$ & $(84.32, 82.93)$ & $(\textbf{85.55}, 85.68)$ \\ \hline
(Average Spey, Average Pren) & $(85.54, 83.95)$ & $(85.42, 83.2)$ & $(84.89, 83.52)$ & $(85.11, 84.3)$ & $(84.66, 83.94)$ & $(88, 86.28)$ \\ \hline
Average Rank  & $3.25$   &  $4.29$  &  $3.89$  &  $4.62$   &  $2.87$ & $2.07$ \\ \hline
 \multicolumn{7}{l}{$^{\dagger}$ represents the proposed model.}\\
 \multicolumn{7}{l}{The boldface and underline indicate the best and second-best models, respectively, in terms of ACC.} \\
\multicolumn{7}{l}{Seny, Spey, and Pren denote the Sensitivity, Specificity, and Precision, respectively.}
\end{tabular}}
\end{table}

Table \ref{Classification performance of UCI in nonLinear Case.} shows the experimental results including ACC, Seny, Spey, and Pren of the proposed MvTPMSVM model along with the baseline SVM-2K, MvTSVM, MVNPSVM, and PSVM-2V models. The optimal parameters of the proposed MvTPSVM model and the baseline models corresponding to the $error~ rates$ are presented in Table \ref{UCI Error rate}. The comparison in terms of ACC, Seny, Spey, Pren and $error ~rates$ indicate that our proposed MvTPMSVM model yields better performance than the baseline models on most of the datasets. From Table \ref{Classification performance of UCI in nonLinear Case.}, the average ACC of the proposed MvTPMSVM model is $85.55\%$. In contrast, the average ACC for SVM-2K, MvTSVM, MVNPSVM, PSVM-2V and MVLDM models stands at $84.48\%$, $82.25\%$, $80.89\%$, $83.22\%$ and $84.32\%$, respectively. As the average ACC can be influenced by exceptional performance in one dataset that compensates for losses across multiple datasets, it might be considered a biased measure. Hence, we utilize the ranking method to assess the effectiveness and evaluate the performance of the models. Here, each classifier is assigned a rank, where the model exhibiting superior performance receives a lower rank, and the model with inferior performance is assigned a higher rank. In the evaluation of $k$ models across $N$ datasets, $r_j^i$ represents the rank of the $j^{th}$ model on the $i^{th}$ dataset. $R_j = \frac{1}{N}\sum_{i=1}^Nr_j^i$ calculates the average rank of $j^{th}$ model. The average ranks for SVM-2K, MvTSVM, MVNPSVM, PSVM-2V, MVLDM and MvTPMSVM are $3.25$, $4.29$, $4.62$, $3.89$, $2.87$ and $2.07$, respectively. The evident observation is that the proposed MvTPMSVM model demonstrates the most favorable average ranks. Hence, the generalization ability of the proposed MvTPMSVM model surpasses that of the baseline models. Now we conduct the statistical tests to determine the significance of the results. To evaluate if there are statistically significant differences between the models, we employ the Friedman test \cite{demvsar2006statistical}. The Friedman test involves comparing the average ranks attributed to different models, subsequently determining the presence of statistical distinctions among the models based on their respective rankings. The null hypothesis posits that all the models share an equal average rank, indicating comparable performance. The Friedman test adheres to the chi-squared distribution $\chi_F^2$ with $(k-1)$ degrees of freedom (d.o.f.), and its calculation involves: $\chi_F^2 = \frac{12N}{k(k+1)}\left[\sum_j R_j^2 -\frac{k(k+1)^2}{4}\right]$. The Friedman statistic $(F_F)$ is calculated as: $F_F = \frac{(N-1)\chi_f^2}{N(k-1)-\chi_F^2}$, where the d.o.f. for the $F$-distribution is $(k - 1)$ and $(N - 1) \times (k - 1)$. At the $5\%$ level of significance, we obtain $\chi_F^2 = 70.1627$ and $F_F = 18.4965$ for $k = 6$ and $N = 55$. With reference to $F_F(5, 270) = 2.4134$ from statistical $F$-distribution table. We reject the null hypothesis as $18.4965 > 2.2474$. Hence, statistical difference exists among the compared models. Consequently, we employ the Nemenyi post hoc test to analyze the pairwise distinctions between the models. The critical difference $(C.D.)$ is computed as follows: $C.D. = q_\alpha \sqrt{\frac{k(k+1)}{6N}}$, where $q_\alpha$ represents the critical value from the distribution table for the two-tailed Nemenyi test. Referring to the statistical $F$-distribution table, with $q_\alpha=2.850$ at a $5\%$ significance level, the calculated $C.D.$ is $1.0167$. The differences in average ranks between the proposed MvTPMSVM model and the baseline SVM-2K, MvTSVM, MVNPSVM, PSVM-2V, and MVLDM models are $1.18$, $2.22$, $2.55$, $1.82$ and $0.80$ respectively. The average rank difference between the proposed MvTPMSVM with the baseline SVM-2K, MvTSVM, MVNPSVM, and PSVM-2V models are greater than the $C.D.$. According to the Nemenyi post hoc test, noteworthy differences are observed between the proposed MvTPMSVM model and the baseline models (except MVLDM). The MvTPMSVM model surpasses MVLDM in terms of average rank. The superior performance of the proposed MvTPMSVM model compared to the baseline models is evident.

\begin{table}[htp]
\centering
    \caption{$Error ~rate$ and optimal parameters of the proposed MvTPMSVM model along with the baseline models on UCI and KEEL datasets with the non-linear kernel.}
    \label{UCI Error rate}
    \resizebox{1\textwidth}{!}{
\begin{tabular}{lcccccc}
\hline
\text{Model} $\rightarrow$ & SVM-2K \cite{farquhar2005two} & MvTSVM \cite{xie2015multi} & PSVM-2V \cite{tang2017multiview}  & MVNPSVM \cite{tang2018multi}  & MVLDM \cite{hu2024multiview}  & MvTPSVM$^{\dagger}$ \\ \hline
Dataset $\downarrow$  & $Error~ rate$  &  $Error ~rate$  &  $Error~ rate$  &  $Error~ rate$   &  $Error~ rate$   &  $Error~ rate$ \\
 & $(C_1, \sigma)$ & $(C_1, \sigma)$ & $(C_1, \gamma, \sigma)$ & $(C_1, D, \sigma)$ & $(C_1, v_1, v_2, \theta, \sigma)$ & $(C_1, C_2,\sigma)$ \\ \hline
abalone9-18 & $0.0639$ & $0.0769$ & $0.0685$ & $0.0711$ & $0.0648$ & $0.0594$ \\
 & $(16, 0.125)$ & $(0.5, 32)$ & $(0.03125, 0.03125, 0.03125)$ & $(1, 0.4, 8)$ & $(2, 0.25, 0.25, 4, 0.25)$ & $(0.25, 4, 0.03125)$ \\
acute\_inflammation & $0$ & $0.0952$ & $0$ & $0$ & $0$ & $0$ \\
 & $(0.125, 1)$ & $(0.03125, 0.125)$ & $(0.03125, 0.03125, 0.03125)$ & $(0.03125, 0.1, 0.25)$ & $(0.25, 0.25, 0.25, 0.25, 0.25)$ & $(0.5, 0.03125, 0.03125)$ \\
acute\_nephritis & $0$ & $0$ & $0$ & $0$ & $0$ & $0$ \\
 & $(0.125, 0.5)$ & $(0.03125, 1)$ & $(0.03125, 0.03125, 0.03125)$ & $(0.03125, 0.1, 0.03125)$ & $(0.25, 0.25, 0.25, 0.25, 0.25)$ & $(0.25, 16, 0.03125)$ \\
aus & $0.1694$ & $0.1739$ & $0.1208$ & $0.1966$ & $0.1884$ & $0.1643$ \\
 & $(0.125, 32)$ & $(0.03125, 32)$ & $(0.5, 0.5, 32)$ & $(0.0625, 0.3, 4)$ & $(0.25, 0.25, 4, 2, 2)$ & $(2, 16, 1)$ \\
Balance & $0.0291$ & $0.0272$ & $0.25$ & $0.0291$ & $0.0116$ & $0.0465$ \\
 & $(8, 8)$ & $(0.03125, 32)$ & $(0.0625, 0.5, 32)$ & $(2, 0.03125, 4)$ & $(0.25, 0.25, 0.25, 0.25, 4)$ & $(0.25, 0.03125, 0.0625)$ \\
bank & $0.1555$ & $0.1226$ & $0.2906$ & $0.1433$ & $0.1114$ & $0.1069$ \\
 & $(8, 16)$ & $(0.03125, 32)$ & $(0.125, 0.5, 1)$ & $(0.0625, 0.3, 4)$ & $(1, 0.25, 0.25, 1, 4)$ & $(0.5, 0.125, 4)$ \\
Bankruptcy & $0.0133$ & $0.0057$ & $0.0133$ & $0.2533$ & $0.0133$ & $0$ \\
 & $(0.0625, 4)$ & $(0.03125, 32)$ & $(8, 0.7, 2)$ & $(0.03125, 0.03125, 0.0625)$ & $(2, 0.25, 0.25, 2, 4)$ & $(2, 2, 1)$ \\
blood & $0.2187$ & $0.2405$ & $0.192$ & $0.2044$ & $0.2143$ & $0.183$ \\
 & $(1, 0.5)$ & $(2, 16)$ & $(0.25, 0.1, 32)$ & $(0.125, 0.5, 1)$ & $(1, 1, 0.25, 0.25, 1)$ & $(0.25, 2, 0.5)$ \\
breast\_cancer & $0.2765$ & $0.2836$ & $0.1882$ & $0.2333$ & $0.2588$ & $0.2118$ \\
 & $(2, 32)$ & $(0.03125, 32)$ & $(32, 0.8, 1)$ & $(0.03125, 2, 1)$ & $(0.5, 0.5, 0.25, 0.25, 0.5, 2)$ & $(0.25, 0.125, 0.25)$ \\
breast\_cancer\_wisc & $0.0431$ & $0.0469$ & $0.0239$ & $0.0565$ & $0.0239$ & $0.0431$ \\
 & $(0.0625, 16)$ & $(0.5, 32)$ & $(0.03125, 0.1, 2)$ & $(0.25, 0.03125, 2)$ & $(4, 0.25, 0.25, 4, 4)$ & $(0.25, 2, 1)$ \\
breast\_cancer\_wisc\_diag & $0.0794$ & $0.0727$ & $0.0294$ & $0.0765$ & $0.0376$ & $0.0353$ \\
 & $(8, 32)$ & $(0.03125, 32)$ & $(0.25, 0.03125, 2)$ & $(0.03125, 0.1, 2)$ & $(0.5, 0.5, 1, 4, 0.5)$ & $(0.25, 1, 0.25)$ \\
breast\_cancer\_wisc\_prog & $0.3203$ & $0.2862$ & $0.1864$ & $0.2898$ & $0.2864$ & $0.2712$ \\
 & $(2, 8)$ & $(0.03125, 32)$ & $(0.25, 0.1, 32)$ & $(32, 1, 16)$ & $(0.25, 0.25, 0.25, 0.25, 0.25)$ & $(0.25, 0.03125, 1)$ \\
brwisconsin & $0.0392$ & $0.0418$ & $0.0441$ & $0.1618$ & $0.0296$ & $0.0343$ \\
 & $(0.5, 32)$ & $(0.03125, 1)$ & $(0.125, 0.03125, 4)$ & $(0.03125, 0.1, 0.125)$ & $(0.25, 0.25, 0.25, 0.25, 0.25)$ & $(4, 1, 2)$ \\
bupa or liver-disorders & $0.2816$ & $0.3058$ & $0.3204$ & $0.3301$ & $0.2833$ & $0.2766$ \\
 & $(0.5, 32)$ & $(16, 32)$ & $(0.03125, 0.1, 0.03125)$ & $(0.5, 0.03125, 32)$ & $(0.25, 0.25, 0.25, 0.25, 0.25)$ & $(0.25, 16, 4)$ \\
checkerboard\_Data & $0.1456$ & $0.207$ & $0.1449$ & $0.1456$ & $0.1884$ & $0.1353$ \\
 & $(8, 8)$ & $(0.0625, 16)$ & $(0.0625, 1, 8)$ & $(0.03125, 0.1, 4)$ & $(0.25, 0.25, 0.25, 0.25, 0.25)$ & $(2, 32, 32)$ \\
chess\_krvkp & $0.1$ & $0.025$ & $0.024$ & $0.1326$ & $0.0084$ & $0.0501$ \\
 & $(8, 32)$ & $(0.03125, 32)$ & $(0.03125, 0.1, 4)$ & $(4, 0.5, 32)$ & $(2, 0.25, 0.25, 2, 4)$ & $(0.25, 32, 0.03125)$ \\
cleve.mat & $0.191$ & $0.1827$ & $0.1798$ & $0.2584$ & $0.1236$ & $0.2247$ \\
 & $(0.25, 32)$ & $(0.03125, 4)$ & $(4, 0.5, 32)$ & $(0.03125, 0.6, 32)$ & $(0.5, 0.5, 0.25, 0.25, 0.5)$ & $(4, 0.125, 1)$ \\
cmc.mat & $0.3107$ & $0.3062$ & $0.3152$ & $0.2203$ & $0.3107$ & $0.3559$ \\
 & $(8, 32)$ & $(0.03125, 32)$ & $(2, 0.1, 32)$ & $(0.25, 0.03125, 4)$ & $(0.5, 1, 4, 0.5, 4)$ & $(2, 0.5, 1)$ \\
congressional\_voting & $0.4077$ & $0.4459$ & $0.3769$ & $0.4385$ & $0.3846$ & $0.3769$ \\
 & $(0.5, 2)$ & $(0.03125, 4)$ & $(0.03125, 0.6, 0.5)$ & $(0.03125, 0.5, 4)$ & $(0.25, 0.25, 0.25, 0.25, 0.25)$ & $(0.25, 0.25, 2)$ \\
conn\_bench\_sonar\_mines\_rocks & $0.0968$ & $0.2877$ & $0.1774$ & $0.3548$ & $0.1129$ & $0.1774$ \\
 & $(2, 16)$ & $(0.03125, 8)$ & $(0.25, 0.7, 8)$ & $(16, 0.03125, 16)$ & $(0.25, 0.25, 0.25, 0.25, 0.25)$ & $(0.25, 0.03125, 0.25)$ \\
credit\_approval & $0.1449$ & $0.236$ & $0.1353$ & $0.1691$ & $0.1401$ & $0.1449$ \\
 & $(0.125, 16)$ & $(0.03125, 8)$ & $(0.03125, 0.0625, 4)$ & $(0.0625, 0.5, 4)$ & $(0.25, 0.25, 2, 4, 0.25)$ & $(0.25, 4, 32)$ \\
cylinder\_bands & $0.268$ & $0.273$ & $0.2222$ & $0.3725$ & $0.268$ & $0.2484$ \\
 & $(16, 8)$ & $(0.03125, 8)$ & $(2, 0.03125, 8)$ & $(0.03125, 0.4, 8)$ & $(1, 1, 0.25, 0.25, 1)$ & $(0.25, 8, 8)$ \\
echocardiogram & $0.2564$ & $0.2174$ & $0.1538$ & $0.2308$ & $0.2051$ & $0.1795$ \\
 & $(0.5, 8)$ & $(0.25, 16)$ & $(0.0625, 0.4, 16)$ & $(0.25, 0.03125, 4)$ & $(0.25, 0.25, 0.25, 0.25, 2)$ & $(0.25, 1, 0.125)$ \\
fertility & $0.2667$ & $0.1443$ & $0.1$ & $0.1333$ & $0.1333$ & $0.1333$ \\
 & $(0.03125, 0.03125)$ & $(0.03125, 0.03125)$ & $(0.03125, 0.03125, 1)$ & $(2, 0.1, 32)$ & $(0.25, 0.25, 0.25, 0.25, 0.25)$ & $(4, 0.125, 1)$ \\
German & $0.28$ & $0.3043$ & $0.2333$ & $0.2633$ & $0.26$ & $0.2533$ \\
 & $(2, 16)$ & $(0.03125, 8)$ & $(16, 0.03125, 16)$ & $(0.25, 0.7, 8)$ & $(0.25, 0.25, 0.25, 0.25, 1)$ & $(0.25, 0.125, 1)$ \\
Haberman & $0.3077$ & $0.2977$ & $0.2308$ & $0.3297$ & $0.3187$ & $0.3036$ \\
 & $(0.03125, 0.03125)$ & $(8, 8)$ & $(32, 0.7, 8)$ & $(0.03125, 32, 0.125)$ & $(0.25, 0.25, 0.25, 0.25, 4)$ & $(4, 1, 2)$ \\
heart\_hungarian & $0.1705$ & $0.2087$ & $0.1591$ & $0.1591$ & $0.1705$ & $0.1473$ \\
 & $(8, 32)$ & $(0.03125, 4)$ & $(0.125, 0.2, 2)$ & $(0.03125, 0.5, 2)$ & $(0.25, 0.25, 0.25, 0.25, 1)$ & $(0.25, 0.03125, 4)$ \\
hepatitis & $0.1522$ & $0.1927$ & $0.1739$ & $0.2174$ & $0.1522$ & $0.087$ \\
 & $(1, 8)$ & $(0.03125, 0.03125)$ & $(0.03125, 0.8, 16)$ & $(1, 1, 8)$ & $(0.25, 0.25, 0.25, 4, 0.25)$ & $(0.25, 0.5, 0.125)$ \\
horse\_colic & $0.1927$ & $0.2558$ & $0.1818$ & $0.2636$ & $0.1918$ & $0.1909$ \\
 & $(2, 32)$ & $(0.03125, 8)$ & $(1, 0.03125, 8)$ & $(1, 0.2, 2)$ & $(0.25, 0.25, 4, 0.25, 4)$ & $(2, 0.03125, 0.03125)$ \\
Ilpd & $0.1747$ & $0.1993$ & $0.2701$ & $0.1952$ & $0.1707$ & $0.1667$ \\
 & $(2, 0.125)$ & $(0.0625, 32)$ & $(16, 0.03125, 0.5)$ & $(0.03125, 0.8, 16)$ & $(1, 4, 0.25, 1, 2)$ & $(0.25, 0.03125, 0.25)$ \\
Iris & $0$ & $0$ & $0$ & $0$ & $0$ & $0$ \\
 & $(0.03125, 2)$ & $(0.03125, 0.5)$ & $(0.25, 0.1, 16)$ & $(0.03125, 0.03125, 0.03125)$ & $(4, 0.25, 0.25, 4, 0.5)$ & $(0.25, 0.03125, 0.25)$ \\
mammographic & $0.191$ & $0.1913$ & $0.191$ & $0.2431$ & $0.1971$ & $0.1875$ \\
 & $(16, 4)$ & $(1, 16)$ & $(1, 0.7, 1)$ & $(1, 0.03125, 1)$ & $(0.5, 0.25, 0.25, 0.5, 1)$ & $(0.5, 0.03125, 2)$ \\
molec\_biol\_promoter & $0.3226$ & $0.28$ & $0.1613$ & $0.2548$ & $0.1613$ & $0.1935$ \\
 & $(1, 32)$ & $(0.03125, 32)$ & $(1, 0.6, 16)$ & $(32, 0.5, 32)$ & $(0.25, 0.25, 0.25, 0.25, 0.25)$ & $(0.25, 0.25, 4)$ \\
monks\_1 & $0.1923$ & $0.1962$ & $0.1506$ & $0.2434$ & $0.1982$ & $0.1867$ \\
 & $(32, 4)$ & $(0.03125, 8)$ & $(8, 0.03125, 2)$ & $(0.0625, 0.2, 8)$ & $(4, 4, 0.25, 0.25, 4)$ & $(1, 0.0625, 0.125)$ \\
monks\_2 & $0.25$ & $0.2684$ & $0.2444$ & $0.2889$ & $0.2267$ & $0.2111$ \\
 & $(16, 1)$ & $(0.03125, 0.25)$ & $(16, 0.9, 16)$ & $(1, 0.03125, 1)$ & $(2, 2, 0.25, 0.25, 2)$ & $(4, 0.25, 0.03125)$ \\
monks\_3 & $0.0663$ & $0.1134$ & $0.0663$ & $0.2831$ & $0.0904$ & $0.0602$ \\
 & $(16, 8)$ & $(4, 16)$ & $(0.03125, 0.1, 8)$ & $(8, 0.5, 4)$ & $(1, 1, 0.25, 0.25, 1)$ & $(2, 0.03125, 4)$ \\
musk\_1 & $0.1845$ & $0.1647$ & $0.1127$ & $0.1366$ & $0.1415$ & $0.1268$ \\
 & $(4, 32)$ & $(0.03125, 16)$ & $(4, 0.03125, 16)$ & $(0.25, 0.9, 32)$ & $(0.25, 0.25, 0.25, 0.25, 0.25)$ & $(0.25, 0.25, 0.03125)$ \\
new-thyroid1 & $0$ & $0.1854$ & $0$ & $0.1406$ & $0.0156$ & $0$ \\
 & $(8, 32)$ & $(0.03125, 0.25)$ & $(0.03125, 0.1, 0.03125)$ & $(1, 0.03125, 4)$ & $(0.25, 0.25, 4, 0.25, 4)$ & $(0.25, 0.125, 1)$ \\
oocytes\_trisopterus\_nucleus\_2f & $0.2702$ & $0.2987$ & $0.1868$ & $0.2959$ & $0.2832$ & $0.2637$ \\
 & $(8, 8)$ & $(0.5, 32)$ & $(4, 0.03125, 4)$ & $(0.25, 0.1, 8)$ & $(2, 0.25, 0.25, 2, 2)$ & $(0.25, 0.0625, 0.5)$ \\
Pima & $0.3013$ & $0.3058$ & $0.3087$ & $0.313$ & $0.3087$ & $0.3$ \\
 & $(0.5, 32)$ & $(0.5, 32)$ & $(0.25, 0.8, 8)$ & $(2, 0.125, 0.5)$ & $(0.5, 0.25, 0.25, 0.5, 4)$ & $(0.25, 0.125, 0.25)$ \\ \hline
\multicolumn{7}{l}{$^{\dagger}$ represents the proposed model.}
 \end{tabular}}
\end{table}

 \begin{table}[ht!]
\ContinuedFloat
\centering
    \caption{(Continued)}
    \resizebox{1\textwidth}{!}{                                                     
    \begin{tabular}{lcccccc}
\hline
\text{Model} $\rightarrow$ & SVM-2K \cite{farquhar2005two} & MvTSVM \cite{xie2015multi} & PSVM-2V \cite{tang2017multiview}  & MVNPSVM \cite{tang2018multi}  & MVLDM \cite{hu2024multiview}  & MvTPSVM$^{\dagger}$ \\ \hline
Dataset $\downarrow$  & $Error~ rate$  &  $Error~ rate$  &  $Error~ rate$  &  $Error~ rate$   &  $Error~ rate$   &  $Error~ rate$ \\
 & $(C_1, \sigma)$ & $(C_1, \sigma)$ & $(C_1, \gamma, \sigma)$ & $(C_1, D, \sigma)$ & $(C_1, v_1, v_2, \theta, \sigma)$ & $(C_1, C_2,\sigma)$ \\ \hline
pittsburg\_bridges\_T\_OR\_D & $0.1333$ & $0.1528$ & $0.1667$ & $0.1333$ & $0.2333$ & $0.1333$ \\
 & $(8, 8)$ & $(0.03125, 0.03125)$ & $(16, 0.25, 8)$ & $(0.125, 0.6, 2)$ & $(0.5, 0.25, 4, 0.5, 0.25)$ & $(0.25, 0.03125, 0.03125)$ \\
planning & $0.2963$ & $0.2656$ & $0.2778$ & $0.3074$ & $0.2963$ & $0.1667$ \\
 & $(1, 1)$ & $(0.03125, 0.03125)$ & $(0.125, 0.1, 0.125)$ & $(0.125, 0.03125, 2)$ & $(0.25, 0.25, 0.25, 0.25, 0.25)$ & $(2, 0.0625, 0.25)$ \\
ripley & $0.1353$ & $0.1483$ & $0.08$ & $0.15$ & $0.1367$ & $0.128$ \\
 & $(0.125, 0.03125)$ & $(1, 16)$ & $(2, 0.7, 16)$ & $(16, 0.0625, 0.125)$ & $(1, 0.25, 0.5, 1, 0.25)$ & $(0.25, 0.25, 4)$ \\
segment0 & $0.0453$ & $0.07$ & $0.2619$ & $0.1257$ & $0.0488$ & $0.0318$ \\
 & $(2, 32)$ & $(0.03125, 2)$ & $(0.03125, 0.1, 0.03125)$ & $(8, 0.25, 4)$ & $(0.25, 0.25, 0.25, 0.25, 4)$ & $(0.5, 0.03125, 0.03125)$ \\
shuttle-6\_vs\_2-3 & $0.0333$ & $0.04$ & $0.0602$ & $0.0311$ & $0.029$ & $0.0145$ \\
 & $(0.03125, 0.03125)$ & $(0.03125, 0.03125)$ & $(0.03125, 0.03125, 16)$ & $(0.125, 0.1, 16)$ & $(0.25, 0.25, 0.25, 0.25, 0.25)$ & $(0.25, 0.125, 0.03125)$ \\
sonar & $0.2097$ & $0.2055$ & $0.181$ & $0.1935$ & $0.1952$ & $0.1774$ \\
 & $(2, 0.25)$ & $(0.03125, 32)$ & $(8, 0.25, 4)$ & $(1, 0.9, 8)$ & $(4, 4, 0.25, 4, 4)$ & $(1, 0.0625, 0.5)$ \\
spectf & $0.1625$ & $0.1723$ & $0.1415$ & $0.1875$ & $0.2$ & $0.1625$ \\
 & $(2, 1)$ & $(2, 16)$ & $(0.0625, 0.1, 32)$ & $(0.5, 0.03125, 4)$ & $(2, 2, 4, 2, 2)$ & $(4, 0.03125, 0.03125)$ \\
statlog\_heart & $0.2299$ & $0.2646$ & $0.1544$ & $0.2235$ & $0.2428$ & $0.2222$ \\
 & $(0.5, 16)$ & $(0.03125, 4)$ & $(4, 0.03125, 32)$ & $(0.03125, 0.7, 8)$ & $(0.25, 0.25, 0.25, 0.25, 0.25)$ & $(2, 0.03125, 4)$ \\
tic\_tac\_toe & $0.0279$ & $0.0149$ & $0.0106$ & $0.0174$ & $0$ & $0$ \\
 & $(4, 4)$ & $(0.03125, 32)$ & $(0.5, 0.9, 32)$ & $(0.5, 0.03125, 4)$ & $(0.25, 0.25, 0.25, 0.25, 0.25)$ & $(2, 0.0625, 1)$ \\
Vehicle & $0.2751$ & $0.2607$ & $0.2354$ & $0.2332$ & $0.2295$ & $0.2292$ \\
 & $(16, 32)$ & $(0.03125, 2)$ & $(0.03125, 0.1, 0.03125)$ & $(0.03125, 2, 2)$ & $(0.25, 0.25, 0.25, 0.25, 0.25)$ & $(0.5, 0.125, 0.125)$ \\
vertebral\_column\_2clases & $0.2311$ & $0.1382$ & $0.246$ & $0.1935$ & $0.1505$ & $0.1558$ \\
 & $(2, 1)$ & $(0.125, 16)$ & $(16, 16, 2)$ & $(0.5, 0.3, 2)$ & $(2, 2, 0.25, 0.25, 2)$ & $(0.5, 1, 0.0625)$ \\
Vote & $0.0785$ & $0.0623$ & $0.0942$ & $0.0538$ & $0.0692$ & $0.0538$ \\
 & $(4, 16)$ & $(0.03125, 8)$ & $(2, 0.3, 8)$ & $(16, 0.03125, 4)$ & $(2, 2, 0.25, 0.25, 2)$ & $(1, 0.125, 0.0625)$ \\
vowel & $0.0111$ & $0.0101$ & $0.0218$ & $0.0473$ & $0$ & $0$ \\
 & $(2, 1)$ & $(0.03125, 2)$ & $(0.03125, 0.1, 0.03125)$ & $(4, 0.03125, 4)$ & $(0.25, 0.25, 0.25, 0.25, 0.25)$ & $(1, 0.125, 32)$ \\
wine & $0.1526$ & $0.15$ & $0.0769$ & $0.1923$ & $0.1426$ & $0.1282$ \\
 & $(2, 32)$ & $(4, 8)$ & $(1, 0.125, 32)$ & $(0.03125, 0.1, 0.03125)$ & $(2, 0.25, 0.25, 2, 4)$ & $(2, 1, 0.03125)$ \\
wpbc & $0.2731$ & $0.2426$ & $0.3899$ & $0.2931$ & $0.2759$ & $0.2069$ \\
 & $(1, 16)$ & $(0.03125, 16)$ & $(0.03125, 0.3, 32)$ & $(1, 0.125, 32)$ & $(0.25, 0.25, 0.25, 0.25, 0.25)$ & $(1, 0.25, 2)$ \\ \hline
Average $Error~ rate$ & $0.1678$ & $0.1775$ & $0.1678$ & $0.1911$ & $0.1552$ & $0.1445$ \\ \hline
\multicolumn{7}{l}{$^{\dagger}$ represents the proposed model.}
\end{tabular}}
\end{table}

\begin{table}[ht!]
\centering
    \caption{Pairwise win-tie-loss test of the proposed MvTPMSVM model in comparison to the baseline models using non-linear kernel on UCI and KEEL datasets.}
    \label{win_tie_loss_UCI}
     \resizebox{1\textwidth}{!}{
\begin{tabular}{lccccc}
\hline
  & SVM-2K \cite{farquhar2005two} & MvTSVM \cite{xie2015multi} & MVNPSVM \cite{tang2018multi} & PSVM-2V \cite{tang2017multiview} & MVLDM \cite{hu2024multiview} \\ \hline
MvTSVM \cite{xie2015multi} & {[}14  \hspace{0.1cm}   2  \hspace{0.1cm}  39{]} &  &  & & \\
MVNPSVM \cite{tang2018multi} & {[}11  \hspace{0.1cm}   4  \hspace{0.1cm}  40{]} & {[}22  \hspace{0.1cm}   2  \hspace{0.1cm}  31{]} &  &  & \\
PSVM-2V \cite{tang2017multiview} & {[}18   \hspace{0.1cm}  9  \hspace{0.1cm}  28{]} & {[}31  \hspace{0.1cm}   2  \hspace{0.1cm}  22{]} & {[}30  \hspace{0.1cm}   6   \hspace{0.1cm} 19{]} & & \\
MVLDM \cite{hu2024multiview} & {[}28   \hspace{0.1cm}  7  \hspace{0.1cm}  20{]}  & {[}37   \hspace{0.1cm}  2  \hspace{0.1cm}  16{]}  & {[}40   \hspace{0.1cm}  4  \hspace{0.1cm}  11{]}  & {[}31   \hspace{0.1cm}  7  \hspace{0.1cm}  17{]} & \\
MvTPMSVM$^{\dagger}$ & {[}39   \hspace{0.1cm}  6  \hspace{0.1cm}  10{]} & {[}47  \hspace{0.1cm}   2   \hspace{0.1cm}   6{]} & {[}47  \hspace{0.1cm}   6    \hspace{0.1cm} 2{]} & {[}43  \hspace{0.1cm}  8   \hspace{0.1cm}   4{]} & {[}26   \hspace{0.1cm}  6  \hspace{0.1cm}  23{]} \\ \hline
\multicolumn{6}{l}{$^{\dagger}$ represents the proposed model.}
\end{tabular}}
\end{table}

\begin{figure}
\begin{minipage}{.238\linewidth}
\centering
\subfloat[Hippopotamus vs Humpback whale]{\includegraphics[scale=0.16]{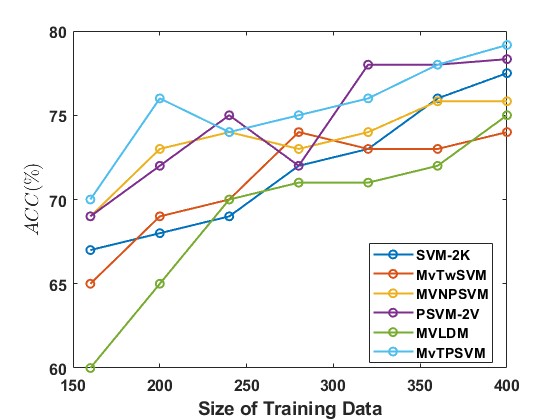}}
\end{minipage}
\begin{minipage}{.238\linewidth}
\centering
\subfloat[Leopard vs \\Seal]{\includegraphics[scale=0.16]{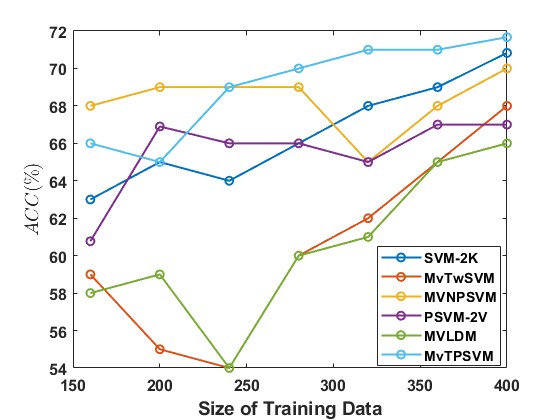}}
\end{minipage}
\begin{minipage}{.236\linewidth}
\centering
\subfloat[Persian cat vs \\Hippopotamus]{\includegraphics[scale=0.16]{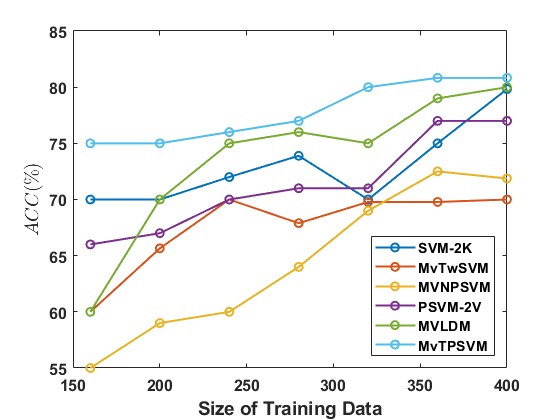}}
\end{minipage}
\begin{minipage}{.238\linewidth}
\centering
\subfloat[Giant panda vs \\ Humpback whale]{\includegraphics[scale=0.16]{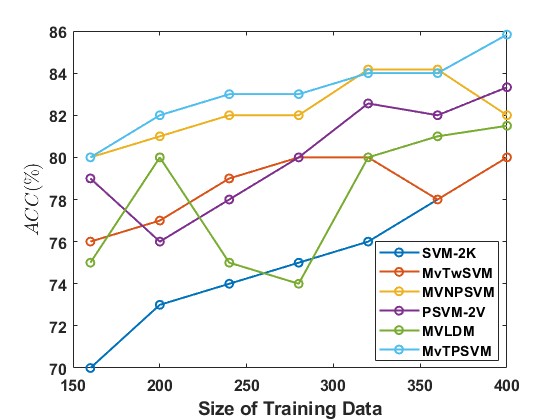}}
\end{minipage}
\caption{The performance comparison of the proposed MvTPMSVM model across various intervals of the training samples using the AwA datasets.}
\label{The performance comparison of the proposed MvTPMSVM model across various intervals of the training dataset using the AwA datasets.}
\end{figure}

Furthermore, in the analysis of models, we employ the pairwise win-tie-loss (W-T-L) sign test. According to the W-T-L sign test, the null hypothesis posits that two models perform equivalently and are expected to secure victories in $N/2$ datasets, where $N$ denotes the count of datasets. If the classification model secures wins in approximately $N/2 + 1.96\sqrt{N}/2$ datasets, then the model is deemed significantly better. In the event of an even number of ties between two models, these ties are evenly distributed between them. Nevertheless, if the number of ties is odd, one tie is disregarded, and the remaining ties are distributed among the specified classifiers. In this instance, with $N = 55$, the occurrence of wins in at least $42.035$ datasets by one of the models signifies a noteworthy distinction between the models. Table \ref{win_tie_loss_UCI} presents the outcomes in terms of pairwise wins, ties, and losses across UCI and KEEL datasets. In each entry, $[x,  y, z]$, $x$ denotes the number of times the model specified in the corresponding row secures a win, $y$ represents the number of ties, and $z$ signifies the occurrences of losses in comparison to the model mentioned in the respective column. Upon meticulous examination of Table \ref{win_tie_loss_UCI}, it is apparent that the proposed MvTPMSVM model outperforms the baseline MvTSVM, MVNPSVM, and PSVM-2V models by achieving a respective number of victories: $47$, $47$, and $43$, out of a total of $55$ datasets. The proposed MvTPMSVM demonstrates a statistically significant distinction compared to the baseline models, except for SVM-2K and MVLDM. Exhibiting a notable level of performance, the MvTPMSVM model succeeds in $39$ out of $55$ datasets over SVM-2K and $26$ out of $55$ datasets over MVLDM. However, the winning percentage of the MvTPMSVM model demonstrates its effectiveness in comparison to the SVM-2K and MVLDM models. It is evident that the proposed MvTPMSVM model exhibits significant superiority when compared to the baseline models.

In Figure \ref{Performance comparison of classification models on UCI and KEEL datasets using Sensitivity, Specificity, and Precision.}, the Seny, Spey, and Pren indicate that the proposed MvTPMSVM model demonstrates competitive performance w.r.t. the baseline models. The proposed MvTPMSVM model achieves an average Seny of $85.68$, Spey of $88.86$, and Pren of $86.28$. These metrics indicate that the MvTPMSVM model outperforms the baseline models, securing the top position in comparison. Thus, the Seny, Spey, and Pren analysis further attests to the superiority of the proposed MvTPMSVM model on UCI and KEEL datasets.

\begin{figure}[htp]
\begin{minipage}{.32\linewidth}
\centering
\subfloat[abalone9-18]{\includegraphics[scale=0.21]{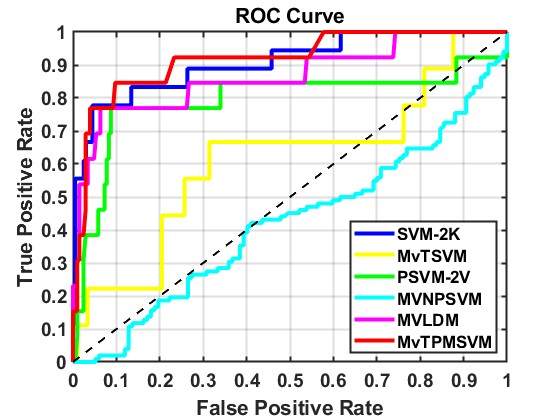}}
\end{minipage}
\begin{minipage}{.32\linewidth}
\centering
\subfloat[blood]{\includegraphics[scale=0.21]{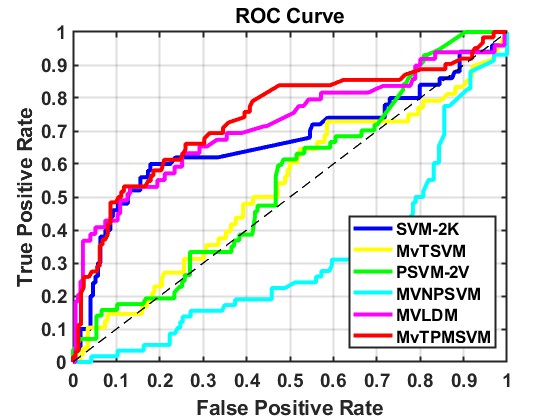}}
\end{minipage}
\begin{minipage}{.32\linewidth}
\centering
\subfloat[monks\_1]{\includegraphics[scale=0.21]{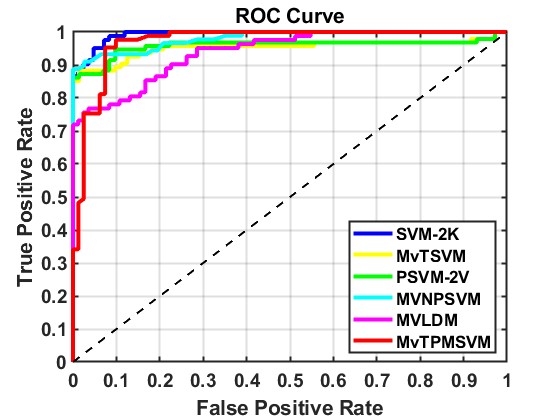}}
\end{minipage}
\caption{ROC curves of the proposed MvTPMSVM model along with the baseline models on the UCI and KEEL datasets.}
\label{ROC curves on the UCI and KEEL datasets.}
\end{figure}

Figure \ref{ROC curves on the UCI and KEEL datasets.} illustrates the ROC curve, highlighting the superior performance of the proposed MvTPMSVM model compared to the baseline models on UCI and KEEL datasets. The ROC curve provides a comprehensive view of the model's diagnostic ability by evaluating the true positive rate against the false positive rate at various threshold settings. The area under the ROC curve (AUC) for the proposed MvTPMSVM model is significantly higher, indicating a better balance between Seny and Spey. This superior AUC suggests that the proposed MvTPMSVM model is more effective in distinguishing between positive and negative cases, leading to more accurate predictions. The proposed MvTPMSVM model means it can better identify true positives, thereby reducing the rate of false negatives and ensuring more reliable detection. These results highlight the robustness and effectiveness of the proposed MvTPMSVM model in handling classification tasks compared to the baseline models.

\begin{figure}
\begin{minipage}{.32\linewidth}
\centering
\subfloat[]{\includegraphics[scale=0.21]{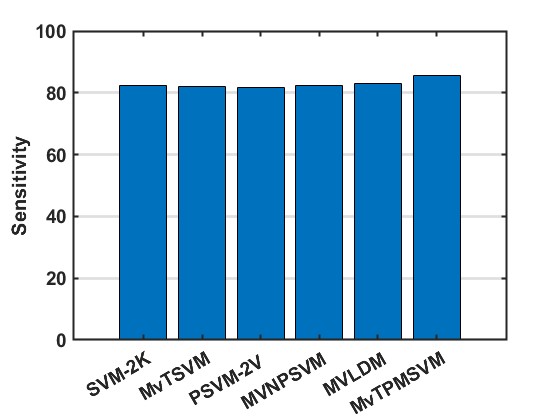}}
\end{minipage}
\begin{minipage}{.32\linewidth}
\centering
\subfloat[]{\includegraphics[scale=0.21]{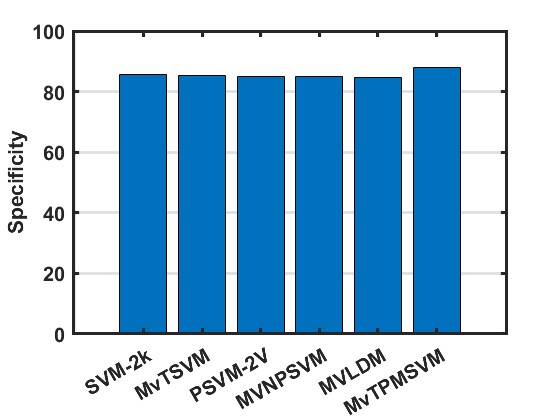}}
\end{minipage}
\begin{minipage}{.32\linewidth}
\centering
\subfloat[]{\includegraphics[scale=0.21]{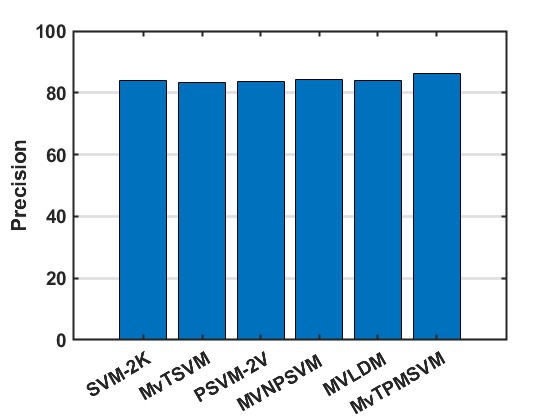}}
\end{minipage}
\caption{Performance comparison of proposed MvTPMSVM model along with the baseline models on UCI and KEEL datasets using Seny, Spey, and Pren}
\label{Performance comparison of classification models on UCI and KEEL datasets using Sensitivity, Specificity, and Precision.}
\end{figure}
\begin{figure}[htp]
\begin{minipage}{.32\linewidth}
\centering
\subfloat[]{\includegraphics[scale=0.21]{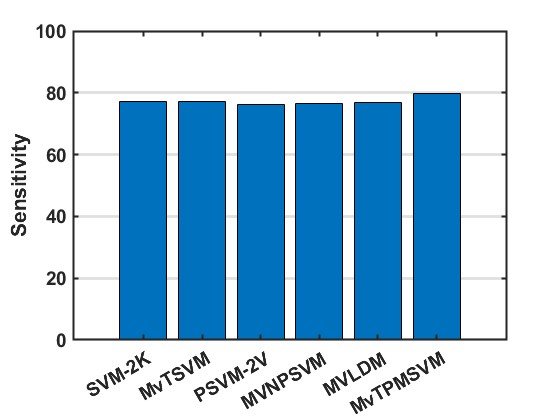}}
\end{minipage}
\begin{minipage}{.32\linewidth}
\centering
\subfloat[]{\includegraphics[scale=0.21]{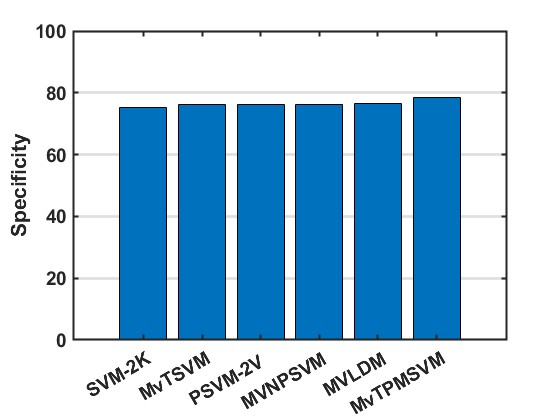}}
\end{minipage}
\begin{minipage}{.32\linewidth}
\centering
\subfloat[]{\includegraphics[scale=0.21]{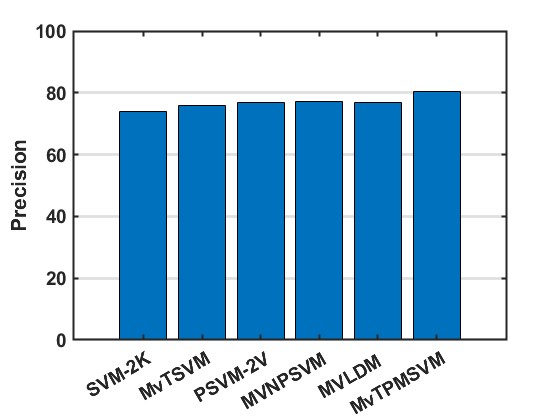}}
\end{minipage}
\caption{Performance comparison of the proposed MvTPMSVM model along with the baseline models on AwA datasets using Seny, Spey, and Pren}
\label{Performance comparison of classification models on AwA datasets using Sensitivity, Specificity, and Precision.}
\end{figure}

\subsection{Experiments on real-world AwA datasets} 
To validate the effectiveness of the proposed MvTPMSVM model, experiments are conducted on the AwA \cite{tang2017multiview} datasets. This dataset comprises $30,475$ images belonging to $50$ different animal classes, and each image comes with six pre-extracted feature representations. We have chosen $10$ animal groups by Tang's studies \cite{tang2017multiview, tang2018multi}, namely persian cat, chimpanzee, pig, hippopotamus, leopard, giant panda, humpback whale, rat, seal, and raccoon. These classes were paired into $45$ binary datasets using a one-vs-one strategy.

\begin{table}[htp]
\centering
    \caption{Performance evaluation of the proposed MvTPMSVM model along with the baseline models on AwA datasets with the non-linear kernel.}
    \label{Classification performance of AwA in nonLinear Case.}
    \resizebox{1\textwidth}{!}{
\begin{tabular}{lcccccc}
\hline
 \text{Model} $\rightarrow$ & SVM-2K \cite{farquhar2005two} & MvTSVM \cite{xie2015multi} & PSVM-2V \cite{tang2017multiview}  & MVNPSVM \cite{tang2018multi}  & MVLDM \cite{hu2024multiview}  & MvTPMSVM$^{\dagger}$ \\ \hline
Dataset $\downarrow$ & (ACC, Seny) & (ACC, Seny) & (ACC, Seny) & (ACC, Seny) & (ACC, Seny) & (ACC, Seny)  \\
 & (Spey, Pren) & (Spey, Pren) & (Spey, Pren) & (Spey, Pren) & (Spey, Pren) & (Spey, Pren) \\ \hline
 Chimpanzee vs Giant panda & $(63.33, 68.85)$ & $(62.5, 65.89)$ & $(68.33, 66.85)$ & $(66.67, 71.68)$ & $(72.22, 66.18)$ & $(68.33, 69.41)$ \\
 & $(77.78, 73.04)$ & $(76.72, 74.85)$ & $(74.06, 72.11)$ & $(73.44, 78.11)$ & $(72.58, 79.23)$ & $(78.24, 80.12)$ \\
Chimpanzee vs Leopard & $(70, 63.64)$ & $(69.17, 61.61)$ & $(72.5, 62.25)$ & $(69.83, 67.21)$ & $(68.75, 62.7)$ & $(74.17, 68.77)$ \\
 & $(82.35, 71.79)$ & $(87.32, 71.36)$ & $(82.82, 78.45)$ & $(88.9, 77.14)$ & $(79.59, 73.41)$ & $(82.5, 79.38)$ \\
Chimpanzee vs Persian cat & $(77.5, 86.21)$ & $(68.33, 82.88)$ & $(70.83, 85.82)$ & $(70, 86.13)$ & $(86.11, 85.9)$ & $(75, 86.87)$ \\
 & $(69.44, 76.92)$ & $(67.27, 77.76)$ & $(74.58, 75.71)$ & $(74.05, 78)$ & $(78.16, 77.01)$ & $(79.53, 78.25)$ \\
Chimpanzee vs Pig & $(72.5, 68.66)$ & $(66.67, 70.14)$ & $(71.67, 68.55)$ & $(68.33, 66.59)$ & $(66.67, 65.22)$ & $(75, 70.33)$ \\
 & $(82.14, 74.8)$ & $(82.63, 73.22)$ & $(74.63, 70.13)$ & $(80.55, 74.89)$ & $(75.22, 75.22)$ & $(78.82, 75.55)$ \\
Chimpanzee vs Hippopotamus & $(74.83, 84.13)$ & $(64.17, 89.24)$ & $(70.83, 82.55)$ & $(74.83, 87.84)$ & $(78.47, 82.5)$ & $(75, 89.12)$ \\
 & $(73.61, 78.52)$ & $(71.98, 74.23)$ & $(74.89, 75.45)$ & $(75.6, 76.12)$ & $(79.52, 80.98)$ & $(83.93, 82.81)$ \\
Chimpanzee vs Humpback whale & $(82.17, 93.75)$ & $(72.5, 100)$ & $(82.5, 97.61)$ & $(70.33, 81.23)$ & $(81.25, 91.18)$ & $(84.17, 100)$ \\
 & $(86.96, 84.23)$ & $(82.85, 82.15)$ & $(88.7, 87.19)$ & $(82.61, 88.61)$ & $(84.7, 82.12)$ & $(85.26, 89.52)$ \\
Chimpanzee vs Raccoon & $(70.83, 60.35)$ & $(64.17, 68.17)$ & $(72.5, 67.39)$ & $(68.33, 72.28)$ & $(72.22, 73.53)$ & $(72.5, 75.52)$ \\
 & $(74.19, 72.27)$ & $(88.82, 78.1)$ & $(89.35, 80.42)$ & $(86.76, 72.31)$ & $(89.44, 79.43)$ & $(82.94, 81.89)$ \\
Chimpanzee vs Rat & $(70.17, 62.12)$ & $(69.67, 60.77)$ & $(70, 61.91)$ & $(70, 66.68)$ & $(68.06, 65.51)$ & $(70.83, 62.46)$ \\
 & $(74.55, 67.77)$ & $(72.13, 63.14)$ & $(77.83, 65.4)$ & $(78.33, 67.38)$ & $(72.11, 64.95)$ & $(75.95, 68.67)$ \\
Chimpanzee vs Seal & $(71.33, 94.74)$ & $(70, 92.86)$ & $(71.17, 93.4)$ & $(70.17, 95.96)$ & $(75.69, 93.33)$ & $(73.33, 98.57)$ \\
 & $(84.29, 76.6)$ & $(81.48, 70.97)$ & $(80.44, 75.44)$ & $(74.07, 74.16)$ & $(79.62, 75.86)$ & $(82.76, 76.87)$ \\
Giant panda vs Leopard & $(60.33, 95.93)$ & $(60.5, 96.32)$ & $(60.83, 92.37)$ & $(60.67, 89.8)$ & $(61.81, 93.42)$ & $(61.67, 94.53)$ \\
 & $(70.21, 69.26)$ & $(75.98, 70.81)$ & $(75.65, 73.79)$ & $(77.26, 76.27)$ & $(73.33, 74.55)$ & $(78.42, 76.11)$ \\
Giant panda vs Persian cat & $(79.5, 76.27)$ & $(76.67, 77.45)$ & $(76.67, 72.45)$ & $(79.17, 77.48)$ & $(66.67, 75)$ & $(80, 78.88)$ \\
 & $(86.54, 81.08)$ & $(87.1, 89.23)$ & $(82.17, 86.27)$ & $(85, 86.1)$ & $(86.67, 88)$ & $(88.28, 85.12)$ \\
Giant panda vs Pig & $(60.83, 75.86)$ & $(65, 75.28)$ & $(61.67, 75.79)$ & $(63.33, 76.2)$ & $(65.97, 79.69)$ & $(66.67, 78.03)$ \\
 & $(57.89, 65.67)$ & $(58, 63.4)$ & $(60.59, 67.34)$ & $(61.74, 66.03)$ & $(62.62, 67.55)$ & $(64.62, 69.73)$ \\
Giant panda vs Hippopotamus & $(67.5, 81.03)$ & $(70, 86.04)$ & $(78.33, 81.01)$ & $(70.67, 74.46)$ & $(74.31, 82.09)$ & $(70.83, 85.94)$ \\
 & $(68.84, 70.15)$ & $(70.33, 75.65)$ & $(65.67, 76.57)$ & $(67.85, 75.11)$ & $(68.75, 74.83)$ & $(67.9, 75.86)$ \\
Giant panda vs Humpback whale & $(85.17, 93.85)$ & $(83.33, 90.91)$ & $(83.33, 87.63)$ & $(84.17, 94.13)$ & $(93.75, 100)$ & $(85.83, 100)$ \\
 & $(88.41, 91.04)$ & $(86.92, 95.61)$ & $(89.55, 98.69)$ & $(85.99, 92.38)$ & $(87.84, 93.53)$ & $(89.1, 96.18)$ \\
Giant panda vs Raccoon & $(60.83, 69.34)$ & $(67.5, 65.17)$ & $(62.5, 67.33)$ & $(65.17, 63.97)$ & $(64.58, 65.71)$ & $(68.33, 68.33)$ \\
 & $(70.59, 60.53)$ & $(66.67, 64.72)$ & $(77.16, 68.78)$ & $(70.03, 70.46)$ & $(70.92, 67.52)$ & $(70.83, 70.48)$ \\
Giant panda vs Rat & $(69.33, 76.56)$ & $(64.17, 72.87)$ & $(68.67, 74.76)$ & $(65.67, 79.14)$ & $(70.14, 83.1)$ & $(70, 84.65)$ \\
 & $(75.38, 75.97)$ & $(79.66, 76.53)$ & $(73.95, 72.85)$ & $(71.17, 72.31)$ & $(75.56, 73.29)$ & $(77.02, 77.78)$ \\
Giant panda vs Seal & $(75, 90)$ & $(72, 91.75)$ & $(74.5, 90.48)$ & $(73.33, 86.8)$ & $(86.81, 87.5)$ & $(75, 82.46)$ \\
 & $(80.6, 85.04)$ & $(78.72, 82.45)$ & $(83.96, 88.33)$ & $(75.02, 87.38)$ & $(86.3, 86.9)$ & $(80.15, 89.81)$ \\
Leopard vs Persian cat & $(81.67, 93.65)$ & $(80, 87.81)$ & $(77.5, 88.42)$ & $(80, 89.47)$ & $(80.56, 89.47)$ & $(81.67, 90.28)$ \\
 & $(76.62, 84.29)$ & $(77.06, 79.84)$ & $(80.05, 78.46)$ & $(82.15, 74.16)$ & $(77.27, 72.93)$ & $(84.73, 86.28)$ \\
Leopard vs Pig & $(70, 73.33)$ & $(66.67, 76.1)$ & $(63.33, 74.39)$ & $(67.5, 78.16)$ & $(68.75, 78.87)$ & $(70, 75.93)$ \\
 & $(70.97, 72.13)$ & $(72.28, 79.09)$ & $(71.65, 70)$ & $(70.33, 76.27)$ & $(75.12, 71.34)$ & $(74.06, 79.49)$ \\
Leopard vs Hippopotamus & $(71.67, 76.12)$ & $(72.5, 74.85)$ & $(70, 75.02)$ & $(69.17, 74.91)$ & $(75, 76.32)$ & $(74.17, 78.98)$ \\
 & $(89.47, 72.26)$ & $(87.27, 76.83)$ & $(82.16, 75.29)$ & $(82.89, 73.55)$ & $(86.32, 76.32)$ & $(86.25, 77.37)$ \\
Leopard vs Humpback whale & $(84.17, 62.59)$ & $(80, 60.68)$ & $(86.33, 60.93)$ & $(85.83, 69.72)$ & $(89.58, 70.12)$ & $(87.5, 70.93)$ \\
 & $(92.59, 92.59)$ & $(98.67, 96.97)$ & $(93.26, 92.87)$ & $(87.51, 91.46)$ & $(85.33, 92.51)$ & $(96.44, 97.18)$ \\
Leopard vs Raccoon & $(62.5, 60.78)$ & $(64.17, 61.08)$ & $(60, 61.67)$ & $(68.33, 65.64)$ & $(56.94, 63.53)$ & $(69.17, 68.33)$ \\
 & $(63.45, 86.88)$ & $(66.36, 87.72)$ & $(64.83, 84.82)$ & $(69.16, 81.04)$ & $(63.19, 82.73)$ & $(69.32, 88.82)$ \\
Leopard vs Rat & $(69.17, 81.48)$ & $(76.67, 84.03)$ & $(70, 84.74)$ & $(75.83, 82.83)$ & $(65.28, 88.24)$ & $(76.67, 90.54)$ \\
 & $(73.33, 77.19)$ & $(79.18, 72.36)$ & $(78.57, 78.47)$ & $(76.06, 74.52)$ & $(78.82, 70.59)$ & $(80.92, 75.61)$ \\
Leopard vs Seal & $(70.83, 70.18)$ & $(68.33, 76.47)$ & $(67.5, 72.82)$ & $(70, 72.39)$ & $(81.25, 73.08)$ & $(71.67, 76.08)$ \\
 & $(85.11, 76.92)$ & $(83.2, 79.69)$ & $(88.98, 73.87)$ & $(81.78, 73.06)$ & $(87.14, 76)$ & $(80.49, 79.24)$ \\
Persian cat vs Pig & $(75.83, 77.78)$ & $(77.5, 74.24)$ & $(73.33, 79.85)$ & $(77.5, 77.08)$ & $(69.44, 72.86)$ & $(77.5, 75.25)$ \\
 & $(72.41, 75)$ & $(65.83, 77.51)$ & $(76.76, 78.33)$ & $(78.74, 76.67)$ & $(77.08, 72.71)$ & $(80, 76.09)$ \\
Persian cat vs Hippopotamus & $(79.33, 73.33)$ & $(70, 76.81)$ & $(77.5, 72.79)$ & $(72.5, 76.14)$ & $(75.69, 76.25)$ & $(80.83, 78.57)$ \\
 & $(89.8, 80.73)$ & $(70.22, 77.64)$ & $(81.99, 87.59)$ & $(82.51, 87.14)$ & $(86.89, 85.18)$ & $(87.89, 88.37)$ \\
Persian cat vs Humpback whale & $(72.5, 88.71)$ & $(70.83, 84.33)$ & $(70.83, 83.68)$ & $(74.67, 83.99)$ & $(85.42, 82.59)$ & $(75, 84.38)$ \\
 & $(79.71, 83.97)$ & $(78.27, 89.88)$ & $(82.03, 87.64)$ & $(72.3, 88)$ & $(83.33, 87.72)$ & $(82.05, 88.09)$ \\
Persian cat vs Raccoon & $(68.33, 65.15)$ & $(70.83, 66.71)$ & $(68.83, 63.4)$ & $(70.83, 61.48)$ & $(65.97, 64.78)$ & $(72.5, 68.47)$ \\
 & $(76.79, 70.49)$ & $(79.77, 76.92)$ & $(70.33, 79.93)$ & $(78.39, 74.89)$ & $(71.43, 75.05)$ & $(72.5, 74.93)$ \\
Persian cat vs Rat & $(62.5, 78.85)$ & $(48.33, 71.96)$ & $(53.33, 78.31)$ & $(49.17, 76.47)$ & $(56.94, 72.24)$ & $(48.33, 78.89)$ \\
 & $(74.67, 64.57)$ & $(79.77, 62.4)$ & $(72.79, 67.89)$ & $(76.45, 68.12)$ & $(73.85, 63.03)$ & $(80.48, 69.82)$ \\
Persian cat vs Seal & $(82.5, 87.93)$ & $(70.83, 88.7)$ & $(85.83, 80.79)$ & $(75.83, 83.35)$ & $(83.33, 83.58)$ & $(82.5, 85)$ \\
 & $(78.46, 12.93)$ & $(79.62, 72.57)$ & $(72.51, 76.73)$ & $(76.12, 78.61)$ & $(81.16, 82.35)$ & $(83.78, 84.28)$ \\
  \hline
  \multicolumn{7}{l}{$^{\dagger}$ represents the proposed model.}\\
 \multicolumn{7}{l}{Seny, Spey, and Pren denote the Sensitivity, Specificity, and Precision, respectively.}
\end{tabular}}
\end{table}
 
 \begin{table}[ht!]
\ContinuedFloat
\centering
    \caption{(Continued)}
   \resizebox{1\textwidth}{!}{                                                     
    \begin{tabular}{lcccccc}
\hline
\text{Model} $\rightarrow$ & SVM-2K \cite{farquhar2005two} & MvTSVM \cite{xie2015multi} & PSVM-2V \cite{tang2017multiview}  & MVNPSVM \cite{tang2018multi}  & MVLDM \cite{hu2024multiview}  & MvTPMSVM \\ \hline
Dataset $\downarrow$ & (ACC, Seny) & (ACC, Seny) & (ACC, Seny) & (ACC, Seny) & (ACC, Seny) & (ACC, Seny)  \\
 & (Spey, Pren) & (Spey, Pren) & (Spey, Pren) & (Spey, Pren) & (Spey, Pren) & (Spey, Pren) \\ \hline
 Pig vs Hippopotamus & $(64.17, 76.27)$ & $(67.5, 76.89)$ & $(68.33, 75.43)$ & $(68.33, 72.27)$ & $(72.22, 70.67)$ & $(71.67, 77.27)$ \\
 & $(60.81, 67.67)$ & $(66.79, 63.75)$ & $(63.25, 61.85)$ & $(70.77, 63.06)$ & $(74.65, 62.6)$ & $(72.86, 77.89)$ \\
 Pig vs Humpback whale & $(85, 89.09)$ & $(82.5, 88.27)$ & $(87.5, 83.86)$ & $(85.83, 88.62)$ & $(88.89, 85.83)$ & $(88.33, 89.61)$ \\
 & $(80.33, 84.48)$ & $(90, 80.74)$ & $(85.23, 80.54)$ & $(88.28, 86.67)$ & $(84.15, 89.61)$ & $(92.73, 89.93)$ \\
Pig vs Raccoon & $(60.83, 67.86)$ & $(55.83, 68.23)$ & $(55, 69.89)$ & $(48.33, 62.85)$ & $(62.5, 66.67)$ & $(57.5, 65.85)$ \\
 & $(57.58, 62.3)$ & $(56.5, 64.41)$ & $(61.78, 65.78)$ & $(61.72, 88.46)$ & $(62.78, 87.87)$ & $(62.78, 89.7)$ \\
Pig vs Rat & $(59.17, 80.36)$ & $(54.17, 82.78)$ & $(54.17, 80.47)$ & $(50.83, 84)$ & $(64.58, 86.81)$ & $(59.33, 88.67)$ \\
 & $(64.22, 64.75)$ & $(68.21, 65.06)$ & $(67.14, 68.27)$ & $(63.46, 65.54)$ & $(60.23, 67.52)$ & $(65.68, 69.22)$ \\
Pig vs Seal & $(75, 75.44)$ & $(71.67, 77.4)$ & $(69.17, 71.16)$ & $(72.67, 72.2)$ & $(72.92, 71.08)$ & $(75.5, 74.13)$ \\
 & $(70.49, 72.88)$ & $(76.96, 71.67)$ & $(73.41, 74.24)$ & $(70.84, 75.1)$ & $(70.59, 75.47)$ & $(79.74, 76.26)$ \\
Hippopotamus vs Humpback whale & $(77.5, 95.16)$ & $(74.17, 91.49)$ & $(78.33, 76.21)$ & $(75.83, 79.79)$ & $(79.86, 73.33)$ & $(79.17, 81.79)$ \\
 & $(80.82, 87.41)$ & $(88.64, 81.47)$ & $(84.97, 89.89)$ & $(83.47, 83.1)$ & $(87.92, 80.54)$ & $(85.36, 87.04)$ \\
Hippopotamus vs Raccoon & $(70.5, 63.49)$ & $(70.67, 62.98)$ & $(70.83, 61.33)$ & $(69.17, 65.69)$ & $(75.69, 65.71)$ & $(71.67, 68.49)$ \\
 & $(78.43, 70.18)$ & $(70.62, 70.06)$ & $(71.58, 76.45)$ & $(79.29, 73.88)$ & $(74.65, 75.18)$ & $(72.09, 78.58)$ \\
Hippopotamus vs Rat & $(68.33, 68.97)$ & $(63.33, 67.14)$ & $(68, 64.09)$ & $(59.17, 61.83)$ & $(64.58, 61.08)$ & $(68.33, 65.57)$ \\
 & $(66.67, 87.8)$ & $(68.57, 88.93)$ & $(68.35, 78.56)$ & $(68.32, 77.09)$ & $(61.86, 79.18)$ & $(70.18, 87.8)$ \\
Hippopotamus vs Seal & $(69.17, 59.68)$ & $(70.33, 58.57)$ & $(70.83, 60.06)$ & $(70.83, 59.51)$ & $(60.42, 60.83)$ & $(70.83, 63.33)$ \\
 & $(75.51, 66.67)$ & $(77.78, 68.11)$ & $(78.8, 62.65)$ & $(75.4, 64.16)$ & $(78.62, 67.15)$ & $(76.67, 68.57)$ \\
Humpback whale vs Raccoon & $(80.67, 90.77)$ & $(74.17, 95.62)$ & $(80.17, 98.42)$ & $(80, 93.4)$ & $(83.33, 92.93)$ & $(80, 99.89)$ \\
 & $(72.19, 71.47)$ & $(74.22, 76.83)$ & $(70.84, 78.56)$ & $(78.98, 71.04)$ & $(77.18, 75.24)$ & $(77.56, 76.92)$ \\
Humpback whale vs Rat & $(80, 82.81)$ & $(78.17, 88.94)$ & $(80, 87.65)$ & $(79.17, 88.08)$ & $(77.78, 85.31)$ & $(80, 85)$ \\
 & $(61.38, 86.89)$ & $(68.46, 83.01)$ & $(66.84, 82.57)$ & $(68.58, 82.48)$ & $(63.56, 79.22)$ & $(62.86, 87.47)$ \\
Humpback whale vs Seal & $(75, 71.43)$ & $(76.67, 75.21)$ & $(77.5, 76.93)$ & $(74.67, 78.48)$ & $(78.47, 80.28)$ & $(77.5, 81.67)$ \\
 & $(81.82, 76.27)$ & $(81.97, 79.26)$ & $(86.74, 81.97)$ & $(85.87, 84.7)$ & $(87.03, 78.62)$ & $(80.24, 82.27)$ \\
Raccoon vs Rat & $(71.33, 78.72)$ & $(72.5, 72.87)$ & $(59.17, 74.32)$ & $(71.67, 74.01)$ & $(65.28, 73.08)$ & $(72.5, 79.09)$ \\
 & $(65.22, 64.91)$ & $(67.05, 66.08)$ & $(68.1, 66.14)$ & $(63.81, 68.55)$ & $(66.28, 69.51)$ & $(69, 68.23)$ \\
Raccoon vs Seal & $(72.5, 74.24)$ & $(68.33, 74.72)$ & $(71.67, 75.81)$ & $(71.67, 75.8)$ & $(75.69, 77.61)$ & $(74.17, 78.57)$ \\
 & $(80.74, 81.67)$ & $(78.13, 88.08)$ & $(82.08, 81.54)$ & $(83.8, 86.43)$ & $(82.22, 84.82)$ & $(80, 89.07)$ \\
Rat vs Seal & $(65, 64.29)$ & $(60, 62.41)$ & $(64.17, 63.22)$ & $(66.67, 60.79)$ & $(69.87, 67.78)$ & $(70, 69.65)$ \\
 & $(61.02, 62.61)$ & $(64.55, 60.22)$ & $(64.17, 69.27)$ & $(68.8, 68.36)$ & $(68.27, 65.81)$ & $(72.34, 67.38)$ \\ \hline
(Average ACC, Average Seny) & $(71.02, 77.11)$ & $(69.4, 77.21)$ & $(71.02, 76.08)$ & $(70.50, 76.59)$ & $(\underline{73.33}, 76.97)$ & $(\textbf{73.57}, 79.65)$ \\
(Average Spey, Average Pren) & $(75.21, 74.06)$ & $(76.23, 76.03)$ & $(76.34, 76.96)$ & $(76.31, 77.09)$ & $(76.66, 76.92)$ & $(78.34, 80.36)$ \\
\hline
Average Rank &  $3.53$  &  $4.78$  &  $3.82$ & $4.19$ & $2.74$  &  $1.93$  \\ \hline
\multicolumn{7}{l}{$^{\dagger}$ represents the proposed model.}\\
 \multicolumn{7}{l}{The boldface and underline indicate the best and second-best models, respectively, in terms of ACC.} \\
 \multicolumn{7}{l}{Seny, Spey, and Pren denote the Sensitivity, Specificity, and Precision, respectively.}
\end{tabular}}
\end{table}

Table \ref{Classification performance of AwA in nonLinear Case.} displays the ACC, Seny, Spey, and Pren of the proposed MvTPMSVM model along with SVM-2K, MvTSVM, MVNPSVM, PSVM-2V, and MVLDM models. The $error~ rates$ and the optimal parameters of the proposed MvTPSVM model and the baseline models are presented in Table \ref{AwA Error rate}. From Table \ref{Classification performance of AwA in nonLinear Case.}, it is evident that the proposed MvTPMSVM model exhibits superior generalization performance across the majority of datasets. The average ACC of the proposed MvTPMSVM model and the baseline SVM-2K, MvTSVM, MVNPSVM, PSVM-2V, and MVLDM models are $73.57\%$, $71.02 \%$, $69.4\%$, $70.50\%$, $71.02\%$ and $73.33\%$, respectively. Table \ref{Classification performance of AwA in nonLinear Case.} presents the average rank of all models, determined by their ACC values. It is worth noting that among all the models, our proposed MvTPMSVM model has the lowest average rank. Moreover, we perform the Friedman statistical test, followed by Nemenyi post hoc tests. For the significance level of $\alpha = 0.05$, we calculate $\chi_F^2 = 50.81$, $F_F = 12.8345$, and $F_F(5, 220) = 2.2550$. The null hypothesis is rejected as $F_F(5, 220)<F_F$. Now, the Nemenyi post-hoc test is employed to identify significant differences among the pairwise comparisons. We calculate $C.D. = 1.1240$, indicating that the average ranking of the models in Table \ref{Classification performance of AwA in nonLinear Case.} should have a minimum difference by $1.1240$ to be considered statistically significant. The differences in average rank between the proposed MvTPMSVM model and the baseline SVM-2K, MvTSVM, MVNPSVM, PSVM-2V, and MVLDM models are $1.60$, $2.85$, $2.26$, $1.89$ and $0.81$, respectively. $C.D.$ is exceeded by the observed differences. Thus, as per the Nemenyi post hoc test, the proposed MvTPMSVM model, exhibits significant distinctions from SVM-2K, MvTSVM, MVNPSVM, and PSVM-2V except MVLDM. Consequently, the proposed MvTPMSVM model demonstrates superior performance compared to baseline models except MVLDM. However, the proposed MvTPMSVM model surpasses MVLDM in terms of average rank. The evident superiority of the proposed MvTPMSVM model is evident when contrasted with the baseline models.

\begin{table}[htp]
\centering
    \caption{$Error~ rate$ and optimal parameters of the proposed MvTPMSVM model along with the baseline models on AwA datasets with the non-linear kernel.}
    \label{AwA Error rate}
    \resizebox{1\textwidth}{!}{
\begin{tabular}{lcccccc}
\hline
\text{Model} $\rightarrow$ & SVM-2K \cite{farquhar2005two} & MvTSVM \cite{xie2015multi} & PSVM-2V \cite{tang2017multiview}  & MVNPSVM \cite{tang2018multi}  & MVLDM \cite{hu2024multiview}  & MvTPMSVM$^{\dagger}$ \\ \hline
Dataset $\downarrow$ & $Error~ rate$  &  $Error~ rate$  &  $Error ~rate$  &  $Error~ rate$   &  $Error~ rate$   &  $Error~ rate$ \\
 & $(C_1, \sigma)$ & $(C_1, \sigma)$ & $(C_1, \gamma, \sigma)$ & $(C_1, D, \sigma)$ & $(C_1, v_1, v_2, \theta, \sigma)$ & $(C_1, C_2,\sigma)$ \\ \hline
Chimpanzee vs Giant panda & $0.3667$ & $0.375$ & $0.3167$ & $0.3333$ & $0.2778$ & $0.3167$ \\
 & $(2, 1)$ & $(0.03125, 1)$ & $(0.0625, 16, 4)$ & $(4, 0.8, 8)$ & $(2, 0.25, 0.25, 0.25, 4, 2)$ & $(2, 0.03125, 0.0625)$ \\
Chimpanzee vs Leopard & $0.3$ & $0.3083$ & $0.275$ & $0.3017$ & $0.3125$ & $0.2583$ \\
 & $(1, 0.5)$ & $(0.03125, 32)$ & $(1, 8, 4)$ & $(8, 0.9, 4)$ & $(2, 0.5, 4, 0.25, 0.25, 0.5)$ & $(8, 0.03125, 0.03125)$ \\
Chimpanzee vs Persian cat & $0.225$ & $0.3167$ & $0.2917$ & $0.3$ & $0.1389$ & $0.25$ \\
 & $(4, 16)$ & $(0.03125, 32)$ & $(0.03125, 4, 8)$ & $(0.03125, 0.1, 1)$ & $(0.5, 0.25, 4, 0.25, 0.25, 4)$ & $(0.0625, 0.03125, 0.5)$ \\
Chimpanzee vs Pig & $0.275$ & $0.3333$ & $0.2833$ & $0.3167$ & $0.3333$ & $0.25$ \\
 & $(2, 0.5)$ & $(0.03125, 32)$ & $(0.03125, 32, 16)$ & $(4, 0.9, 8)$ & $(0.25, 0.25, 0.25, 0.25, 0.5, 1)$ & $(1, 0.03125, 0.03125)$ \\
Chimpanzee vs Hippopotamus & $0.2517$ & $0.3583$ & $0.2917$ & $0.2517$ & $0.2153$ & $0.25$ \\
 & $(2, 32)$ & $(0.03125, 32)$ & $(0.03125, 2, 1)$ & $(32, 0.8, 1)$ & $(0.5, 0.5, 0.25, 0.25, 0.5, 2)$ & $(2, 2, 0.0625)$ \\
Chimpanzee vs Humpback whale & $0.1783$ & $0.275$ & $0.175$ & $0.2967$ & $0.1875$ & $0.1583$ \\
 & $(4, 2)$ & $(0.03125, 32)$ & $(0.25, 0.0625, 2)$ & $(0.03125, 0.1, 1)$ & $(2, 2, 0.25, 4, 2, 2)$ & $(0.0625, 32, 0.25)$ \\
Chimpanzee vs Raccoon & $0.2917$ & $0.3583$ & $0.275$ & $0.3167$ & $0.2778$ & $0.275$ \\
 & $(4, 32)$ & $(0.03125, 32)$ & $(0.03125, 1, 2)$ & $(8, 0.9, 4)$ & $(4, 4, 0.25, 0.25, 4, 2)$ & $(8, 0.03125, 0.03125)$ \\
Chimpanzee vs Rat & $0.2983$ & $0.3033$ & $0.3$ & $0.3$ & $0.3194$ & $0.2917$ \\
 & $(1, 4)$ & $(0.03125, 32)$ & $(0.03125, 4, 1)$ & $(0.0625, 0.1, 1)$ & $(1, 1, 0.25, 0.25, 1, 1)$ & $(8, 0.03125, 0.125)$ \\
Chimpanzee vs Seal & $0.2867$ & $0.3$ & $0.2883$ & $0.2983$ & $0.2431$ & $0.2667$ \\
 & $(2, 16)$ & $(0.03125, 32)$ & $(1, 0.25, 1)$ & $(0.125, 0.2, 2)$ & $(0.25, 0.25, 0.25, 0.25, 0.25, 2)$ & $(0.25, 0.03125, 0.0625)$ \\
Giant panda vs Leopard & $0.3967$ & $0.395$ & $0.3917$ & $0.3933$ & $0.3819$ & $0.3833$ \\
 & $(4, 1)$ & $(0.03125, 32)$ & $(0.0625, 16, 2)$ & $(8, 0.9, 4)$ & $(1, 2, 2, 4, 0.25, 1)$ & $(8, 1, 0.03125)$ \\
Giant panda vs Persian cat & $0.205$ & $0.2333$ & $0.2333$ & $0.2083$ & $0.3333$ & $0.2$ \\
 & $(2, 0.5)$ & $(0.03125, 32)$ & $(2, 8, 2)$ & $(0.0625, 0.1, 1)$ & $(4, 0.25, 0.25, 0.25, 4, 2)$ & $(2, 0.5, 0.03125)$ \\
Giant panda vs Pig & $0.3917$ & $0.35$ & $0.3833$ & $0.3667$ & $0.3403$ & $0.3333$ \\
 & $(2, 0.5)$ & $(8, 32)$ & $(0.5, 0.25, 1)$ & $(0.0625, 0.1, 0.5)$ & $(0.25, 0.25, 4, 4, 0.5, 2)$ & $(8, 0.25, 0.0625)$ \\
Giant panda vs Hippopotamus & $0.325$ & $0.3$ & $0.2167$ & $0.2933$ & $0.2569$ & $0.2917$ \\
 & $(1, 1)$ & $(0.03125, 32)$ & $(0.125, 32, 4)$ & $(0.03125, 0.1, 0.5)$ & $(1, 0.5, 0.25, 0.25, 4, 4)$ & $(2, 0.03125, 0.0625)$ \\
Giant panda vs Humpback whale & $0.1483$ & $0.1667$ & $0.1667$ & $0.1583$ & $0.0625$ & $0.1417$ \\
 & $(8, 32)$ & $(0.03125, 32)$ & $(0.25, 0.03125, 1)$ & $(0.03125, 0.1, 1)$ & $(4, 0.25, 0.25, 0.25, 2, 1)$ & $(0.5, 0.25, 0.5)$ \\
Giant panda vs Raccoon & $0.3917$ & $0.325$ & $0.375$ & $0.3483$ & $0.3542$ & $0.3167$ \\
 & $(4, 8)$ & $(0.03125, 32)$ & $(0.5, 1, 2)$ & $(8, 0.9, 4)$ & $(2, 1, 0.25, 0.25, 4, 4)$ & $(0.5, 0.125, 0.03125)$ \\
Giant panda vs Rat & $0.3067$ & $0.3583$ & $0.3133$ & $0.3433$ & $0.2986$ & $0.3$ \\
 & $(2, 1)$ & $(0.03125, 32)$ & $(0.0625, 4, 2)$ & $(0.0625, 0.1, 1)$ & $(0.5, 0.25, 4, 4, 1, 4)$ & $(2, 0.03125, 0.25)$ \\
Giant panda vs Seal & $0.25$ & $0.28$ & $0.255$ & $0.2667$ & $0.1319$ & $0.25$ \\
 & $(2, 1)$ & $(0.03125, 32)$ & $(0.25, 0.25, 2)$ & $(0.03125, 0.1, 0.5)$ & $(1, 0.5, 0.25, 0.25, 2, 1)$ & $(0.5, 1, 0.125)$ \\
Leopard vs Persian cat & $0.1833$ & $0.2$ & $0.225$ & $0.2$ & $0.1944$ & $0.1833$ \\
 & $(8, 8)$ & $(0.03125, 32)$ & $(2, 8, 2)$ & $(0.03125, 0.1, 0.5)$ & $(2, 0.25, 0.25, 0.25, 4, 1)$ & $(0.5, 0.125, 1)$ \\
Leopard vs Pig & $0.3$ & $0.3333$ & $0.3667$ & $0.325$ & $0.3125$ & $0.3$ \\
 & $(2, 0.5)$ & $(0.03125, 32)$ & $(0.5, 0.5, 0.5)$ & $(0.03125, 0.1, 0.25)$ & $(1, 0.25, 0.25, 0.25, 2, 0.5)$ & $(1, 0.0625, 0.125)$ \\
Leopard vs Hippopotamus & $0.2833$ & $0.275$ & $0.3$ & $0.3083$ & $0.25$ & $0.2583$ \\
 & $(1, 1)$ & $(0.03125, 32)$ & $(1, 0.03125, 1)$ & $(0.03125, 0.1, 0.25)$ & $(2, 0.5, 0.25, 0.25, 0.25, 2)$ & $(0.125, 0.0625, 0.125)$ \\
Leopard vs Humpback whale & $0.1583$ & $0.2$ & $0.1367$ & $0.1417$ & $0.1042$ & $0.125$ \\
 & $(4, 0.5)$ & $(0.03125, 32)$ & $(0.03125, 8, 2)$ & $(0.03125, 0.1, 1)$ & $(0.5, 0.5, 0.25, 0.25, 0.25, 2)$ & $(8, 32, 1)$ \\
Leopard vs Raccoon & $0.375$ & $0.3583$ & $0.4$ & $0.3167$ & $0.4306$ & $0.3083$ \\
 & $(8, 0.5)$ & $(0.03125, 32)$ & $(0.03125, 2, 1)$ & $(0.03125, 0.1, 0.125)$ & $(4, 1, 0.25, 0.25, 1, 2)$ & $(2, 0.03125, 0.03125)$ \\
Leopard vs Rat & $0.3083$ & $0.2333$ & $0.3$ & $0.2417$ & $0.3472$ & $0.2333$ \\
 & $(1, 0.5)$ & $(0.03125, 32)$ & $(4, 0.03125, 2)$ & $(0.03125, 0.1, 1)$ & $(2, 0.5, 0.25, 0.25, 4, 2)$ & $(0.03125, 1, 0.5)$ \\
Leopard vs Seal & $0.2917$ & $0.3167$ & $0.325$ & $0.3$ & $0.1875$ & $0.2833$ \\
 & $(4, 0.5)$ & $(0.03125, 32)$ & $(0.5, 1, 2)$ & $(0.03125, 0.1, 0.5)$ & $(1, 0.5, 0.25, 0.25, 4, 1)$ & $(16, 0.5, 0.03125)$ \\
Persian cat vs Pig & $0.2417$ & $0.225$ & $0.2667$ & $0.225$ & $0.3056$ & $0.225$ \\
 & $(1, 2)$ & $(0.03125, 32)$ & $(8, 0.03125, 8)$ & $(4, 0.9, 8)$ & $(4, 2, 0.25, 0.25, 0.25, 4)$ & $(0.03125, 0.25, 0.03125)$ \\
Persian cat vs Hippopotamus & $0.2067$ & $0.3$ & $0.225$ & $0.275$ & $0.2431$ & $0.1917$ \\
 & $(2, 1)$ & $(0.03125, 32)$ & $(2, 0.5, 2)$ & $(0.0625, 0.9, 2)$ & $(2, 0.25, 0.25, 0.25, 2, 4)$ & $(2, 0.03125, 0.03125)$ \\
Persian cat vs Humpback whale & $0.275$ & $0.2917$ & $0.2917$ & $0.2533$ & $0.1458$ & $0.25$ \\
 & $(1, 1)$ & $(0.03125, 32)$ & $(0.0625, 0.125, 2)$ & $(0.0625, 0.9, 16)$ & $(0.5, 0.5, 0.25, 4, 2, 4)$ & $(16, 0.0625, 0.125)$ \\
Persian cat vs Raccoon & $0.3167$ & $0.2917$ & $0.3117$ & $0.2917$ & $0.3403$ & $0.275$ \\
 & $(4, 1)$ & $(0.03125, 32)$ & $(0.03125, 0.03125, 0.03125)$ & $(16, 0.9, 2)$ & $(4, 0.25, 0.25, 0.25, 1, 1)$ & $(8, 32, 0.03125)$ \\
Persian cat vs Rat & $0.375$ & $0.5167$ & $0.4667$ & $0.5083$ & $0.4306$ & $0.5167$ \\
 & $(2, 4)$ & $(0.03125, 32)$ & $(8, 4, 8)$ & $(4, 0.9, 8)$ & $(0.5, 1, 0.25, 0.25, 0.25, 4)$ & $(1, 0.03125, 0.25)$ \\
Persian cat vs Seal & $0.175$ & $0.2917$ & $0.1417$ & $0.2417$ & $0.1667$ & $0.175$ \\
 & $(1, 4)$ & $(0.03125, 32)$ & $(0.25, 0.03125, 2)$ & $(0.0625, 0.9, 4)$ & $(0.5, 0.25, 0.25, 0.25, 0.5, 2)$ & $(0.5, 2, 0.03125)$ \\
Pig vs Hippopotamus & $0.3583$ & $0.325$ & $0.3167$ & $0.3167$ & $0.2778$ & $0.2833$ \\
 & $(8, 16)$ & $(0.03125, 0.03125)$ & $(1, 0.03125, 8)$ & $(0.03125, 0.1, 0.25)$ & $(0.25, 0.25, 0.25, 0.25, 2, 1)$ & $(32, 4, 0.0625)$ \\
Pig vs Humpback whale & $0.15$ & $0.175$ & $0.125$ & $0.1417$ & $0.1111$ & $0.1167$ \\
 & $(8, 16)$ & $(0.03125, 32)$ & $(0.0625, 32, 2)$ & $(0.03125, 0.1, 0.5)$ & $(1, 0.25, 0.25, 0.25, 0.25, 4)$ & $(32, 2, 0.5)$ \\
Pig vs Raccoon & $0.3917$ & $0.4417$ & $0.45$ & $0.5167$ & $0.375$ & $0.425$ \\
 & $(8, 32)$ & $(0.03125, 0.03125)$ & $(16, 1, 16)$ & $(2, 0.9, 16)$ & $(1, 0.25, 0.25, 0.25, 2, 0.5)$ & $(0.0625, 4, 0.03125)$ \\
Pig vs Rat & $0.4083$ & $0.4583$ & $0.4583$ & $0.4917$ & $0.3542$ & $0.4067$ \\
 & $(2, 2)$ & $(0.03125, 0.03125)$ & $(0.125, 1, 2)$ & $(0.03125, 0.1, 0.5)$ & $(4, 0.25, 0.25, 0.25, 4, 2)$ & $(8, 0.125, 0.25)$ \\
Pig vs Seal & $0.25$ & $0.2833$ & $0.3083$ & $0.2733$ & $0.2708$ & $0.245$ \\
 & $(2, 1)$ & $(0.03125, 32)$ & $(0.25, 0.03125, 2)$ & $(0.03125, 0.1, 1)$ & $(1, 0.25, 0.25, 0.25, 1, 2)$ & $(4, 32, 0.0625)$ \\
Hippopotamus vs Humpback whale & $0.225$ & $0.2583$ & $0.2167$ & $0.2417$ & $0.2014$ & $0.2083$ \\
 & $(1, 4)$ & $(0.03125, 32)$ & $(0.25, 8, 4)$ & $(0.03125, 0.1, 0.5)$ & $(1, 0.25, 0.25, 0.25, 0.25, 4)$ & $(32, 0.03125, 0.25)$ \\
Hippopotamus vs Raccoon & $0.295$ & $0.2933$ & $0.2917$ & $0.3083$ & $0.2431$ & $0.2833$ \\
 & $(4, 8)$ & $(0.03125, 32)$ & $(0.5, 0.03125, 1)$ & $(0.0625, 0.9, 1)$ & $(1, 0.25, 2, 4, 1, 4)$ & $(8, 2, 0.03125)$ \\
Hippopotamus vs Rat & $0.3167$ & $0.3667$ & $0.32$ & $0.4083$ & $0.3542$ & $0.3167$ \\
 & $(1, 0.5)$ & $(0.03125, 32)$ & $(0.03125, 2, 2)$ & $(0.03125, 0.1, 1)$ & $(1, 1, 0.25, 0.25, 2, 0.5)$ & $(4, 4, 0.25)$ \\
Hippopotamus vs Seal & $0.3083$ & $0.2967$ & $0.2917$ & $0.2917$ & $0.3958$ & $0.2917$ \\
 & $(2, 0.5)$ & $(0.03125, 16)$ & $(0.5, 0.5, 2)$ & $(0.03125, 0.1, 0.5)$ & $(2, 2, 4, 4, 0.25, 4)$ & $(8, 16, 0.125)$ \\
Humpback whale vs Raccoon & $0.1933$ & $0.2583$ & $0.1983$ & $0.2$ & $0.1667$ & $0.2$ \\
 & $(4, 2)$ & $(0.03125, 32)$ & $(2, 0.03125, 2)$ & $(4, 0.5, 4)$ & $(1, 0.5, 0.25, 0.25, 0.5, 2)$ & $(8, 32, 0.0625)$ \\
Humpback whale vs Rat & $0.2$ & $0.2183$ & $0.2$ & $0.2083$ & $0.2222$ & $0.2$ \\
 & $(2, 2)$ & $(0.03125, 32)$ & $(0.03125, 16, 8)$ & $(8, 0.9, 4)$ & $(1, 0.5, 0.25, 2, 0.25, 0.5)$ & $(8, 32, 0.125)$ \\ \hline
 \multicolumn{7}{l}{$^{\dagger}$ represents the proposed model.}
 \end{tabular}}
\end{table}

 \begin{table}[htp]
\ContinuedFloat
\centering
    \caption{(Continued)}
    \resizebox{1\textwidth}{!}{                                                     
    \begin{tabular}{lcccccc}
\hline
\text{Model} $\rightarrow$ & SVM-2K \cite{farquhar2005two} & MvTSVM \cite{xie2015multi} & PSVM-2V \cite{tang2017multiview}  & MVNPSVM \cite{tang2018multi}  & MVLDM \cite{hu2024multiview}  & MvTPMSVM$^{\dagger}$ \\ \hline
Dataset $\downarrow$ & $Error~ rate$  &  $Error ~rate$  & $Error~ rate$  &  $Error~ rate$   &  $Error~ rate$   &  $Error~ rate$ \\
 & $(C_1, \sigma)$ & $(C_1, \sigma)$ & $(C_1, \gamma, \sigma)$ & $(C_1, D, \sigma)$ & $(C_1, v_1, v_2, \theta, \sigma)$ & $(C_1, C_2,\sigma)$ \\ \hline
Humpback whale vs Seal & $0.25$ & $0.2333$ & $0.225$ & $0.2533$ & $0.2153$ & $0.225$ \\
 & $(1, 1)$ & $(2, 32)$ & $(2, 0.03125, 8)$ & $(8, 0.9, 4)$ & $(2, 0.5, 0.25, 0.25, 0.5, 4)$ & $(8, 0.125, 0.0625)$ \\
Raccoon vs Rat & $0.2867$ & $0.275$ & $0.4083$ & $0.2833$ & $0.3472$ & $0.275$ \\
 & $(2, 0.5)$ & $(0.03125, 32)$ & $(0.03125, 2, 2)$ & $(0.0625, 0.1, 1)$ & $(0.25, 1, 0.25, 0.25, 0.25, 4)$ & $(0.25, 0.5, 0.03125)$ \\
Raccoon vs Seal & $0.275$ & $0.3167$ & $0.2833$ & $0.2833$ & $0.2431$ & $0.2583$ \\
 & $(1, 2)$ & $(0.03125, 32)$ & $(0.5, 0.03125, 2)$ & $(0.03125, 0.1, 1)$ & $(1, 0.5, 0.25, 0.25, 0.25, 4)$ & $(8, 0.25, 0.25)$ \\
Rat vs Seal & $0.35$ & $0.4$ & $0.3583$ & $0.3333$ & $0.3013$ & $0.3$ \\
 & $(2, 1)$ & $(0.03125, 32)$ & $(0.03125, 32, 16)$ & $(0.0625, 0.9, 2)$ & $(0.5, 0.25, 0.25, 4, 0.5, 8)$ & $(2, 0.03125, 0.03125)$ \\ \hline
 Average $Error~ rate$ &  $0.2808$  & $0.3060$ & $0.2898$  &  $0.2950$  & $0.2667$  &  $0.2643$ \\ \hline
 \multicolumn{7}{l}{$^{\dagger}$ represents the proposed model.}
\end{tabular}}
\end{table}

\begin{figure}
\begin{minipage}{.238\linewidth}
\centering
\subfloat[abalone9-18]{\includegraphics[scale=0.16]{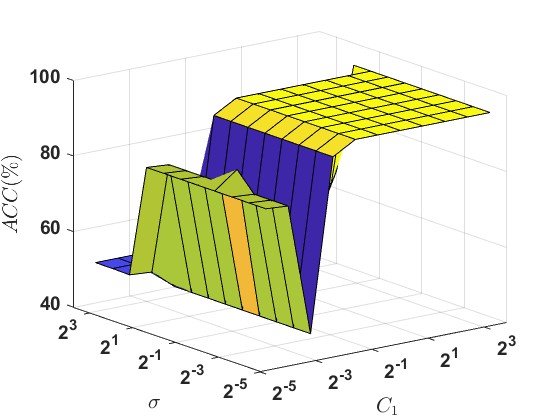}}
\end{minipage}
\begin{minipage}{.238\linewidth}
\centering
\subfloat[Blood]{\includegraphics[scale=0.16]{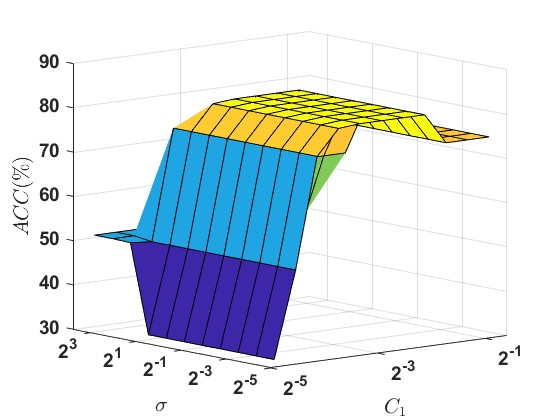}}
\end{minipage}
\begin{minipage}{.238\linewidth}
\centering
\subfloat[bupa or liver-disorders]{\includegraphics[scale=0.16]{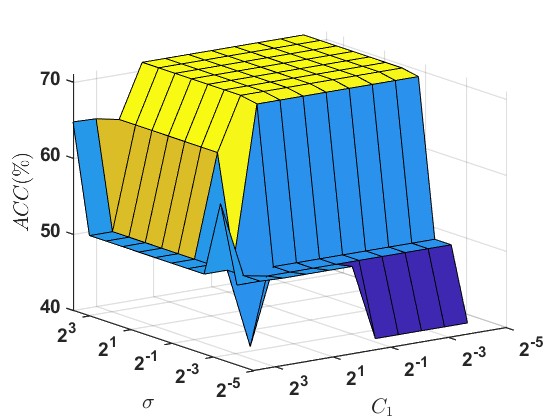}}
\end{minipage}
\begin{minipage}{.238\linewidth}
\centering
\subfloat[Fertility]{\includegraphics[scale=0.16]{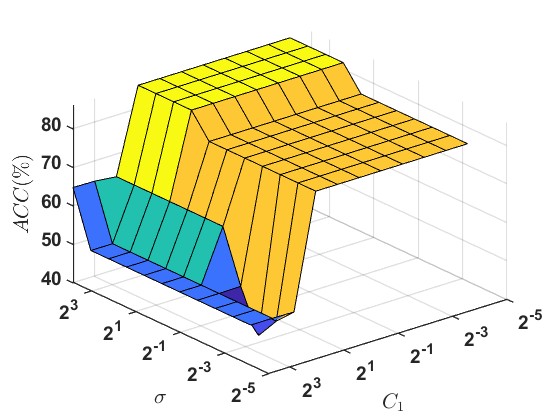}}
\end{minipage}
\caption{The effect of hyperparameters $(C_1, \sigma)$ tuning on the ACC of UCI and KEEL datasets on the performance of proposed MvTPMSVM model.}
\label{Insensitivity performance of the proposed MvTPMSVM model for classification of real-world UCI and KEEL datasets to the user-specified parameters}
\end{figure}
Subsequently, additional experiments were performed by adjusting the sizes of the training sets, and their respective performance is illustrated in Figures \ref{The performance comparison of the proposed MvTPMSVM model across various intervals of the training dataset using the AwA datasets.}. We employ a random selection method, using $30\%$ of the samples for testing, while systematically altering the set sizes allocated for training. Figure \ref{The performance comparison of the proposed MvTPMSVM model across various intervals of the training dataset using the AwA datasets.}, exhibiting different training sizes, consistently demonstrates that the proposed MvTPMSVM model generally outperforms the baseline models. Moreover, as the training sizes increase, the MvTPMSVM model exhibits superior performance compared to baseline models. 

Based on Figure \ref{Performance comparison of classification models on AwA datasets using Sensitivity, Specificity, and Precision.}, the Seny, Spey, and Pren analysis indicates that the performance of the proposed MvTPMSVM model achieved the highest position in the baseline models, demonstrating its competitive nature. The average Seny, Spey, and Pren are $79.65$, $78.34$, and $80.36$, respectively. The statistical tests provide further evidence that the proposed models outperformed all the existing models in the AwA datasets and achieved the highest rankings in Seny, Spey, and Pren analyses.



\subsection{Effect of hyperparameters \texorpdfstring{$C_1$}{C1} and \texorpdfstring{$\sigma$}{sigma}}
In this subsection, we conduct an in-depth investigation to understand the influence of the hyperparameters $C_1$ and $\sigma$ on the comprehensive predictive capacity of the proposed MvTPMSVM model. Figure \ref{Insensitivity performance of the proposed MvTPMSVM model for classification of real-world UCI and KEEL datasets to the user-specified parameters} depicts the results of sensitivity analysis conducted on both UCI and KEEL datasets. The ACC is assessed by altering the parameters $C_1$ and $\sigma$. It is observable that with the escalation of $C_1$ and $\sigma$ values, there is a corresponding increase in ACC. 
However, once a specific threshold is exceeded, the ACC stabilizes, indicating that further increments in $C_1$ and $\sigma$ 
exhibit a diminishing effect on testing ACC. Consequently, meticulous selection of hyperparameters for the proposed MvTPMSVM model is imperative to attain optimal generalization performance.

\section{Conclusions and Future Work}
\label{Conclusions}
In this paper, we proposed a novel multiview twin parametric margin support vector machine (MvTPMSVM) to address the challenges faced by MvTSVM. The proposed MvTPMSVM model alleviates the influence of heteroscedastic noise present in the data, simultaneously reducing the computational costs typically associated with SVM-2K,  MvTSVM, and their variants. Furthermore, the proposed MvTPMSVM model directly applies the kernel trick for nonlinear cases, enabling it to solve the exact formulation. In order to showcase the effectiveness and efficiency of the proposed MvTPMSVM model, we performed an extensive series of experiments and subjected them to thorough statistical analyses. We conducted experiments on $55$ UCI and KEEL datasets, and these experimental results underwent a comprehensive statistical assessment employing a ranking scheme, Friedman test, Nemenyi post-hoc test, and win-tie-loss sign test. Our findings from the experiments, combined with the statistical analyses, indicate that the proposed MvTPMSVM model outperforms baseline models in terms of generalization performance. Furthermore, the proposed model underwent testing on both synthetic and $45$ real-world AwA datasets. The outcomes demonstrated notable enhancements in ACC and generalization performance by employing multiple views in the analysis. The proposed MvTPMSVM model exhibits superior performance compared to certain early and late fusion models, as well as selected state-of-the-art multiview models. Our proposed model has shown exceptional performance in binary classification problems, their evaluation in multiview (incorporating more than two views) multiclass problems has not been conducted. A critical avenue for future research would involve adapting the proposed model to be suitable for multiview multiclass problems. 
Also, our model designed to handle balanced datasets, has the potential to enhance its robustness by integrating techniques tailored to address class imbalance learning. This refers to scenarios where the distribution of classes in the training data is highly skewed, with one or more classes significantly outnumbering others. By incorporating methods specifically designed to tackle class imbalance, the model becomes more adept at effectively handling datasets with disparate class distributions. The source code of the proposed MvTPMSVM model is available at \url{https://github.com/mtanveer1/MvTPMSVM}.

\section*{Acknowledgment}
This work is supported by Indian government's Department of Science and Technology (DST) through the MTR/2021/000787 grant as part of the Mathematical Research Impact-Centric Support (MATRICS) scheme.

\bibliography{refs.bib}

\begin{thebibliography}{48}
\providecommand{\natexlab}[1]{#1}
\providecommand{\url}[1]{\texttt{#1}}
\expandafter\ifx\csname urlstyle\endcsname\relax
  \providecommand{\doi}[1]{doi: #1}\else
  \providecommand{\doi}{doi: \begingroup \urlstyle{rm}\Url}\fi

\bibitem[Chao and Sun(2016)]{chao2016consensus}
Guoqing Chao and Shiliang Sun.
\newblock Consensus and complementarity based maximum entropy discrimination for multi-view classification.
\newblock \emph{Information Sciences}, 367:\penalty0 296--310, 2016.

\bibitem[Cortes and Vapnik(1995)]{cortes1995support}
C~Cortes and V~Vapnik.
\newblock Support-vector network-. machine learning 20: 273--297.
\newblock \emph{Portfolio Selection, Journal of Global Optimization}, 43\penalty0 (2-3), 1995.
\newblock \doi{https://doi.org/10.1007/BF00994018}.

\bibitem[Dem{\v{s}}ar(2006)]{demvsar2006statistical}
Janez Dem{\v{s}}ar.
\newblock Statistical comparisons of classifiers over multiple data sets.
\newblock \emph{The Journal of Machine Learning Research}, 7:\penalty0 1--30, 2006.

\bibitem[Derrac et~al.(2015)Derrac, Garcia, Sanchez, and Herrera]{derrac2015keel}
J~Derrac, S~Garcia, L~Sanchez, and F~Herrera.
\newblock K{EEL} data-mining software tool: Data set repository, integration of algorithms and experimental analysis framework.
\newblock \emph{J. Mult. Valued Log. Soft Comput}, 17:\penalty0 255--287, 2015.

\bibitem[Dua and Graff(2017)]{dua2017uci}
Dheeru Dua and Casey Graff.
\newblock {UCI} machine learning repository.
\newblock \emph{Available: http://archive.ics.uci.edu/ml}, 2017.

\bibitem[Farquhar et~al.(2005)Farquhar, Hardoon, Meng, Shawe-Taylor, and Szedmak]{farquhar2005two}
Jason Farquhar, David Hardoon, Hongying Meng, John Shawe-Taylor, and Sandor Szedmak.
\newblock Two view learning: S{VM-2K}, theory and practice.
\newblock \emph{Advances in Neural Information Processing Systems}, 18, 2005.

\bibitem[Ganaie et~al.(2022)Ganaie, Tanveer, and Lin]{ganaie2022large}
M.~A. Ganaie, M.~Tanveer, and C.~T. Lin.
\newblock Large-scale fuzzy least squares twin {SVM}s for class imbalance learning.
\newblock \emph{IEEE Transactions on Fuzzy Systems}, 30\penalty0 (11):\penalty0 4815--4827, 2022.

\bibitem[Gupta and Gupta(2023)]{gupta2023least}
Umesh Gupta and Deepak Gupta.
\newblock Least squares structural twin bounded support vector machine on class scatter.
\newblock \emph{Applied Intelligence}, 53\penalty0 (12):\penalty0 15321--15351, 2023.

\bibitem[Hao(2010)]{hao2010new}
Pei-Yi Hao.
\newblock New support vector algorithms with parametric insensitive/margin model.
\newblock \emph{Neural Networks}, 23\penalty0 (1):\penalty0 60--73, 2010.

\bibitem[Hu et~al.(2024, doi: 10.1109/TNNLS.2023.3349142)Hu, Xiao, Zheng, Zhu, and Hsu]{hu2024multiview}
Kun Hu, Yingyuan Xiao, Wenguang Zheng, Wenxin Zhu, and Ching-Hsien Hsu.
\newblock Multiview large margin distribution machine.
\newblock \emph{IEEE Transactions on Neural Networks and Learning Systems}, 2024, doi: 10.1109/TNNLS.2023.3349142.

\bibitem[Jayadeva et~al.(2007)Jayadeva, Khemchandani, and Chandra]{khemchandani2007twin}
Jayadeva, Reshma Khemchandani, and Suresh Chandra.
\newblock Twin support vector machines for pattern classification.
\newblock \emph{IEEE Transactions on Pattern Analysis and Machine Intelligence}, 29\penalty0 (5):\penalty0 905--910, 2007.

\bibitem[Kumar and Gopal(2009)]{kumar2009least}
M.~Arun Kumar and M.~Gopal.
\newblock Least squares twin support vector machines for pattern classification.
\newblock \emph{Expert Systems with Applications}, 36\penalty0 (4):\penalty0 7535--7543, 2009.

\bibitem[Li et~al.(2006)Li, Allinson, Tao, and Li]{li2006multitraining}
Jing Li, Nigel Allinson, Dacheng Tao, and Xuelong Li.
\newblock Multitraining support vector machine for image retrieval.
\newblock \emph{IEEE Transactions on Image Processing}, 15\penalty0 (11):\penalty0 3597--3601, 2006.

\bibitem[Mangasarian and Wild(2005)]{mangasarian2005multisurface}
Olvi~L Mangasarian and Edward~W Wild.
\newblock Multisurface proximal support vector machine classification via generalized eigenvalues.
\newblock \emph{IEEE Transactions on Pattern Analysis and Machine Intelligence}, 28\penalty0 (1):\penalty0 69--74, 2005.

\bibitem[Moosaei et~al.(2023)Moosaei, Ganaie, Hlad{\'\i}k, and Tanveer]{moosaei2023inverse}
Hossein Moosaei, M.~A. Ganaie, Milan Hlad{\'\i}k, and M.~Tanveer.
\newblock Inverse free reduced universum twin support vector machine for imbalanced data classification.
\newblock \emph{Neural Networks}, 157:\penalty0 125--135, 2023.

\bibitem[Moosaei et~al.(2024)Moosaei, Bazikar, and Hlad{\'\i}k]{moosaei2024multi}
Hossein Moosaei, Fatemeh Bazikar, and Milan Hlad{\'\i}k.
\newblock Multi-task twin support vector machine with universum data.
\newblock \emph{Engineering Applications of Artificial Intelligence}, 132:\penalty0 107951, 2024.

\bibitem[Peng(2011)]{peng2011tpmsvm}
Xinjun Peng.
\newblock T{PMSVM}: a novel twin parametric-margin support vector machine for pattern recognition.
\newblock \emph{Pattern Recognition}, 44\penalty0 (10-11):\penalty0 2678--2692, 2011.

\bibitem[Quadir and Tanveer(2024)]{abdul2024granular}
A.~Quadir and M.~Tanveer.
\newblock Granular {B}all {T}win {S}upport {V}ector {M}achine {W}ith {P}inball {L}oss {F}unction.
\newblock \emph{IEEE Transactions on Computational Social Systems}, 2024.
\newblock \doi{10.1109/TCSS.2024.3411395}.

\bibitem[Richhariya et~al.(2020)Richhariya, Tanveer, Rashid, and Initiative]{richhariya2020diagnosis}
B.~Richhariya, M.~Tanveer, A.~H. Rashid, and Alzheimer’s Disease~Neuroimaging Initiative.
\newblock Diagnosis of {A}lzheimer's disease using universum support vector machine based recursive feature elimination ({USVM-RFE}).
\newblock \emph{Biomedical Signal Processing and Control}, 59:\penalty0 101903, 2020.

\bibitem[Sajid et~al.(2024)Sajid, Sharma, Beheshti, Tanveer, and {Alzheimer's Disease Neuroimaging Initiative}]{sajid2024decoding}
M.~Sajid, R.~Sharma, I.~Beheshti, M.~Tanveer, and {Alzheimer's Disease Neuroimaging Initiative}.
\newblock Decoding cognitive health using machine learning: {A} comprehensive evaluation for diagnosis of significant memory concern.
\newblock \emph{Wiley Interdisciplinary Reviews: Data Mining and Knowledge Discovery}, page e1546, 2024.
\newblock \doi{https://doi.org/10.1002/widm.1546}.

\bibitem[Sch{\"o}lkopf et~al.(2000)Sch{\"o}lkopf, Smola, Williamson, and Bartlett]{scholkopf2000new}
Bernhard Sch{\"o}lkopf, Alex~J Smola, Robert~C Williamson, and Peter~L Bartlett.
\newblock New support vector algorithms.
\newblock \emph{Neural Computation}, 12\penalty0 (5):\penalty0 1207--1245, 2000.

\bibitem[Shao et~al.(2011)Shao, Zhang, Wang, and Deng]{shao2011improvements}
Yuan-Hai Shao, Chun-Hua Zhang, Xiao-Bo Wang, and Nai-Yang Deng.
\newblock Improvements on twin support vector machines.
\newblock \emph{IEEE Transactions on Neural Networks}, 22\penalty0 (6):\penalty0 962--968, 2011.

\bibitem[Shao et~al.(2023)Shao, Lv, Huang, and Bai]{shao2023twin}
Yuan-Hai Shao, Xiao-Jing Lv, Ling-Wei Huang, and Lan Bai.
\newblock Twin {SVM} for conditional probability estimation in binary and multiclass classification.
\newblock \emph{Pattern Recognition}, 136:\penalty0 109253, 2023.

\bibitem[Sindhwani et~al.(2005)Sindhwani, Niyogi, and Belkin]{sindhwani2005co}
Vikas Sindhwani, Partha Niyogi, and Mikhail Belkin.
\newblock A co-regularization approach to semi-supervised learning with multiple views.
\newblock In \emph{Proceedings of ICML Workshop on Learning with Multiple views}, pages 74--79. Citeseer, 2005.

\bibitem[Sun et~al.(2021)Sun, Fujita, Zheng, and Ai]{sun2021multi}
Jie Sun, Hamido Fujita, Yujiao Zheng, and Wenguo Ai.
\newblock Multi-class financial distress prediction based on support vector machines integrated with the decomposition and fusion methods.
\newblock \emph{Information Sciences}, 559:\penalty0 153--170, 2021.

\bibitem[Szedmak and Shawe-Taylor(2007)]{szedmak2007synthesis}
Sandor Szedmak and John Shawe-Taylor.
\newblock Synthesis of maximum margin and multiview learning using unlabeled data.
\newblock \emph{Neurocomputing}, 70\penalty0 (7-9):\penalty0 1254--1264, 2007.

\bibitem[Tang et~al.(2017)Tang, Tian, Zhang, and Liu]{tang2017multiview}
Jingjing Tang, Yingjie Tian, Peng Zhang, and Xiaohui Liu.
\newblock Multiview privileged support vector machines.
\newblock \emph{IEEE Transactions on Neural Networks and Learning Systems}, 29\penalty0 (8):\penalty0 3463--3477, 2017.

\bibitem[Tang et~al.(2018)Tang, Li, Tian, and Liu]{tang2018multi}
Jingjing Tang, Dewei Li, Yingjie Tian, and Dalian Liu.
\newblock Multi-view learning based on nonparallel support vector machine.
\newblock \emph{Knowledge-Based Systems}, 158:\penalty0 94--108, 2018.

\bibitem[Tanveer(2015)]{tanveer2015application}
M.~Tanveer.
\newblock Application of smoothing techniques for linear programming twin support vector machines.
\newblock \emph{Knowledge and Information Systems}, 45\penalty0 (1):\penalty0 191--214, 2015.

\bibitem[Tanveer et~al.(2016)Tanveer, Khan, and Ho]{tanveer2016robust}
M.~Tanveer, M.~A. Khan, and Shen-Shyang Ho.
\newblock Robust energy-based least squares twin support vector machines.
\newblock \emph{Applied Intelligence}, 45:\penalty0 174--186, 2016.

\bibitem[Tanveer et~al.(2019{\natexlab{a}})Tanveer, Sharma, and Suganthan]{tanveer2019general}
M.~Tanveer, A.~Sharma, and P.~N. Suganthan.
\newblock General twin support vector machine with pinball loss function.
\newblock \emph{Information Sciences}, 494:\penalty0 311--327, 2019{\natexlab{a}}.

\bibitem[Tanveer et~al.(2019{\natexlab{b}})Tanveer, Tiwari, Choudhary, and Jalan]{tanveer2019sparse}
M.~Tanveer, A.~Tiwari, R.~Choudhary, and S.~Jalan.
\newblock Sparse pinball twin support vector machines.
\newblock \emph{Applied Soft Computing}, 78:\penalty0 164--175, 2019{\natexlab{b}}.

\bibitem[Tanveer et~al.(2021)Tanveer, Sharma, and Suganthan]{tanveer2021least}
M.~Tanveer, A.~Sharma, and P.~N. Suganthan.
\newblock Least squares {KNN}-based weighted multiclass twin {SVM}.
\newblock \emph{Neurocomputing}, 459:\penalty0 454--464, 2021.

\bibitem[Tanveer et~al.(2022{\natexlab{a}})Tanveer, Ganaie, Bhattacharjee, and Lin]{tanveer2022intuitionistic}
M~Tanveer, M.~A. Ganaie, A~Bhattacharjee, and C~T Lin.
\newblock Intuitionistic fuzzy weighted least squares twin svms.
\newblock \emph{IEEE Transactions on Cybernetics}, 53\penalty0 (7):\penalty0 4400--4409, 2022{\natexlab{a}}.

\bibitem[Tanveer et~al.(2022{\natexlab{b}})Tanveer, Rajani, Rastogi, Shao, and Ganaie]{tanveer2022comprehensive}
M.~Tanveer, T.~Rajani, R.~Rastogi, Y.~H. Shao, and M.~A. Ganaie.
\newblock Comprehensive review on twin support vector machines.
\newblock \emph{Annals of Operations Research}, pages 1--46, 2022{\natexlab{b}}.
\newblock \doi{https://doi.org/10.1007/s10479-022-04575-w}.

\bibitem[Tanveer et~al.(2022{\natexlab{c}})Tanveer, Tiwari, Choudhary, and Ganaie]{tanveer2022large}
M.~Tanveer, A.~Tiwari, R.~Choudhary, and M.~A. Ganaie.
\newblock Large-scale pinball twin support vector machines.
\newblock \emph{Machine Learning}, pages 1--24, 2022{\natexlab{c}}.
\newblock \doi{https://doi.org/10.1007/s10994-021-06061-z}.

\bibitem[Wang et~al.(2019)Wang, Yang, Liu, and Fujita]{wang2019study}
Hao Wang, Yan Yang, Bing Liu, and Hamido Fujita.
\newblock A study of graph-based system for multi-view clustering.
\newblock \emph{Knowledge-Based Systems}, 163:\penalty0 1009--1019, 2019.

\bibitem[Wang et~al.(2023)Wang, Zhu, and Zhang]{wang2023safe}
Huiru Wang, Jiayi Zhu, and Siyuan Zhang.
\newblock Safe screening rules for multi-view support vector machines.
\newblock \emph{Neural Networks}, 166:\penalty0 326--343, 2023.

\bibitem[Xiao et~al.(2024)Xiao, Pan, Liu, Zhao, Kong, and Hao]{xiao2024privileged}
Yanshan Xiao, Guitao Pan, Bo~Liu, Liang Zhao, Xiangjun Kong, and Zhifeng Hao.
\newblock Privileged multi-view one-class support vector machine.
\newblock \emph{Neurocomputing}, 572:\penalty0 127186, 2024.

\bibitem[Xie(2020)]{xie2020multi}
Xijiong Xie.
\newblock Multi-view semi-supervised least squares twin support vector machines with manifold-preserving graph reduction.
\newblock \emph{International Journal of Machine Learning and Cybernetics}, 11:\penalty0 2489--2499, 2020.

\bibitem[Xie and Sun(2015)]{xie2015multi}
Xijiong Xie and Shiliang Sun.
\newblock Multi-view twin support vector machines.
\newblock \emph{Intelligent Data Analysis}, 19\penalty0 (4):\penalty0 701--712, 2015.

\bibitem[Xie et~al.(2023{\natexlab{a}})Xie, Li, and Sun]{xie2023deep}
Xijiong Xie, Yanfeng Li, and Shiliang Sun.
\newblock Deep multi-view multiclass twin support vector machines.
\newblock \emph{Information Fusion}, 91:\penalty0 80--92, 2023{\natexlab{a}}.

\bibitem[Xie et~al.(2023{\natexlab{b}})Xie, Sun, Qian, Guo, Zhang, Ye, and Wang]{xie2023laplacian}
Xijiong Xie, Feixiang Sun, Jiangbo Qian, Lijun Guo, Rong Zhang, Xulun Ye, and Zhijin Wang.
\newblock Laplacian {L}p norm least squares twin support vector machine.
\newblock \emph{Pattern Recognition}, 136:\penalty0 109192, 2023{\natexlab{b}}.

\bibitem[Yang et~al.(2020)Yang, Jiang, Tian, Wang, Zhou, and Fujita]{yang2020inverse}
Xiaohui Yang, Xiaoying Jiang, Chenxi Tian, Pei Wang, Funa Zhou, and Hamido Fujita.
\newblock Inverse projection group sparse representation for tumor classification: A low rank variation dictionary approach.
\newblock \emph{Knowledge-Based Systems}, 196:\penalty0 105768, 2020.

\bibitem[Yang et~al.(2023)Yang, Hua, Zhang, Fan, Zhang, Ye, and Fu]{yang2023preferred}
Xubing Yang, Zhichun Hua, Li~Zhang, Xijian Fan, Fuquan Zhang, Qiaolin Ye, and Liyong Fu.
\newblock Preferred vector machine for forest fire detection.
\newblock \emph{Pattern Recognition}, 143:\penalty0 109722, 2023.

\bibitem[Ye et~al.(2021)Ye, Huang, Zhang, Zheng, Fu, and Yang]{ye2021multiview}
Qiaolin Ye, Peng Huang, Zhao Zhang, Yuhui Zheng, Liyong Fu, and Wankou Yang.
\newblock Multiview learning with robust double-sided twin {SVM}.
\newblock \emph{IEEE Transactions on Cybernetics}, 52\penalty0 (12):\penalty0 12745--12758, 2021.

\bibitem[Zhao et~al.(2017)Zhao, Xie, Xu, and Sun]{zhao2017multi}
Jing Zhao, Xijiong Xie, Xin Xu, and Shiliang Sun.
\newblock Multi-view learning overview: Recent progress and new challenges.
\newblock \emph{Information Fusion}, 38:\penalty0 43--54, 2017.

\bibitem[Zhu et~al.(2022)Zhu, Wang, Li, and Zhang]{zhu2022fast}
Jiayi Zhu, Huiru Wang, Hongjun Li, and Qing Zhang.
\newblock Fast multi-view twin hypersphere support vector machine with consensus and complementary principles.
\newblock \emph{Applied Intelligence}, 52\penalty0 (11):\penalty0 12684--12703, 2022.

\end{thebibliography}
\bibliographystyle{plainnat}
\end{document}